\documentclass{article}

\usepackage{arxiv}
\usepackage[utf8]{inputenc} 
\usepackage[T1]{fontenc}    
\usepackage{hyperref}       
\usepackage{url}            
\usepackage{booktabs}       
\usepackage{amsfonts}       
\usepackage{nicefrac}       
\usepackage{microtype}      
\usepackage{lipsum}		
\usepackage{graphicx}
\usepackage{doi}
\usepackage{comment}
\usepackage{amsfonts}
\usepackage{amssymb}
\usepackage{amsmath}
\usepackage[authoryear]{natbib} 
\usepackage[table]{xcolor}
\usepackage{subcaption}
\usepackage{graphicx}
\usepackage[table]{xcolor}
\usepackage[normalem]{ulem}
\usepackage{lscape}
\usepackage{longtable}
\usepackage{stmaryrd}
\usepackage{dsfont}
\usepackage{multirow}

\usepackage{tikz}
\usetikzlibrary{arrows.meta}

\usepackage{afterpage} 
\usepackage{rotating}
\newcommand{\modified}[1]{#1}

\author{ Céline Finet \\
	INRIA Rennes\\
	France\\
	\texttt{celine.finet@inria.fr} \\
	\And
	Stephane Da Silva Martins \\
	SATIE - CNRS\\ 
	Université Paris-Saclay\\
	France\\
	\texttt{stephane.da-silva-martins@universite-paris-saclay.fr} \\
	\And
	\href{https://orcid.org/0000-0002-3323-0510}{\includegraphics[scale=0.06]{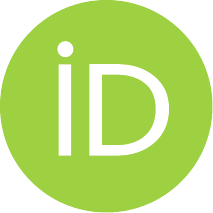}\hspace{1mm}Jean-Bernard Hayet} \\
	Department of Computer Science\\
	Centro de Investigación en Matemáticas\\
	Guanajuato, México \\
     \And
     \href{https://orcid.org/0009-0000-4315-6556}{\includegraphics[scale=0.06]{images/orcid.pdf}\hspace{1mm}Ioannis Karamouzas} \\
      University of California, Riverside\\
      USA \\
	 \And
	 \href{https://orcid.org/0000-0003-4484-7134}{\includegraphics[scale=0.06]{images/orcid.pdf}\hspace{1mm}Javad Amirian} \\
     Sorbonne Université, Paris\\
     France\\
	 \And    
	 \href{https://orcid.org/0000-0001-8494-2289}{\includegraphics[scale=0.06]{images/orcid.pdf}\hspace{1mm}Sylvie Le Hégarat-Mascle} \\
	SATIE - CNRS\\ 
	Université Paris-Saclay\\
	France\\
	 \And  
	 \href{https://orcid.org/0000-0003-1812-1436}{\includegraphics[scale=0.06]{images/orcid.pdf}\hspace{1mm}Julien Pettré} \\
   	 INRIA Rennes\\
	 France\\
     \And
	\href{https://orcid.org/0000-0001-7065-4809}{\includegraphics[scale=0.06]{images/orcid.pdf}\hspace{1mm}Emanuel Aldea} \\
SATIE - CNRS\\ 
	Université Paris-Saclay\\
	France\\
}

\date{}

\renewcommand{\shorttitle}{Recent Advances in Multi-Agent Human Trajectory Prediction: A Comprehensive Review}

\hypersetup{
pdftitle={A template for the arxiv style},
pdfsubject={q-bio.NC, q-bio.QM},
pdfauthor={David S.~Hippocampus, Elias D.~Striatum},
pdfkeywords={First keyword, Second keyword, More},
}

\usepackage{subcaption}
\usepackage{tabularx}
\newcolumntype{Y}{>{\centering\arraybackslash}X}

\title{Recent Advances in Multi-Agent Human Trajectory Prediction: A Comprehensive Review}

\begin{document}




\maketitle 

\begin{abstract}
With the emergence of powerful data-driven methods in human trajectory prediction (HTP), gaining a finer understanding of multi-agent interactions lies within hand's reach, with important implications in areas such as social robot navigation, autonomous \modified{driving}, and crowd modeling. This survey reviews some of the most recent advancements in deep learning-based multi-agent trajectory prediction, focusing on studies published between 2020 and 2025. We categorize the existing methods based on their architectural design, their input representations, and their overall prediction strategies, placing a particular emphasis on models evaluated using the ETH/UCY benchmark. Furthermore, we highlight key challenges and future research directions in the field of multi-agent HTP.
\end{abstract}
\keywords{Review, survey, human trajectory prediction, deep learning, robotics, autonomous navigation, crowd modeling}

\section{Introduction}

Human Trajectory Prediction (HTP) focuses on forecasting the future movements of individuals or groups given past observations and application-dependent context. For instance, in the context of robotics, accurate trajectory prediction is fundamental for safe and socially compliant navigation, enabling autonomous systems to anticipate human motion and adapt their behaviors in shared spaces ~\citep{Rudenko2020, core, stratton2024}. Historically, physics-based (expert-driven) approaches have been central: the Social Force Model (SFM) introduced by  \cite{Helbing_1995} conceptualizes pedestrians as particles influenced by attractive goals and repulsive obstacles, while velocity-based methods such as Reciprocal Velocity Obstacles (RVO) \citep{vandenBerg2008} and Optimal Reciprocal Collision Avoidance (ORCA) \citep{vandenBerg2011} reason about collision-free velocities and have been widely used for trajectory prediction in crowds. 

\begin{figure}[ht]
    \centering
    \includegraphics[width=1\linewidth]{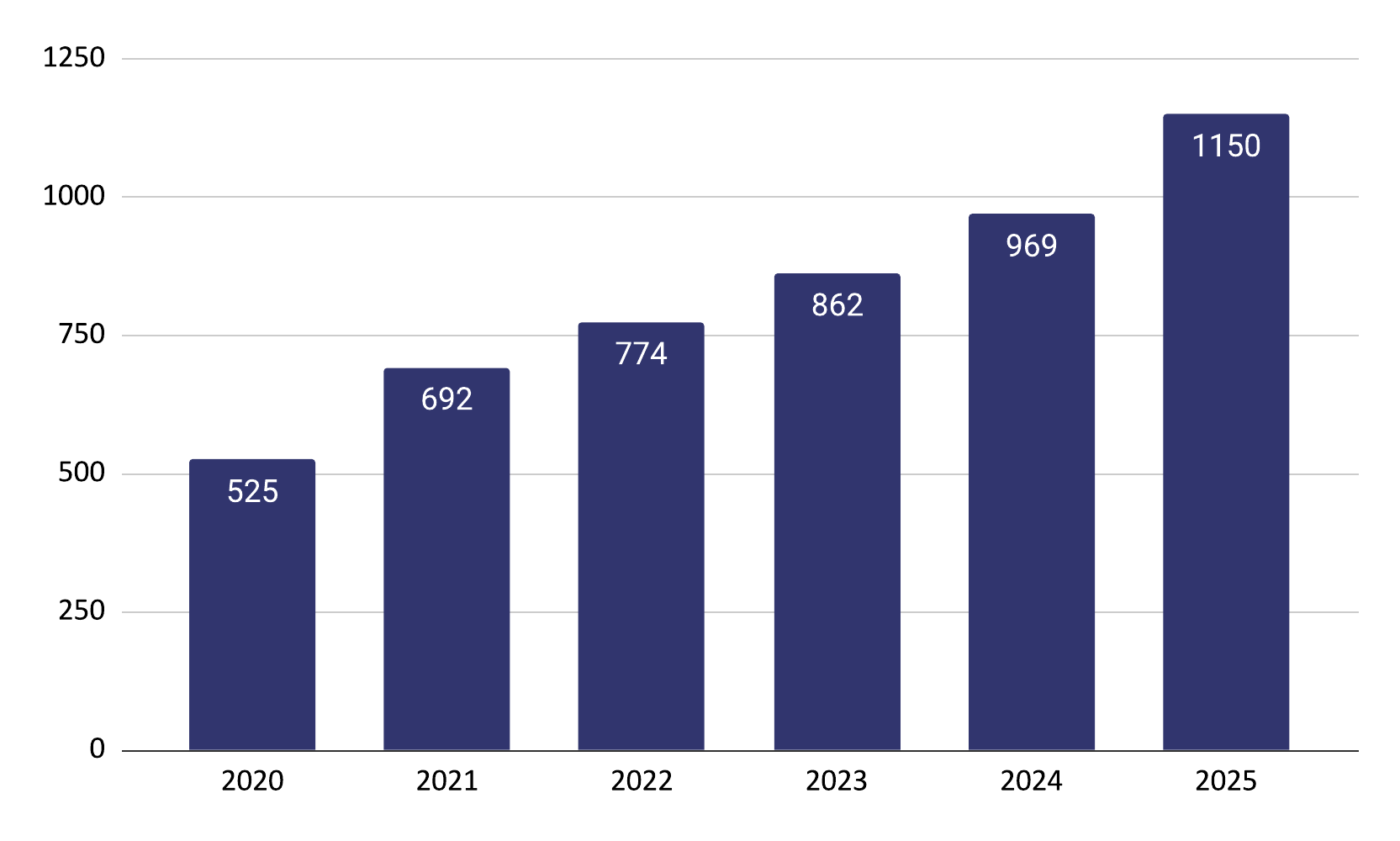}
    \caption{Evolution of the yearly number of Human Trajectory Prediction (HTP) publications indexed in Google Scholar, from 2020 to 2025. 
    The plot shows a consistent upward trend \modified{in the period}, culminating in a total of \modified{4,972} publications over the six-year period. 
    }
    \label{fig:chart}
\end{figure}

In their influential survey, ~\citet{Rudenko2020} organized the HTP literature into two broad paradigms: data-driven models that learn statistical regularities from data, and expert-driven models grounded in physics and hand-crafted rules. Since then, comparative studies on standard benchmarks (e.g., ETH/UCY, SDD) have consistently shown that deep, data-driven approaches achieve significantly lower displacement errors than classical physics-based baselines. \modified{For instance, the seminal work on Social-LSTM \citep{Alahi_2016} demonstrated a substantial reduction in Average Displacement Error (ADE) compared to the Social Force Model \citep{Helbing_1995}, highlighting the advantages of learned interaction modeling over hand-crafted rules.}
This performance gap, together with the lack of new work on physics-based approaches, explains why, in this study, we have decided to focus only on data-driven approaches~\citep{Review_pedestrian}. 
Since the publication of Rudenko's survey, HTP data-driven models have experienced rapid advances, particularly in their use of deep learning techniques. Among these advancements, multi-agent models have become a key area of focus, as they address the intricate interactions and dependencies between individuals in dynamic environments.
Figure~\ref{fig:chart} illustrates the steady increase in the number of papers on HTP over the years. 
These models aim to capture the collective motion patterns of groups while maintaining a high level of accuracy for individual predictions, making them essential for many applications, as shown in Table~\ref{tab:app}.

\begin{figure}[ht]
    \centering
    \includegraphics[width=1.02\linewidth]{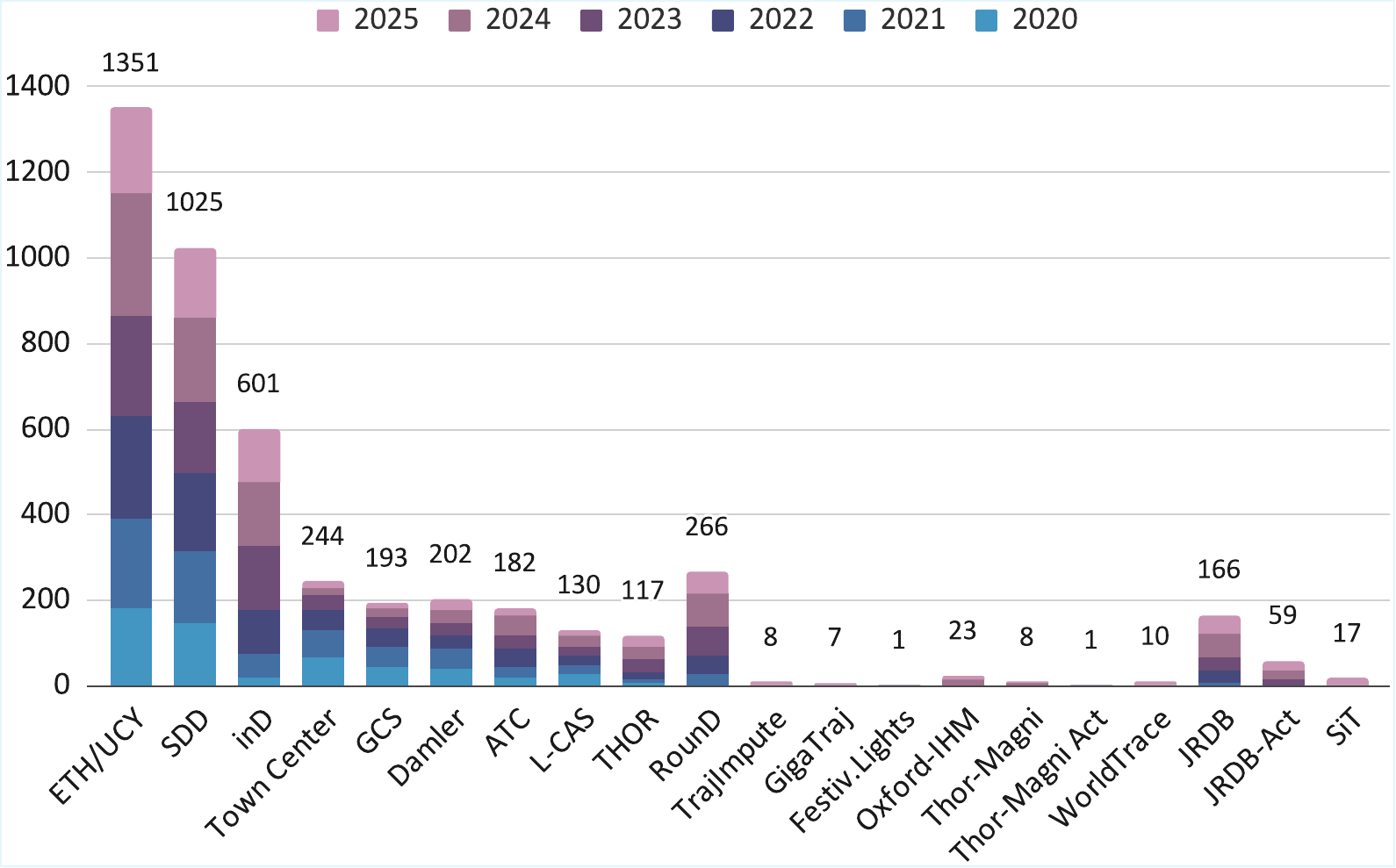}
    \caption{\modified{Yearly citation counts (2020–2025) for major trajectory‑prediction datasets, based on Google Scholar citations of their reference papers.}  
    The stacked bars show the yearly citation counts, highlighting the continued dominance of the ETH/UCY and SDD datasets, followed by inD and RoundD, while more recent datasets such as JRDB and SiT Dataset show growing interest. 
    }
    \label{fig:citations-by-dataset}
\end{figure}

As HTP embraces data-driven models, the availability of high-quality datasets becomes essential to learn intricate patterns of human motion and interactions. Consequently, datasets and benchmarks are not just tools but foundational elements that enable meaningful advancements in this domain. Among the most widely 
\modified{referenced} datasets, the ETH/UCY benchmark~\citep{ETH, Lerner_2007} has emerged as the \textit{de facto} main benchmark for evaluating State-Of-The-Art (SOTA) HTP methods on socially interactive environments, as it provides rich scenarios with diverse real-world interactions, a trend corroborated by our citation analysis in Figure~\ref{fig:citations-by-dataset}, showing ETH/UCY as the most cited dataset. However, while these datasets provide a solid foundation, effectively capturing intricate interactions within dynamics remains a significant challenge. Many existing approaches still focus on single-agent prediction or fail to adequately address the complexities of modeling agent interactions in dynamic, multi-agent settings.
 
\begin{figure}[t]
    \centering
    \begin{subfigure}{0.43\linewidth}
        \centering
        \includegraphics[width=\linewidth]{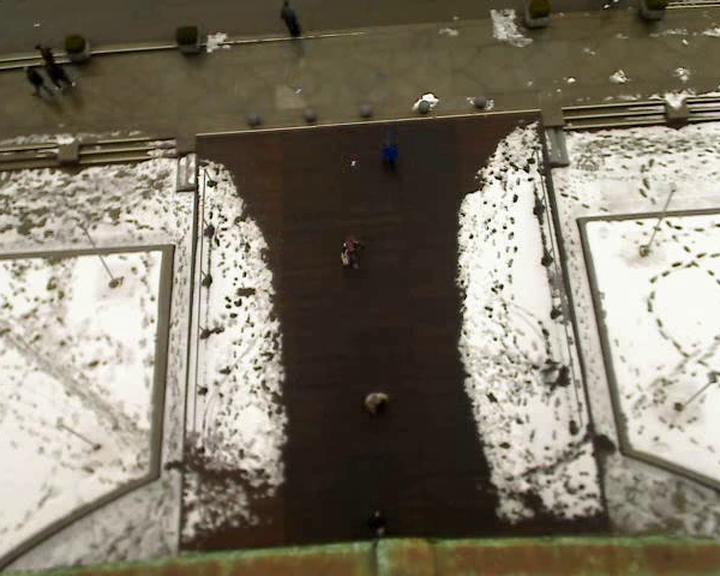}
        \caption{\centering ETH}
    \end{subfigure}
    \hfill
    \begin{subfigure}{0.43\linewidth}
        \centering
        \includegraphics[width=\linewidth]{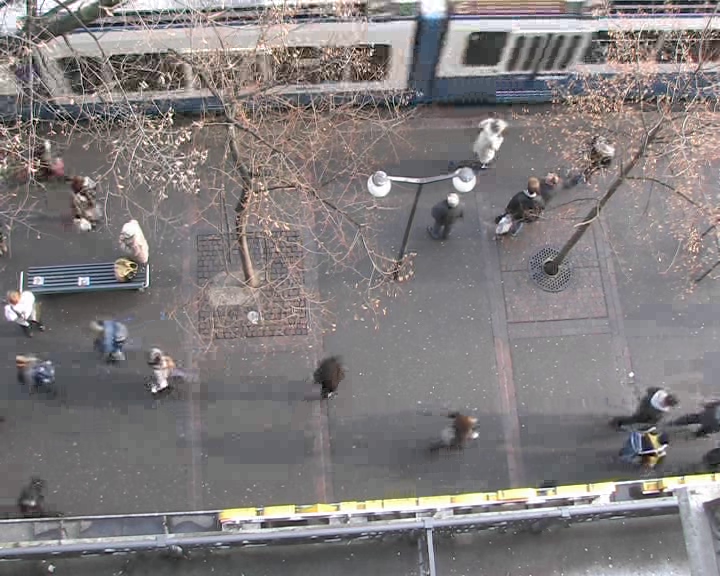}
        \caption{\centering HOTEL}
    \end{subfigure}
    \hfill
    \begin{subfigure}{0.45\linewidth}
        \centering
        \includegraphics[width=\linewidth]{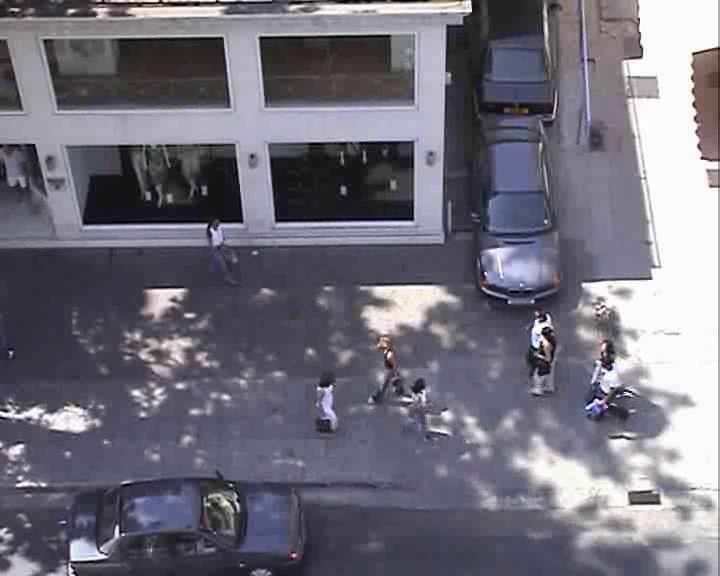}
        \caption{\centering ZARA}
    \end{subfigure}
    \hfill
    \begin{subfigure}{0.45\linewidth}
        \centering
        \includegraphics[width=\linewidth]{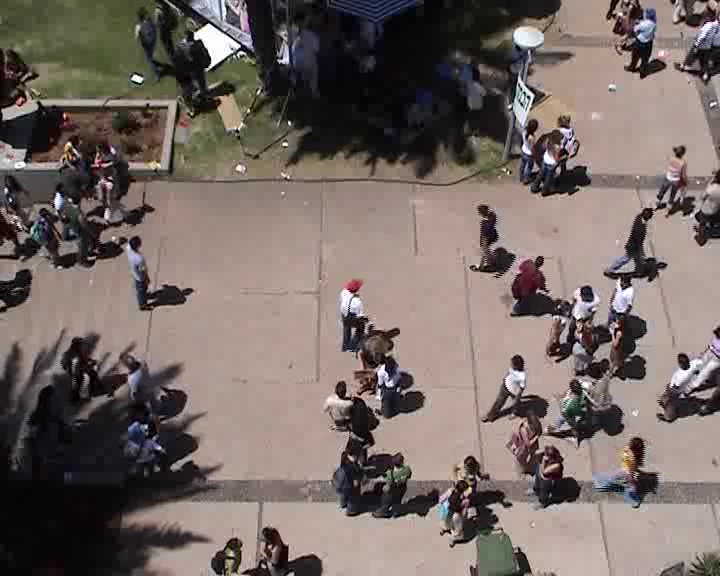}
        \caption{\centering UNIV}
    \end{subfigure}
    \caption{Scenes of the ETH/UCY benchmark. 
    }
    \label{fig:eth_ucy}
\end{figure}

This survey covers the most significant progress in HTP since~\cite{Rudenko2020} with a focused review of state-of-the-art multi-agent trajectory prediction approaches that utilize deep learning techniques and a particular emphasis on methods evaluated using the ETH/UCY dataset (illustrated Figure~\ref{fig:eth_ucy}).

\subsection{Problem Definition}

Consider a scene where $N$ agents are observed over $T_{\text{obs}}$ time steps, 
with their state vectors given by:
\begin{center}
    $\mathbf{S}_{1:T_{\text{obs}}} = 
    \left\{ \mathbf{s}_t^i \in \mathbb{R}^d 
    ~\middle|~ t \in \llbracket 1, T_{\text{obs}} \rrbracket,~ 
    i \in \llbracket 1, N \rrbracket \right\}.$
\end{center}

Each state $\mathbf{s}_t^i$ usually includes position, velocity, orientation, and other dynamic features relevant to the agent's motion.
The objective of a multi-agent trajectory prediction model is to generate a set of $K$ plausible future trajectories over $T_{\text{pred}}$ time steps:
\begin{center}
\scalebox{0.9}{
    $\hat{\mathbf{Y}}^{k} = 
    \left\{ \hat{\mathbf{y}}_t^{i,k} \in \mathbb{R}^2 
    ~\middle|~ t \in \llbracket T_{\text{obs}}+1, T_{\text{pred}} \rrbracket,~ 
    i \in \llbracket 1, N \rrbracket,~ k \in \llbracket 1, K \rrbracket \right\}.$
}
\end{center}

The corresponding ground-truth trajectories are:
\begin{center}
    $\mathbf{Y}_{\text{gt}} = 
    \left\{ \mathbf{y}_t^i \in \mathbb{R}^2 
    ~\middle|~ t \in \llbracket T_{\text{obs}}+1, T_{\text{pred}} \rrbracket,~ 
    i \in \llbracket 1, N \rrbracket \right\}.$
\end{center}

\subsection{Terminology}

\paragraph{} An \emph{agent} is any entity whose movement is modeled, tracked, and predicted within a specific environment. In the context of human trajectory prediction, agents typically refer to pedestrians.

\paragraph{} The \emph{environment} consists of spatial and contextual elements that may influence agents movements. This refers to either static elements (e.g.\modified{,} obstacles) or dynamic elements (e.g.\modified{,} other agents) that constrain or modify the trajectories of the agents.

\paragraph{} \emph{Single-agent prediction} refers to forecasting a single pedestrian's future trajectory based on its past behavior and the available context. It focuses exclusively on the individual agent's movement and intentions without considering its interactions with other agents. On the contrary, \emph{multi-agent prediction} forecasts the trajectories of multiple agents by considering their interactions, including the effects of social dynamics, shared environments, potential collisions or cooperation, making the task complex since it should capture the interdependencies in their behaviors.

\paragraph{} \emph{Joint prediction} refers to multi-agent prediction 
where the trajectories of multiple pedestrians are predicted \emph{simultaneously}, explicitly modeling their mutual social interactions.
In contrast, \emph{marginal prediction} estimates each pedestrian's trajectory individually, while still accounting for social interactions with other agents. 

\paragraph{} \emph{Direct prediction} involves generating the entire trajectory in a single step, directly mapping past state vectors into future coordinates. On the contrary, \emph{auto-regressive prediction} predicts the trajectory recursively, generating one state at a time and using this output as input for the subsequent prediction.

\paragraph{} \emph{Target-conditioned prediction} forecasts trajectories using long-term goals or intermediate waypoints as explicit latent variables within the trajectory inference process.

\paragraph{} \emph{Prediction refinement} iteratively improves an initial coarse trajectory prediction with additional refinement steps. 

\begin{figure*}[ht]
    \centering
    \includegraphics[width=1\linewidth]{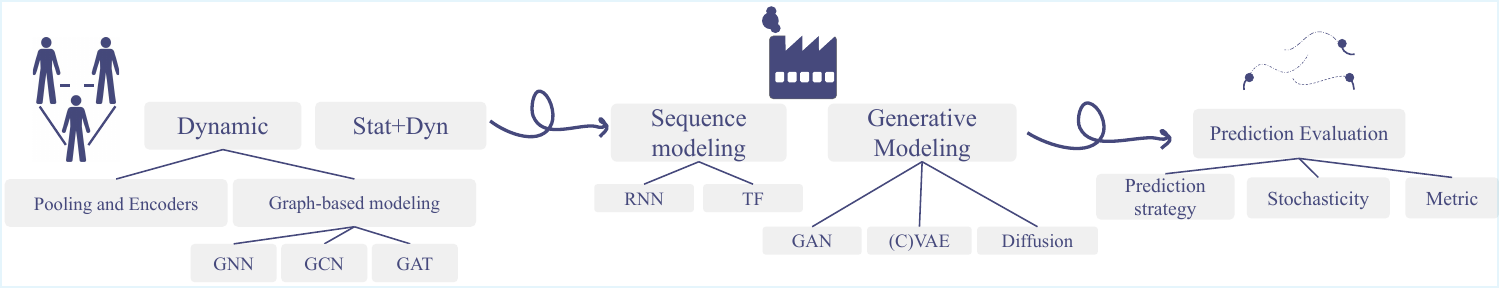}
   \caption{An overview of the global framework for HTP algorithms that also describes the structure of this paper. The context part implies embedding contextual information that can be dynamic or 
   both static and dynamic in the network (developed in Section~\ref{subsection:Context}). The Backbone architecture is the general framework of the model, which takes the embedded context and processes it with a sequential approach, a generative one or both (developed in Section~\ref{subsection:archi}). The \modified{Prediction} and Evaluation \modified{parts} involve the way the predicted trajectories are generated and evaluated, which can be the prediction strategy, the choice between joint and marginal prediction, the handling of the stochasticity, and finally the evaluation metrics (developed in Section~\ref{subsection:output}).}
    \label{fig:overview}
\end{figure*}

\begin{table*}[ht]
\centering
\small
\begin{tabular}{|l|p{13cm}|}
\hline
\textbf{Category} & \textbf{Description} \\ \hline
\textbf{Robotic Navigation} & Robots' effectiveness in dynamic environments relies on their ability to anticipate human movements and future trajectories. Socially acceptable interactions allow robots to navigate and collaborate with humans by accurately forecasting behaviors and responding in real-time to environmental changes~\citep{Human--robot, trautman2015robot, nishimura2020risk,biswas2022socnavbench,gao2022evaluation, core, muchen2024mixed, stratton2024}.
 \\ \hline
\textbf{Autonomous Vehicles} & 
Companies developing self-driving technology need robust embedded algorithms to predict pedestrian and vehicle movements to ensure passenger and pedestrian safety~\citep{monosurvey}. 
\modified{Recent collaborations between academia and industry (e.g., automotive OEMs and AV companies) have further underscored the importance of reliable trajectory prediction in real‑world deployment scenarios~\citep{TrajPRedZhou_2024}. Numerous challenges have arisen in this context, such as the the CARLA challenge~\citep{CARLA}, the Lyft Motion Prediction for Autonomous Vehicles challenge~\citep{houston2021one}, or the Waymo Open Dataset Challenges~\citep{waymo}.}
\\ \hline
\textbf{Public Safety} &  Safety agencies use HTP to monitor crowds and prevent incidents by identifying abnormal movements. The organization of challenges like TrajNet++~\citep{kothari2021} or GigaTraj~\citep{Lin2024gigatraj} highlights the importance of accurate trajectory prediction in public safety applications.\\ \hline
\textbf{Planning} & Physical simulation models using HTP algorithms are valuable for architects and system designers in crowd dynamics and safety planning. \modified{They can be useful for space dimensioning and evacuation optimization~\citep{egress2017,hu2020predicting,Kun2024}}, but typically provide short-term predictions. For large environments or emergencies, there is a crucial need for more robust, longer-term forecasting algorithms. \\ \hline
\textbf{Crowd Modeling} & Crowd modeling and HTP are inherently dual problems. While HTP predicts future trajectories of individuals in dynamic environments, crowd simulation models and generates realistic crowd behaviors. As crowd simulation increasingly adopts data-driven methods, it benefits from insights provided by HTP and leverages large human trajectories datasets  to 
\modified{enable real-time predictions and more realistic simulations in large-scale environments like massive human gatherings~\citep{pelechano2008virtual, pelechano2016simulating, vanToll2021Algorithms, Survey_crowd_modelling, lemonari2022authoring}}. \\ \hline
\textbf{Sports Analytics} & Predicting trajectories using HTP can provide strategic insights into players' decision-making processes, team coordination, and tactical analysis across various team sports, including basketball, baseball and soccer~\citep{NBA, xu2024sports}.\\ \hline
\textbf{Retail Applications} & Trajectory prediction can also optimize store layouts or locations based on customer movement patterns for marketing strategy optimization
~\citep{li2019mobile,retail}. \\ \hline
\end{tabular}
\vspace{0.25cm}
\caption{Applications of Human Trajectory Prediction (HTP).}
\label{tab:app}
\end{table*}

\subsection{Survey organization}
\label{section:core}

Section~\ref{sec:dataset} presents the main datasets and benchmarks used to evaluate HTP models, describing their sensing modalities, annotation types, and crowd characteristics.%

Section~\ref{section:survey} presents our systematic survey protocol and critically reviews previous surveys on human trajectory prediction, specifying their scope and limitations. The section further provides a rationale for focusing on multi-agent models evaluated using the ETH/UCY benchmarks.

Forecasting pedestrian paths is an inherently complex challenge, as it involves accounting for multiple factors such as pedestrians' intentions, social interactions, and interactions with the surrounding static or dynamic environment. To tackle this complexity, many proposed solutions have adopted \emph{modular} architectures, with each module focusing on a specific aspect of the problem. Given this modularity, we have structured the discussion of architectures into three key sections illustrated in Figure~\ref{fig:overview}.  

Section~\ref{subsection:Context} examines how different models represent the environment and agents, focusing on both static and dynamic contextual information. We also highlight how graph-based representations and attention mechanisms are used to capture dynamic obstacles or static features in the environment.

Section~\ref{subsection:archi} reviews the backbone architectures of the prediction models themselves. We categorize the models according to their primary architectural choices, such as generative models, Transformers, and recurrent networks, and explain how these architectures interact with the contextual inputs to generate predictions.

Section~\ref{subsection:output} explores the methods used for trajectory generation, including prediction strategies, stochastic prediction, and the distinction between marginal and joint prediction methods. We also review how the predictions are evaluated using various metrics.

Finally, Section~\ref{sec:discussion} provides a transversal discussion of current limitations and future directions in HTP, focusing on evaluation metrics, datasets, architectural choices, and challenges toward real-world deployment.

\section{Datasets}
\label{sec:dataset}

Prediction performance is usually evaluated by comparing predicted trajectories or predictive distributions with ground-truth motion data. Qualitatively, researchers frequently visualize predicted paths alongside observed movements to assess the alignment between model outputs and ground truth pedestrian trajectories They also often examine failure cases, such as abrupt direction changes or interactions with dynamic obstacles, to identify the model limitations and future improvements.

To collect the data necessary for the previously mentioned quantitative and qualitative analysis, different sensors have been considered: fixed overhead cameras or range-sensing systems such as LiDAR, depth sensors, or stereo rigs, which may be static or ego-centric. Each pedestrian receives a persistent identifier, and positions are recorded in global or image coordinates $(x, y)$ with corresponding timestamps $t$, resulting in a set of ordered state vectors. \modified{ Beyond basic positions, these datasets frequently incorporate dynamic features (e.g., velocity, acceleration), orientation or gaze direction, as well as higher-level contextual information like social grouping or behavioral labels. }
Benchmark datasets for trajectory prediction are constructed to capture diverse interaction patterns, varying crowd densities, complex environmental geometries, semantic scene context, and temporally continuous observations of pedestrian motion.

Table~\ref{tab:dataset} reviews datasets used for pedestrian trajectory prediction. These datasets provide the empirical foundation for HTP research, supporting quantitative evaluation and comparison of modeling approaches. Although they differ in sensing modality, spatial coverage, and annotation details, all aim to capture human motion in a structured and measurable manner.

Furthermore, the inclusion of citation counts since 2020 highlights the evolving relevance of each dataset. Highly cited datasets such as ETH/UCY and SDD remain essential benchmarks for a fair comparison across studies, whereas more recent datasets (e.g., TrajImpute, Festival of Lights, GigaTraj) introduce novel sensing modalities or richer annotations that capture social intent and multimodal cues. Large-scale datasets such as JRDB and THÖR further expand available annotations by providing 3D trajectories, full-body skeletons, orientation angles, social group labels, and multimodal sensor streams (RGB, depth, LiDAR).



\begin{sidewaystable*} 
\centering
\scriptsize

\resizebox{0.85\textwidth}{!}{%
\begin{tabular}{
p{2.9cm}  
p{1.5cm}  
p{2cm}    
p{2.5cm}  
p{1.6cm}  
p{6.5cm}  
p{1.6cm}  
p{1.4cm}  
p{1.6cm}  
}
\toprule
\textbf{Dataset} & \textbf{Duration} & \textbf{Num. Pedestrians} & \textbf{Annotations} & \textbf{Num. Citations} & \textbf{Description} & \textbf{Avg Count/Frame} & \textbf{Max Count/Frame} & \textbf{Std Count/Frame}\\
\midrule

ETH/UCY~\cite{ETH, Lerner_2007} & 54.5 min & 2,206 & 2D Positions & 1,350 & Outdoor trajectories from elevated cameras recorded across five scenes~(ETH, HOTEL, UNIV, ZARA1, and ZARA2)& 14.96 & 75 & 4.89\\ \hline
SDD~\cite{sdd} & 5 hours & 1,036 & Bounding Boxes & 1,030 & Outdoor trajectories from overhead cameras recorded across eight campus scenes & 7.86 & 32 & 4.7\\\hline
inD~\cite{ind} & 10 hours & 2,900 & 2D posititons, Velocity, Acceleration and Heading orientation & 601 & Outdoor trajectories from overhead cameras recorded at four German intersections & 3.15 & 15 & 1.79\\\hline
Town Center Dataset~\cite{town} & 5 min & 230 & Bounding Boxes & 243 & Outdoor trajectories from a fixed surveillance camera recorded on a pedestrian street in Oxford & 15.88 & 28 & 4.7\\\hline
Grand Central Station Dataset~\cite{gcs} & 33.20 min & 17,000 & 2D Positions & 203 & Indoor trajectories from an elevated camera recorded in Grand Central Terminal (New York) & 79.5 & 289 & 48.5\\\hline
Daimler Pedestrian Path Prediction Dataset~\cite{daimier} & 4.53 min & 68 & 2D Positions, Bounding Boxes and Stereo Images & 202 & Outdoor trajectories from vehicle-mounted stereo cameras recorded in urban traffic & 23.13 & 58 & 19.8\\\hline
ATC~\cite{atc} & 92 days & - & 3D Positions, Velocity and Facing Angle & 182 & Indoor trajectories from 3D range sensors recorded in a shopping mall & 102.03 & 1818 & 41.7\\\hline
L-CAS~\cite{lcas} & 37 min & 6,140 single-person and 3,054 group & 3D Positions & 130 & Indoor trajectories from 3D LiDAR recorded in the Minerva Building at the University of Lincoln & 2.25 & 58 & 2.98\\\hline
THÖR~\cite{thor} & 60 min & 600+ & 3D Positions, 3D Head Orientation and Eye Gaze & 119 & Indoor trajectories from motion-capture sensors recorded in a lab with head orientation and gaze & 4.3 & 9 & 2.18\\\hline
rounD~\cite{round} & 6 hours & 25 & Bounding Boxes & 26 & Outdoor trajectories from overhead cameras recorded at German roundabouts & – & – & –\\\hline
SiT Dataset~\cite{sit} & - & - & 2D and 3D bounding boxes & 17 & Indoor/outdoor trajectories from 360$^\circ$ cameras and 3D LiDAR recorded by a mobile robot & 37.84 & 126 & 12.95\\\hline
TrajImpute~\cite{trajimpute} & 54.5 min & 2,206 & 2D Positions & 8 & Outdoor trajectories from elevated cameras recorded in ETH/UCY scenes with synthetically missing observations & – & – & –\\\hline
GigaTraj~\cite{Lin2024gigatraj} & - & 15,520 & 2D Positions and bounding boxes & 7 & Outdoor trajectories from ultra-wide FOV (gigapixel) cameras recorded in large-scale scenes & 313 & – & –\\\hline
Festival of Lights~\cite{lights} & 1.45 min & 6,735 & 2D Positions & 1 & Outdoor trajectories from elevated cameras recorded during Lyon’s Festival of Lights & 58.98 & 595 & 44.47\\\hline
Oxford-IHM~\cite{oxfordihm} & 60 min & - & 3D Positions, 2D Maps and RGB-D & 23 & Indoor trajectories recorded with fixed and mobile RGB-D cameras and motion-capture systems for human-robot interaction studies & – & – & –\\\hline
THÖR-MAGNI~\cite{magni} & 3.5 hours & 30 & 3D Positions, Head Orientation, Eye Gaze and 3D LiDAR & 20 & Indoor trajectories from human-robot navigation scenarios recorded with multimodal sensors (gaze, LiDAR, robot odometry) & 4.1 & 9 & 1.32\\\hline
THÖR-MAGNI Act~\cite{magniact} & 8.3 hours & 76 & 3D Positions, Head Orientation, Eye Gaze, 3D LiDAR and Human Actions & 1 & Indoor trajectories with synchronized egocentric video and fine-grained action labels for human-robot collaboration & 2.93 & 3 & 0.32\\\hline
WorldTrace~\cite{worldtrace} & 28 months & 2.45M & GPS points & 10 & Large-scale global dataset of human trajectories collected across 70 countries for foundation model pretraining & 361,669.88 & 2,400,000 & 639,553.85\\\hline
JRDB~\cite{jrdb} & 64 min & 3,611 & 2D/3D Bounding Boxes & 166 & Egocentric outdoor/indoor trajectories recorded from a mobile robot equipped with RGB, LiDAR, and odometry sensors & 2.59 & 14 & 1.19\\\hline
JRDB-Act~\cite{jrdbact} & 64 min & 3,611 & 2D/3D Bounding Boxes, Human Actions and social group annotation & 59 & Egocentric indoor/outdoor trajectories with action and group annotations recorded from a mobile robot equipped with RGB, LiDAR, and odometry sensors & 2.59 & 14 & 1.19\\\hline
NBA~\cite{NBAdata} & 626 games & 350+ players & 2D Positions and Ball Trajectory & X & Indoor trajectories from multi-camera tracking in professional basketball arenas & 10 & 10 & 0\\ \hline
Zucker~\cite{zucker} & - & 28 & 2D positions and gyroscopes & 9 & Indoor trajectories of humans interacting with a small differential-drive service robot in a laboratory environment under 3 robot controllers & 2.9 & 4 & 0.37 \\ \hline
TBD~\cite{TBD} & 759 min & 11,716 & 2D/3D Positions & 8 & Indoor trajectories from top-down cameras and ego-centric sensors & - & - & - \\ \hline
EgoTraj-Bench~\cite{egotraj} & 210 min & - & Noisy FPV histories + Clean BEV past/future trajectories & 0 & eal-world ego-centric benchmark pairing noisy first-person video observations with clean BEV ground truth from TBD dataset & - & - & - \\
\bottomrule
\end{tabular}
}
\caption{Overview of pedestrian trajectory datasets, ranked by citation counts accrued from 2020 to  present. The table reports the recording duration, the number of unique pedestrians, the available annotations, and a brief description for each dataset; entries marked with ``-'' indicate information not reported by the original sources. 
\modified{We computed the pedestrian count per frame (average, maximum, and standard deviation) to represent the scene occupancy and crowd size.} }

\label{tab:dataset}
\end{sidewaystable*}


\section{Systematic survey}\label{section:survey}

\subsection{Existing surveys}
\label{subsec:existing_surveys}
    Several surveys have examined the different aspects of HTP and highlighted key advancements and ongoing challenges. Among them,~\cite{Rudenko2020} provides a comprehensive review up to 2019. Complementing this, \cite{kothari2021} reviews deep learning–focused approaches on human trajectory forecasting. \citet{Review_pedestrian} compares deep learning and knowledge-based models, showing that deep networks achieve higher local accuracy, knowledge-based methods better capture collective behaviors. They advocate hybrid frameworks to improve interpretability and scalability.
Given the rapid \modified{developments} in HTP, an updated survey is needed to focus on recent progress, particularly in deep learning-based multi-modal and multi-agent HTP. 
 
Sections~\ref{road_users} and~\ref{social_robot} summarize existing application-oriented surveys, covering domains ranging from autonomous driving and urban pedestrian modeling to socially aware robots and warehouse logistics.

\subsubsection{On the motion of road users.} \label{road_users}

Road users encompass all agents sharing pathways, including vehicles, cyclists and pedestrians. The following surveys have covered different topics related to the motion of road users, and some do intersect with the particular case of human motion prediction. ~\citet{huang2022survey} provide a comparative overview of trajectory-prediction methods developed for autonomous driving, categorizing approaches into the following classes: physics models, classical machine learning, deep learning, and reinforcement learning. \modified{Recently, ~\citet{wang2025trends} revisit the generalization and deployability of motion prediction models across robotics and autonomous driving, highlighting the critical gap between idealized research benchmarks and the complexities of real-world, open-world conditions within closed-loop autonomy stacks.}
 
Other surveys, such as~\cite{Sharma_2022},~\cite{Zhang_2023}, and~\cite{Capy_2023}, focus on intention prediction and behavior modeling for vulnerable road users, providing a detailed analysis of prediction methods used in urban environments and addressing the unique challenges posed by pedestrians and cyclists.
\cite{Kong_2023} expand the discussion to mobility trajectory generation, outlining common application scenarios and providing definitions of mobility trajectory generation. ~\cite{Galvao_2023} cover behavior prediction for pedestrians and vehicles, noting improvements in prediction horizons and accuracy. Additionally,~\cite{Fu_2024} discuss the challenges and future directions of pedestrian trajectory prediction in autonomous driving, underscoring the necessity for robust and accurate prediction algorithms.


\subsubsection{On social robot navigation.} \label{social_robot}
Surveys addressing trajectory and intention prediction methods for socially aware robot navigation analyse the application of human trajectory prediction in robot navigation scenarios~\citep{Singamaneni_2024}. \citep{core} critically highlight the fundamental limitations in current socially aware navigation systems and categorize persistent challenges in planning, behavior design, and evaluation. Finally,~\cite{Almeida_2023} present a comparative analysis of deep learning models used for role-conditioned human motion prediction, using the THOR-Magni dataset~\citep{magni}.

\subsection{Article selection}
\label{section:selection}
 
This survey targets specifically data-driven multi-agent HTP algorithms published between 2020 and 2025.  We aim to provide an up-to-date overview by restricting our scope to recent literature. 

Our survey builds upon the foundation laid by~\cite{Rudenko2020}, employing a systematic approach to identify and analyze relevant studies in this domain, where we have observed a significant increase in the number of publications since 2020 (see Figure~\ref{fig:chart}). We focus on scenarios that involve multiple agents, including those with a large number of participants, which can be particularly complex due to the variety of social interactions. Consequently, we have broadened our search for articles to include studies on less dense crowds, such as those evaluated using the ETH/UCY dataset~\citep{ETH, Lerner_2007}, as it is the most widely adopted benchmark for human trajectory prediction, as illustrated in Figure~\ref{fig:citations-by-dataset}.

\begin{table}[t]
    \centering
    \begin{tabular}{cc}
        \hline
            \textbf{Database} & \textbf{Search keywords} \\
             \hline
             Google Scholar & ``human trajectory prediction'' \\ 
             & and ``ETH/UCY''  Year:2020-2025\\
             \hline
    \end{tabular}
    \caption{Article search keywords in Google Scholar database.}
    \label{tab:prisma}
\end{table}

Using the Preferred Reporting Items for Systematic Reviews and Meta-Analyses (PRISMA) methodology~\citep{moher2009preferred}, we have performed a qualitative analysis of multi-agent HTP studies tested on the ETH/UCY dataset. Our search for relevant articles published from 2020 to 2025 has relied on the Google Scholar database.

\begin{figure}[ht]
    \centering
    \includegraphics[width=0.35\linewidth]{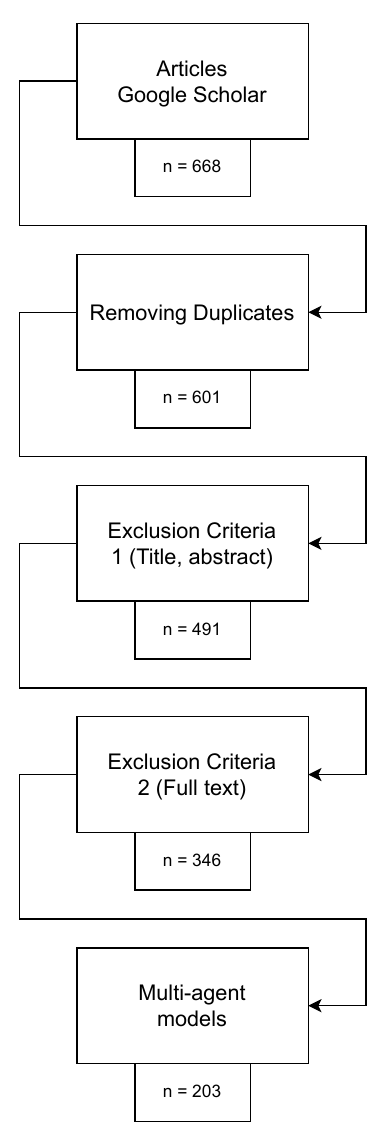}
    \caption{The PRISMA methodology flow chart. The criteria for exclusion in the first-round screening include supplementary materials, non-reviewed papers, Master's and PhD theses, non-open-access papers, and papers with no citations, except for those published in 2024 and 2025. The criteria for exclusion in the second-round screening are papers that do not explicitly mention being tested on the ETH/UCY dataset, those focusing on autonomous driving applications or single-agent prediction. 
    }
    \label{fig:prisma}
\end{figure}

In the initial phase of our research, 
$668$ records have been identified from Google Scholar using the keywords described in Table~\ref{tab:prisma}. 
After the initial search, we have identified the articles, removed duplicates and conducted two rounds of screening to ensure a global evaluation. We have used predefined exclusion criteria during this process to maintain thoroughness. For further insights into our study identification, screening, eligibility, and inclusion process, you can refer to the PRISMA flow chart, presented in Figure~\ref{fig:prisma}. Finally, $203$ papers with direct relevance to our research objectives have been selected for this review.

\section{Contextual information 
representation }
\label{subsection:Context}

This section explores how multi-agent trajectory prediction models incorporate contextual information. \emph{Context} refers to external factors that influence agent trajectories, including the static environment (\textit{e.g}., obstacles like buildings, walls, and other permanent structures) and dynamic elements, such as the movements and behaviors of surrounding agents (pedestrians, vehicles, etc.), temporary obstacles, and environmental changes (\textit{e.g.}, doors opening, furniture being moved, construction, street closures). By examining how models use this information to enhance prediction accuracy, we can better understand their underlying mechanisms.

First, we discuss techniques for encoding \emph{dynamic} contextual data. This involves moving obstacles, such as other agents whose movements may influence the trajectory of the target agent. \modified{We review techniques such as} pooling, encoders, and Graph Neural Networks (GNN), which aggregate and simplify this complex data. Then, we examine models that combine both static and dynamic information. 

\subsection{Dynamic contextual information} \label{subsubsec:dynamic}
\modified{A target agent rarely moves alone: nearby agents can constrain, attract, or deflect its motion based on proximity and social rules. Accounting for these implicit factors is therefore crucial for accurate trajectory prediction. Even if interactions are seldom annotated, most datasets provide the full set of agent positions over time, allowing the agent's surroundings to be inferred from observed dynamics. In the following, we review the main strategies to encode the neighborhood context: \emph{Pooling and Encoders} as early solutions, \emph{Graph-based} methods as the current dominant paradigm for modeling complex interactions, and a final set of alternative approaches.} 
Figure~\ref{fig:histogram} highlights the timeline of those different categories and the papers we have selected to explain their principles.

 
\subsubsection{Pooling and encoders.} \label{subsubsubsection:pool}

Modeling interactions with all surrounding agents is computationally expensive \modified{and conceptually challenging, largely because the number of neighbors varies over time and across scenes}. \emph{Pooling} \modified{addresses} this \modified{issue} by \emph{aggregating} 
\modified{neighborhood information} into a fixed-size representation \modified{that can be processed efficiently}.

\modified{A first line of work achieves this fixed-size encoding} through spatial quantization. SocialGAN~\citep{Gupta_2018} 
\modified{introduced} grid-based pooling 
\modified{by} discretizing the scene
\modified{in an} agent\modified{-centric coordinate frame. This choice is fast and simple, but it imposes a} rigid structure
\modified{that can blur fine spatial cues. Later approaches therefore moved toward \emph{continuous} neighborhood representations to reduce discretization artifacts, e.g., by explicitly constructing spatial context vectors as in SCAN~\citep{SCAN} or by encoding the neighborhood as agent-centered interaction maps processed by CNNs following Social-PEC~\citep{zhao2020noticing},  which can be more interpretable than classical pooling layers while preserving relative positional fidelity. Another limitation of uniform pooling is that it treats all neighbors as} equally \modified{informative, an assumption that} rarely 
\modified{holds} in crowded scenes. 
\modified{R}ecent architectures 
\modified{instead introduce} \emph{selection} mechanisms \modified{ that let the model decide \emph{which, when, and how}} agents matter. 
\modified{This ranges from explicit filtering via ``social'' masks, as in PECNet by~\citet{PECNet} to soft attention that adapts weights to motion states rather than mere distance as in AgentFormer by~\citet{AgentFormer} and ScePT by~\citet{ScePT}.} \modified{In practice, this shift from hard aggregation to attention trades the simplicity of pooling for a richer, context-dependent interaction encoding, at the cost of increased computation when attention becomes dense}.

%
\modified{Beyond geometric cues such as} position and velocity, \modified{a natural next step is to model \emph{what kind of relation} binds agents, i.e., who is moving \emph{with} whom, and under which shared intent, rather than treating neighbors as interchangeable obstacles. One way to do so is to enrich the neighborhood encoding with higher-level \emph{semantic states} that promote coherent collective motion, or to leverage additional modalities to reveal latent group membership, e.g., by fusing visual and acoustic cues to detect grouping behaviors~}\citep{zou2025walks}. 
\modified{Once group structure is treated as a first-class signal, interaction modeling becomes explicitly \emph{relational} and \emph{time-varying}, tracking pairwise and higher-order links as they form and dissolve. This is achieved either with dynamic graph/hypergraph constructions and pooling, as shown in EvolveGraph~\citep{evolvegraph}, EvolveHypergraph~\citep{EvolveHypergraph}, and GPGraph~\citep{GPGraph}, or with memory updates that maintain a latent social context over time}~\citep{SMEMO}. 

\modified{Finally, neighborhood encoding is tightly coupled with \emph{how history is summarized}: interaction cues matter only if they can be propagated through time. Sequential encoders such as RNNs and Variational Recurrent Graph Encoders have been the standard choice~\citep{Trajectron++,MATRIX} but they can be bottlenecked by step-by-step processing and may weaken long-range dependencies. 
Recently, a rapidly emerging trend shifts the representation from the time domain to the frequency domain, using Fourier or wavelet embeddings to make global temporal structures and periodicities more explicit, as in MHTraj~\citep{MHTraj:} and MSWTE-GNN~\citep{MSWTE-GNN}. 
SEI~\citep{SEI} bridges both views by combining multi-scale local convolutional encoders with global attention, so that short-term cues and longer-range social influences remain coherent over time.}

\modified{Overall, pooling-based methods face an inherent trade-off: aggressive pooling (like grids) can oversimplify complex local topology and miss critical safety details, whereas more expressive encoders (like spectral or attention-based) increase representational power but often at the cost of  
computational overhead. They may also reduce interpretability compared to lightweight RNN baselines 
such as TPNMS~\citep{TPNMS} or LB-EBM~\citep{LB-EBM}
.} 


\subsubsection{Graph-based modeling.}\label{subsubsubsection:graph}

Graph-based representations 
\modified{offer a natural alternative to} pooling: 
\modified{instead of collapsing a variable-size neighborhood into a single vector, they keep an explicit interaction structure by modeling} agents as nodes and \modified{their dependencies} as edges. \modified{This explicit relational scaffold is especially useful in crowded scenes, where prediction may depend as much on \emph{who influences whom} as on each agent's individual motion history}. This section surveys graph-based 
\modified{interaction modeling under the broader umbrella of} \emph{Graph Neural Networks} (GNNs), \modified{which include} \emph{Graph Convolutional Networks} (GCNs) and \emph{Graph Attention Networks} (GATs)\modified{. We organize the literature around three recurring design choices: (i) how the interaction topology is built, (ii) how neighbor information is aggregated (often through convolutional or attention-based mechanisms), and (iii) how temporal evolution is handled}.

\paragraph{\modified{From pairwise proximity to group-level relations.}}
\modified{Many early graph formulations start with pairwise edges, typically induced by } distance, \modified{which already captures basic avoidance patterns  as in }CARPe Posterum~\citep{CARPe}. 
\modified{Yet social motion is not always reducible to dyads: groups can emerge, move coherently, and reconfigure. Hypergraphs generalize edges to sets of agents and therefore provide a convenient handle to represent such group-level effects. This perspective is explored in} DynGroupNet~\citep{DynGroupNet}, EvolveHypergraph~\citep{EvolveHypergraph, X11}, and MART~\citep{MART}\modified{, while} OST-HGCN~\citep{OST-HGCN} further 
\modified{separates social vs.\ temporal relations through two dedicated hypergraphs. Importantly, topology is not only about distance-based connectivity but also about \emph{plausibility}: proximity alone can create spurious links. Several works therefore constrain edges with physical or perceptual priors, for instance through physical masks}
\citep{PCHGCN} and collision-aware \modified{designs}~\citep{su2025improving} 
\modified{or through asymmetric pruning aligned with field-of-view constraints as in OV-SKTGCNN~\citep{OVSKTGCNN}.}

\paragraph{\modified{Aggregation: from uniform mixing to selective influence.}}
\modified{Once the graph is defined, the main question becomes how to combine neighbor signals. GCN-style layers offer an efficient baseline by applying shared filters over the graph. Social-STGCNN~\citep{Social-STGCNN}  uses symmetric adjacency normalization to stabilize graph convolutions. To keep the computational load under control when graphs become dense, some approaches restrict propagation to a small number of hops~\citep{hop} or leverage sparse adjacency matrices and sparse matrix multiplications~\citep{SGCN}. However, physical adjacency does not necessarily imply social influence. Attention-based aggregation makes this selectivity explicit by learning edge weights: GATraj~\citep{GATraj} and SocialVAE~\citep{SocialVAE} dynamically emphasize the most relevant neighbors. Transformer-based graph variants extend this toward more global interaction reasoning as in MHTraj~\citep{MHTraj:} and STAR~\citep{STAR}, often with higher computational and memory cost due to (near-) global attention over agents (and sometimes time). Hybrid approaches keep GCN-style message passing but modulate it with learned interaction weights (attention/gating), so that neighbor contributions are reweighted without resorting to fully global self-attention as in AVGCN~\citep{AVGCN} and  DSTIGCN~\citep{DSTIGCN}.}

\paragraph{\modified{Handling time: sequential pipelines vs spatio-temporal graphs.}}
\modified{A parallel axis concerns how time is introduced. Many pipelines historically alternate spatial graph operations with temporal encoders (e.g., LSTMs or TCNs), which can work well but may create bottlenecks or lose information at the interface. 
Recent methods increasingly favor fusing spatial and temporal relations into a single embedding, as in STG-KNet~\citep{STG-KNet} and UniEdge~\citep{UniEdge}. To improve robustness across different scenes, another emerging trend is to embed geometric equivariance or invariances directly into the model so that predictions respect the geometric symmetries of human movement. Along this direction, EqMotion~\citep{Eqmotion} decouples trajectories into equivariant geometric features for motion and invariant reasoning components for stable agent interactions, while STFlow~\citep{STFlow} employs a geometric flow‑matching framework to denoise trajectories from conditioned random walks.
Therefore, predictions remain consistent under rotation and translation, ensuring robust trajectory forecasting regardless of the scene's global coordinate system. 
Finally, some approaches revisit temporal modeling through the spectral lens, where wavelet/Fourier representations can make long-range structure and periodicities more accessible to graph operations as in MSWTE-GNN~\citep{MSWTE-GNN} 
EigenTrajectory~\citep{EigenTrajectory}.} 

\paragraph{\modified{A lightweight alternative: implicit interaction embeddings.}}
\modified{While explicit graphs offer 
interpretable structure (edges, weights, attention), some methods void committing to a topology and learn latent interaction features instead. SGANv2~\citep{SGANv2} and Social-Implicit~\citep{Social-Implicit} embed social context with MLPs/CNNs encoders, which can reduce the quadratic complexity of dense graphs. The flip side is that the learned interactions may be less transparent than the explicit edges and attention weights used
in GAT-based models.}

\subsubsection{Topological and structural priors in multi-agent trajectories.}
Recent works use topological priors to encode global interaction patterns, providing structured alternatives to purely data-driven methods. For instance,~\cite{MTP} propose Multiple Topologies Prediction to group trajectories into topological modes. While this effectively reduces multimodal ambiguity by predicting representative paths, its reliance on structured intersections limits its generalization to open spaces. More recently, \modified{social homology~\citep{SHINE} has been used to classify paths into homology classes that align with human behaviors, constraining robot motion accordingly}. This offers a rigorous framework for social compliance, though computing these classes can be computationally demanding and yield overly conservative behavior in dense crowds. Finally, SRefiner~\citep{SRefiner} uses “soft-braid attention” to capture fine-grained spatio-temporal relationships. By explicitly modeling how trajectories intertwine, it excels at reducing collisions, but introduces substantial computational overhead that limits high-frequency, real-time inference.

\begin{figure*}[t]
\centering
    \includegraphics[width=0.8\textwidth]{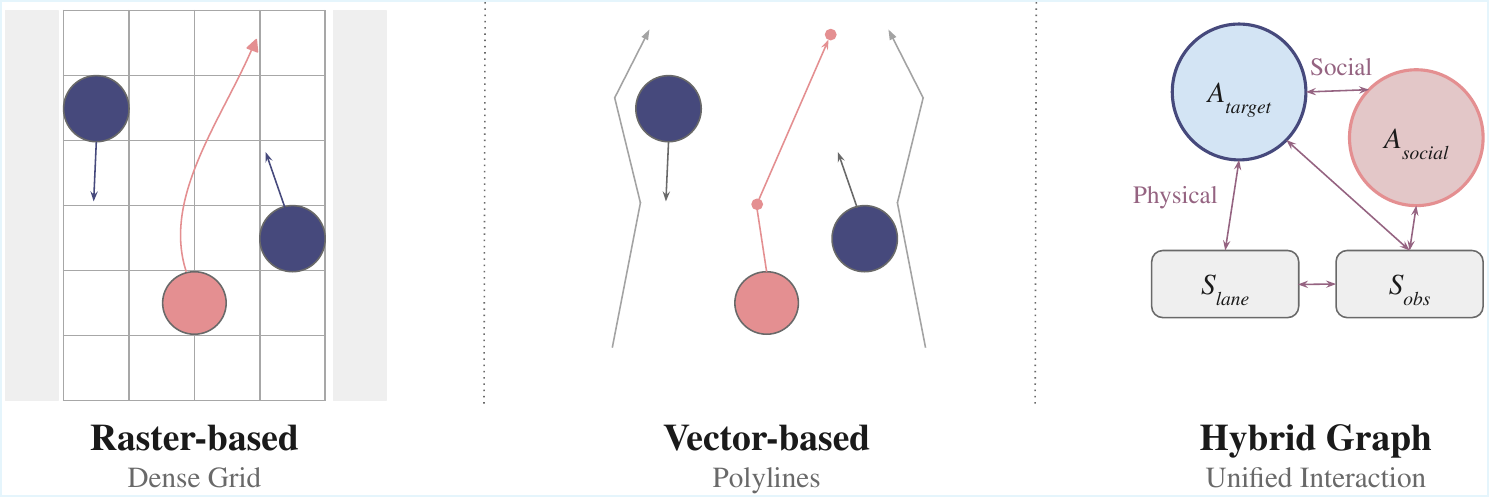}
    \caption{\modified{Comparison of contextual encoding paradigms for trajectory prediction: (a) Raster-based encoding capturing dense spatial and social grids, (b) Vector-based representations using discrete polylines for agents and map boundaries, and (c) Hybrid Graphs integrating both dynamic agents ($A$) and static scene elements ($S$) into a unified message-passing framework.}}
    \label{fig:context_encoding}
\end{figure*}

\subsection{Dynamic and static contextual information} \label{subsubsec:dyn+stat}

\modified{Effective trajectory prediction requires integrating both the \emph{dynamic} social context and the \emph{static} environmental context. The core challenge lies in fusing these heterogeneous data sources into a coherent representation that preserves both spatial geometry and social dependencies.}

\modified{A fundamental design choice is how the environment is encoded as outlined in Figure~\ref{fig:context_encoding}. 
Many approaches rely on \emph{raster-based} representations, which provide dense spatial coverage and integrate naturally with CNN or Vision Transformer (ViT) backbones for processing semantic maps or satellite imagery.
For instance, SceneAware~\citep{SceneAware} and Meta-IRLSOT++~\citep{Meta-IRLSOT++} use ViTs and TopFormer encoders to extract visual features, while CSCNet~\citep{CSCNet} and STHGLU~\citep{STHGLU} rely on CNNs to process occupancy grids and heatmaps. While these methods excel at capturing unstructured traversable areas, they are computationally heavy and lack explicit topological structure. Conversely, \emph{vector-based} methods encode scene elements (e.g., lanes, obstacles) as graph nodes. As an example, GAIN~\citep{GAIN} introduces a Road Geometry Graph to explicitly model traffic rules. This vectorization, also seen in Hypertron~\citep{Hypertron}, offers a more structured and lightweight alternative to pixel-level encoding, though it requires higher-level semantic pre-processing.}

\modified{Once environmental features are extracted, the fusion mechanism determines how the model reasons about agent-agent and agent-scene interactions. Early architectures like NMMP~\citep{NMMP} and LSSTA~\citep{LSSTA} employed \emph{late fusion}, processing social and scene features in parallel streams (e.g., via LSTMs and CNNs) and concatenating them before decoding. While simple, this approach often fails to capture the intricate, spatially-grounded dependencies between an agent and specific map features. To address this, recent models have shifted toward \emph{interaction-centric fusion} using cross-attention mechanisms. For example, AMD~\citep{AMD} and CINet~\citep{CINet} dynamically weigh the relevance of specific map regions to the target agent, rather than compressing the entire map into a static vector. Similarly, SPU-BERT~\citep{SPU-BERT} and Snapshot~\citep{uhlemann2025snapshot} leverage Transformer architectures to unify spatial and temporal tokens within a shared embedding space. Another powerful paradigm is the \emph{Hybrid Graph}. Instead of using separate encoders, models like STGT~\citep{STGT} design unified graphs where both agents and scene elements act as interacting nodes. This allows message passing to propagate physical constraints (e.g., ``wall implies repulsion'') directly through the graph edges, although scaling such heterogeneous networks to dense environments remains computationally demanding.}

\modified{Beyond learned representations, ensuring physical plausibility remains a hurdle. Several models enforce this explicitly through \emph{energy maps} and potential fields. SocialCVAE~\citep{SocialCVAE} and SSA-GAN~\citep{SSA-GAN} generate energy landscapes where obstacles create high-cost regions, guiding predictions away from collisions, while ARP-STGCN~\citep{ARP-STGCN} formalizes this as an attraction-repulsion potential. However, correlation does not imply causation: a pedestrian stops \emph{because} of a red light, not merely because of proximity to it. Recognizing this limitation, CICR~\citep{CICR} moves beyond standard attention by applying causal inference and counterfactual reasoning to disentangle social and scene cues. Finally, an emerging trend explores semantic reasoning via Large Language Models (LLMs); for instance, LG-Traj~\citep{LGTraj} uses LLMs to interpret scene contexts conceptually rather than purely geometrically.}

\modified{Taken together, the above approaches highlight that effective HTP hinges not only on extracting social and environmental cues, but on representing them in a form that downstream architectures can reason over reliably. Having established how these contextual signals are encoded and fused, it becomes clear that capturing their dynamics relies heavily on the expressiveness of the underlying backbone. The following section will therefore focus on the deep learning architectures used to extract the spatio-temporal patterns governing pedestrian movement from the complex contextual inputs.}

\section{Backbone architecture}
\label{subsection:archi}
 
The HTP field has progressed significantly in tandem with advancements in deep learning. It has evolved from early recurrent encoder–decoder architectures~\citep{rnn, Hochreiter_1997, GRU}, through the emergence of generative models~\citep{Goodfellow_2014}, to today’s transformer-based systems and large language models~\citep{Transformer2017, gpt}. 
The variety of deep learning modules used in HTP and the timeline of their use \modified{are depicted in} Figure~\ref{fig:histogram}\modified{,} showing that HTP follows closely the deep learning advances.
In this section, we describe how existing deep learning modules are integrated into the HTP models’ \emph{backbone architecture}, meaning the intermediate layers responsible for processing information, excluding the input and output parts.
The key modules of interest are sequential and generative modules, presented in the following two subsections. 

\begin{figure}[ht]
     \centering
     \includegraphics[width=0.35\textwidth]{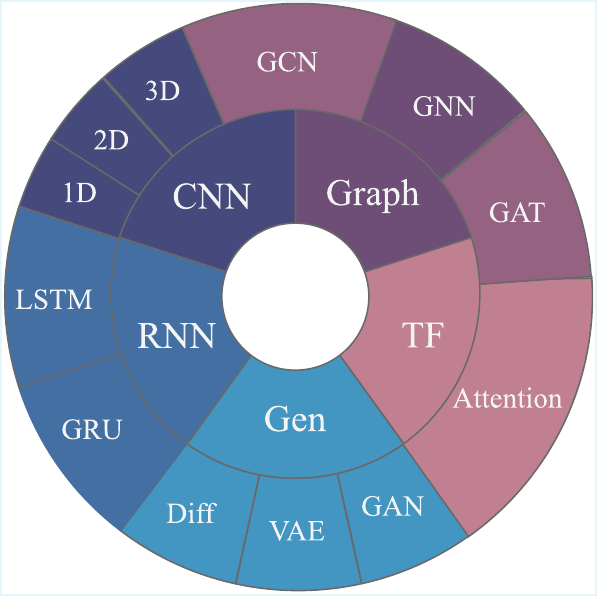}
     \caption{\footnotesize Deep-learning modules used in HTP. This figure takes inspiration from Fig.~11 in~\citep{Fu_2024}. It represents the different kind of modules used in HTP algorithms. Notice that GCN overlaps between CNN and Graph section and GAT overlaps between Graph and Transformers. ``Gen'' stands for Generative module; ``Attention'' for Attention mechanism module; ``TF'' for Transformers; ``Diff'' for Diffusion.}
     \label{fig:modules_rate}
 \end{figure}
\subsection{Sequence modeling}
\label{subsec:sequence}

\modified{Sequence modeling is a cornerstone of HTP, providing tools to handle the temporal dependencies inherent to pedestrian motion. Currently, the field is dominated by three main paradigms: Recurrent Neural Networks (RNNs), Transformers, and, more recently, Selective State Space Models (SSMs).}


\modified{Recurrent Networks have historically been the standard for capturing temporal dependencies, making them highly effective for short-term motion prediction (see Figure~\ref{fig:histogram}). However, their sequential processing nature inherently bottlenecks the modeling of long-range dependencies and complex spatial interactions in crowded scenes. Transformers address these limitations by leveraging self-attention mechanisms, which enable parallel processing and direct relational reasoning over extended time horizons. Finally, SSMs offer a newly emerging alternative, combining the long-horizon modeling capabilities of Transformers with the linear-time complexity of RNNs. The following paragraphs explore how these three module families have been adapted for HTP.}

\subsubsection{Recurrent Networks.}

\modified{Recurrent networks rely on sequential memory mechanisms to embed past motion context. While vanilla RNNs~\citep{rnn} established the foundation, Long Short-Term Memory (LSTM) networks~\citep{Hochreiter_1997} and Gated Recurrent Units (GRUs)~\citep{GRU} have become the de facto standards due to their ability to mitigate the vanishing gradient problem. The choice between LSTMs and GRUs is typically empirical, trading off between parameter count and convergence speed.}

\paragraph{Encoding and Decoding with LSTMs and GRUs.}
\modified{During the \emph{encoding phase}, LSTMs and GRUs embed agents' motion histories into compact latent representations~\citep{CIDNN, DERGCN, kothari2021, TrajPRedZhou_2024, SocialCVAE, Group-Obstacle-LSTM, SR-LSTM, ganeshaaraj2025enhancing, CMPT}. To enrich this representation, several architectures incorporate attention mechanisms over LSTM states, such as Fuzzy Query Attention~\citep{FQA,X13,X16}, or utilize bidirectional LSTMs to capture both forward and backward temporal contexts, a common feature in generative frameworks like BCDiff~\cite{BCDiff} and MG-GAN~\citep{MG-GAN}. Similarly, GRU encoders often feature dynamic memory regulation, such as adaptive forgetting controllers that modulate memory retention based on motion dynamics~\citep{SMEMO, AFC-RNN}, or work in tandem with Transformer modules to capture heterogeneous dependencies across different agent types~\citep{AMD}.}

\modified{In the \emph{decoding phase}, RNNs are tasked with unrolling the latent state into future coordinate sequences. ConvLSTMs are favored when decoding requires maintaining explicit spatio-temporal grid structures~\citep{FLEAM, SSP}, whereas standard LSTMs are prevalent in generative setups where they combine trajectory, scene, and social characteristics with latent noise~\citep{GATraj, MG-GAN, SSA-GAN}. GRU-based decoders offer similar capabilities~\citep{SocialVAE, GroupNet, MATRIX}, with some models employing forward/backward GRUs~\citep{SRGAT, SocialVAE} or unique bidirectional modules~\citep{BRAINet} to better capture the multimodality of future paths. Gaussian Mixture Model (GMM) heads are also frequently appended to GRU decoders to adapt trajectory generation to both common and anomalous behaviors~\cite{ganeshaaraj2025enhancing}.}

\modified{Despite their widespread use, the sequential bottleneck of RNNs naturally limits their capacity to model very long-range interactions, prompting the field's shift toward attention-based architectures.  Specifically, as prediction horizons extend, the recursive nature of these networks often leads to information dilution, making it difficult to retain critical early behavioral cues. By bypassing this sequential dependency, attention-based models allow for parallel processing and direct connections across all time steps, providing a more robust framework for forecasting complex, prolonged pedestrian dynamics as discussed below. }

\subsubsection{Transformers.}

\modified{Originating in Natural Language Processing~\citep{Transformer2017}, the Transformer architecture has fundamentally reshaped HTP. By replacing recurrence with self- and cross-attention mechanisms, Transformers enable parallel processing and direct modeling of the complex, long-range spatiotemporal patterns observed in crowded scenes and multimodal data integration.}

\paragraph{Transformers as holistic frameworks.}
\modified{Transformers are frequently deployed as holistic, end-to-end forecasting frameworks. They naturally bridge spatial and temporal dimensions, either through joint modeling of multi-agent dynamics~\citep{ForceFormer}, by unifying trajectory and semantic map tokens within a single stream~\citep{SPU-BERT}, or by employing explicit spatial and temporal attention heads~\citep{VIKT, GDDDL}. The architecture's flexibility allows for advanced structural designs, such as masking for partially observed sequences~\citep{X6}, dual-stream networks separating coarse trajectory and fine context encoding~\citep{X26}, or unified architectures that merge distinct predictive modules~\citep{TUTR, MART}. Recent works have also explored the spectral domain, combining frequency-domain convolutions and temporal attention to explicitly capture periodic motion dynamics~\citep{MHTraj:}, or decomposing motion into hierarchical kinematic streams (position, velocity, acceleration) processed by dedicated attention modules~\cite{huang2025learning}. Finally, Transformers can be coupled with discrete associative memories to perform autoregressive reasoning over quantized motion tokens~\citep{FMTP}, or integrated with vision-based scene features to enforce physical plausibility~\citep{SceneAware}.} 
\modified{Taken together, the above architectures illustrate how researchers use the Transformer’s flexible tokenization interface to impose structure on spatiotemporal data, whether through unified streams, factorized motion representations, or auxiliary memory and perception modules. 
However, this flexibility also makes holistic designs highly sensitive to tokenization choices: poorly structured tokens can obscure geometric relationships or inflate sequence length, and the absence of strong inductive biases places greater burden on data quality and model regularization.
}

\paragraph{The critical role of attention mechanisms.}
\modified{The core strength of Transformers in HTP lies in their versatile \emph{attention mechanisms}, which are heavily engineered to parse social dynamics. Self-attention is predominantly used to integrate an agent's own multimodal history, such as
pose and trajectory~\citep{HST}, while cross-attention allows to identify relevant social influences from surrounding agents or environmental cues~\citep{Social-SSL, GA-STT, AFC-RNN, CINet, ASTRA}. To capture nuanced social hierarchies, models like SocialMOIF~\citep{SocialMOIF} implement multi-order attention to distinguish direct from indirect neighbor influences. Other approaches refine social reasoning via dual-attention designs that explicitly filter relevant neighbors before integrating their cues~\citep{SocialTrans, FTPN}. Ultimately, these attention maps collectively highlight how attention serves not only as a feature aggregator but also as an interpretable interface for modeling collision avoidance and social coordination~\citep{X14}. However, despite the transformative impact of self- and cross- attention mechanisms, their quadratic complexity poses a clear bottleneck, limiting their efficiency for long-range dependencies.}

\paragraph{Enhancing Generative Models and Graph Networks.}
\modified{Transformers serve as powerful backbones for generative frameworks, effectively replacing RNNs to improve sequence realism and diversity. In CVAE settings, they
encode past motions before latent sampling~\citep{su2025improving, LADM}, or support dual CVAE-VAE structures that enforce social coherence through masked sequence reconstruction~\cite{damirchi2025socially}. In adversarial settings, Transformers act as goal-conditioned generators~\citep{PPNet, AgentFormer} or sophisticated discriminators that evaluate multi‑agent consistency~\citep{SGANv2}. 
More recently, they have become critical components of diffusion models, managing the temporal coherence of the denoising process~\citep{gu2022MID, Leapfrog, BCDiff} or acting as lightweight context encoders that guide diffusion generators based on past trajectories and map priors~\citep{LCD, shi2025leveraging}. A parallel line of work integrates graph structure directly into the attention mechanism. These \emph{Graph-enhanced Transformers} apply attention directly across graph nodes, merging the topological priors of GCNs with the dynamic weighting of Transformers to capture evolving physical and social constraints~\citep{GAIN, STGT, GTP-Force, LSSTA, MSRL, PTP-STGCN}. However, despite offering strong structural priors, graph‑enhanced variants may overconstrain interactions when topology is uncertain or rapidly changing.}

\modified{In sum, Transformers offer unmatched flexibility for unifying spatial, temporal, and social cues, but this expressiveness comes with trade‑offs. The quadratic cost of attention limits scalability in dense scenes, and their performance depends heavily on how trajectories, agents, and graph nodes are tokenized and encoded. As a result, Transformer‑based HTP models excel when data is abundant and interactions are well‑structured, but they require careful design to remain robust in crowded, noisy, or resource‑constrained settings.
}


\subsubsection{Selective State Space Models (SSMs).}

\modified{Recently, Selective State Space Models, notably Mamba-based architectures~\citep{mamba}, have emerged as highly efficient alternatives. They capture long-range temporal dependencies via learnable continuous-time dynamics while maintaining linear computational complexity, thus avoiding the quadratic scaling bottleneck of self-attention. Early applications in HTP show significant promises: architectures like SAMD and TOTP~\citep{ren2025stochastic,ren2025totp} use Motion-Selective SSMs to balance trajectory stability and multimodal diversity within diffusion processes. Similarly, models like MambaPTP~\citep{MambaPTP} integrate Mamba encoders with graph attention or physical priors, using bidirectional gated state-space modeling to filter redundant interactions and enhance long-horizon temporal reasoning. Despite these advantages, SSM‑based predictors remain in an early stage for HTP, with their ability to fully replace attention‑based architectures in dense, socially complex scenes still an open question. Current SSM formulations can underfit fine‑grained social cues, their inductive biases are less expressive than attention for multi‑agent interactions, and their stability depends heavily on careful gating and normalization.}

\modified{Because pedestrian behavior is inherently non-deterministic, sequence modeling alone is often insufficient; predicting a single average path fails to capture the true distribution of plausible futures. Consequently, much of the recent literature relies on the sequence modeling architectures detailed above as foundational backbones for \emph{generative modeling}. The next section will delve into how these generative frameworks map deterministic histories into multimodal future distributions.}

\subsection{Generative modeling}
\label{subsec:generative}
\begin{figure*}[t]
    \centering
    \includegraphics[width=0.8\linewidth]{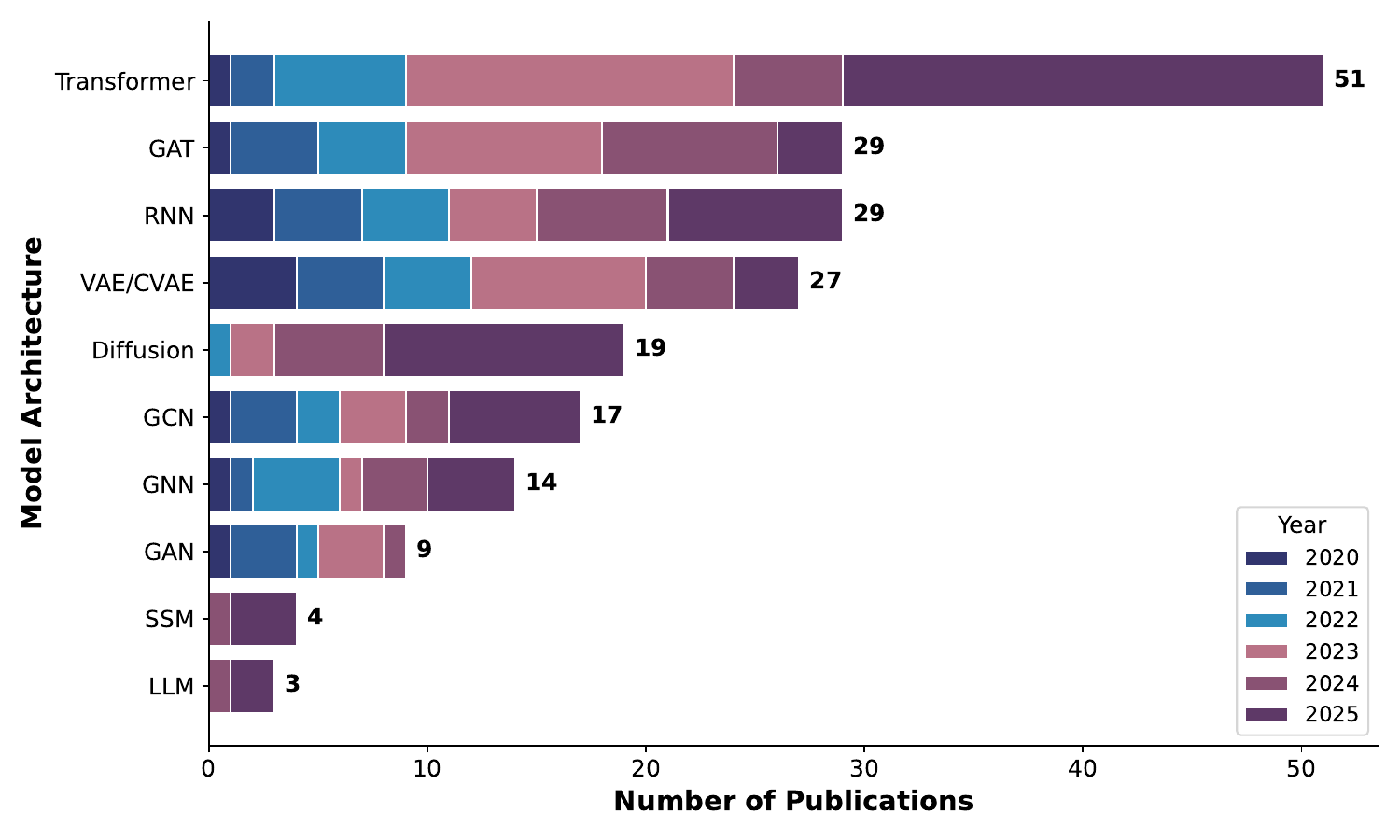}
    \caption{Evolution of the yearly number of publications categorized by the deep learning architectures and modules utilized from 2020 to 2025. The data presented in this chart is exclusively derived from the articles selected through our systematic literature review using the PRISMA methodology. 
    }
    \label{fig:histogram}
\end{figure*}

\modified{To address the non-deterministic nature of pedestrian behavior, generative modeling has become central to HTP. These models provide the necessary mathematical frameworks to quantify uncertainty and output diverse, multimodal distributions of plausible future paths. Among the most notable techniques are Generative Adversarial Networks (GANs)~\citep{Goodfellow_2014}, Variational Auto-Encoders (VAEs)~\citep{vae}, and, more recently, Diffusion models~\citep{IDM} and Large Language Models (LLMs)~\citep{brown2020}. This section explores how these generative frameworks have been adapted for HTP and the main innovations they introduce (see Figure~\ref{fig:histogram}).}


\subsubsection{GANs.}

\modified{Generative Adversarial Networks~\citep{Goodfellow_2014} were among the first successful generative models adapted for trajectory forecasting. They rely on an adversarial zero-sum game: a generator maps random latent noise to trajectory coordinates, while a discriminator attempts to distinguish these generated paths from ground-truth data.}
\modified{In HTP, GANs are primarily explicitly tailored to enforce social and physical compliance. SocialGAN~\citep{Gupta_2018} pioneered this space by training a generator to output plausible paths that are evaluated by a discriminator, effectively learning a distribution aligned with social norms. To capture a wider array of pedestrian multimodality and mitigate mode collapse, MG-GAN~\citep{MG-GAN} extends this setup by deploying multiple generators and dynamically selecting among them via a dedicated module. }

\modified{To better ground the adversarial generation in reality, modern GAN architectures frequently integrate contextual encoders. BR-GAN~\citep{BR-GAN} and SSA-GAN~\citep{SSA-GAN} combine LSTM decoding with geographical and social features for nuanced intention modeling, while DTGAN~\citep{DTGAN} uses Graph Attention Networks to explicitly encode pedestrian interactions. Ensuring that generated trajectories are both diverse and physically safe remains a priority: models like SGANv2~\citep{SGANv2}, STSF-Net~\citep{STSF-Net}, and TPNMS~\citep{TPNMS} incorporate safety-aware constraints and multi-supervision to enhance prediction quality. Further innovations include reciprocal learning mechanisms~\citep{X13} and embedded social norms constraints~\citep{X5} to guarantee that the generated paths respect group cohesion and collision avoidance.}

\subsubsection{(C)VAEs.}

\modified{Variational Auto-Encoders~\citep{vae}, and their Conditional variants (CVAEs), offer a more statistically stable alternative to GANs. They learn a stochastic mapping from data to a latent space (via an encoder) and a reverse mapping to reconstruct the data (via a decoder). In inference, future trajectories are generated by sampling from a learned prior distribution conditioned on past observations.}

\paragraph{Goal-driven and Endpoint-conditioned Generation.}
\modified{A prominent trend in CVAE-based forecasting is conditioning the latent space on spatial targets. Models such as PECNet~\citep{PECNet}, CTSGI~\citep{CTSGI}, and MINet~\citep{MINet} explicitly sample short-term goals or endpoints first, and then condition the trajectory generation on these spatial anchors, significantly improving the physical plausibility of the sampled paths.} 
\modified{However, this strategy implicitly assumes that intermediate waypoints or final destinations can be reliably inferred at test time. In practice, goal prediction is itself a hard problem, and errors in the sampled endpoints can propagate through the generative process, leading to over‑confident or systematically biased forecasts. Moreover, endpoint‑conditioned models may underrepresent behaviors that do not follow clear goal‑directed patterns (e.g., reactive avoidance), limiting their expressiveness in settings where intent is ambiguous or multimodal.}

\paragraph{Social and Graph-enhanced Latent Spaces.}
\modified{To handle complex multi-agent scenarios, CVAEs are frequently coupled with graph or hypergraph networks. ForceFormer~\citep{ForceFormer} and AgentFormer~\citep{AgentFormer} model individual and collective actions through CVAEs to represent social-temporal dependencies. Similarly, Hypertron~\citep{Hypertron}, and Tri-HGNN~\citep{Tri-HGNN} integrate hypergraph reasoning into the latent space to capture higher-order group interactions, while DynGroupNet~\citep{DynGroupNet} and Grouptron use specialized latent variables to forecast dense crowds and group-aware stochastic behaviors. NMRF~\citep{NMRF} further formalizes this by implementing unary and pairwise interaction potentials within a neuralized Markov Random Field framework, sampling latent variables to generate interaction-aware futures.} 
\modified{These relational latent spaces offer high expressiveness, but they come with scalability and robustness trade‑offs: the number of edges or hyperedges grows quickly with crowd size, leading to heavy computation unless aggressively pruned. Performance can also degrade when perception noise corrupts the underlying interaction graph, since the latent structure is tightly coupled to agent detection and neighborhood selection. As a result, these models excel in structured, moderately sized scenes but can struggle to maintain efficiency and stability in large, cluttered, or noisy environments.}

\paragraph{Temporal Dynamics and Uncertainty.}
\modified{Modeling the temporal evolution of uncertainty is another key focus. SocialVAE~\citep{SocialVAE} introduces timewise latent variables and backward RNN structures to capture non-linear multi-agent dynamics, while SocialCVAE~\citep{SocialCVAE} isolates a specific latent variable to represent ``socially reasonable randomness.'' Hierarchical latent structures, such as those in Trajectron++~\citep{Trajectron++}, ASTRA~\citep{ASTRA}, and M2P3~\citep{M2P3}, allow models to separate high-level intent from low-level motion dynamics. A wide array of other architectures—including DROGON-E~\citep{DROGON}, LSSTA~\citep{LSSTA}, MSRL~\citep{MSRL}, SBD~\citep{SBD}, ScePT~\citep{ScePT}, Sparse-GAMP~\citep{Sparse-GAMPS}, SPU-BERT~\citep{SPU-BERT}, SSALVM~\citep{SSALVM}, STGlow~\citep{STGlow}, UEN~\citep{UEN}, and works by~\citet{X16} and~\citet{X2}—demonstrate the versatility of CVAEs in fusing recurrent, graph, and environmental features to produce diverse motion patterns.}
\modified{Despite this diversity, latent dimensions often lack interpretability, and hierarchical structures may not scale gracefully in dense multi‑agent scenes. As a result, uncertainty‑aware generative models must balance expressiveness with tractability, ensuring that latent variables remain both semantically meaningful and computationally manageable. }

\subsubsection{Diffusion Models and Normalizing Flows.}

\modified{While originally developed for high-fidelity image synthesis, Diffusion Models have rapidly overtaken the HTP landscape~\citep{X8}. They operate by gradually corrupting ground-truth trajectories with Gaussian noise in a forward process, and then training a neural network to estimate the reverse-time dynamics (denoising), mapping pure noise back to coherent future paths. They excel at capturing complex, multimodal distributions without suffering from mode collapse.}

\paragraph{Conditioning and Guidance.}
\modified{To ensure that the denoised trajectories are socially and contextually accurate, diffusion processes in HTP are heavily conditioned. MID~\citep{gu2022MID} pioneered this by refining noisy inputs into multiple plausible futures based on motion history. IDM~\citep{IDM} separates the sources of uncertainty by applying two distinct diffusion processes: one for the endpoint and another for the trajectory conditioned on it. SA-SGM~\citep{SASGM} takes a continuous-time score-based approach, complementing the diffusion process with a learned social attention module.}

\paragraph{Efficiency and Latent Diffusion.}
\modified{Because standard diffusion requires hundreds of iterative denoising steps, computational efficiency is a major hurdle for real-time HTP. LADM~\citep{LADM} and~\citet{hu2025cvae} tackle this by performing the diffusion process within a lower-dimensional VAE latent space rather than the raw coordinate space. CDDM~\citep{CDDM} accelerates inference through a collaborative progressive distillation framework that transfers knowledge from a large teacher model to a lightweight student via DDIM-based supervision. EPD~\citep{EPD} refines initial trajectory distributions generated by an energy-based model using very few denoising steps.}
\modified{Despite these advantages, latent space diffusion depends heavily on the quality of the underlying VAE, and compression can obscure fine‑grained motion cues or introduce reconstruction bias. Distillation‑based and few‑step accelerators offer additional speedups but may reduce multimodality or rely strongly on high‑quality initial proposals. As a result, efficiency‑oriented variants provide meaningful speedups over vanilla diffusion but must balance computational savings against potential losses in fidelity, diversity, or robustness.}

\paragraph{Bidirectional Processing and Refinement.}
\modified{Several works leverage diffusion for bidirectional reconstruction and iterative refinement. BCDiff~\citep{BCDiff} and Diffusion²~\citep{Diffusion2} employ sequential or bidirectional setups to first reconstruct missing past observations before predicting the future. Instead of starting from pure noise, Diff-Refiner~\citep{DiffRefiner} initializes the diffusion process from coarse baseline predictions, acting as a plug-and-play ODE-based sampler to enhance multi-agent accuracy. Leapfrog~\citep{Leapfrog} further optimizes the forward and reverse phases to better align with specific human kinematic patterns, while SAMD~\citep{ren2025stochastic} integrates Mamba-based Selective State Space Modules to improve denoising stability and long-range temporal consistency.} 
\modified{While these refinement strategies offer strong corrective capabilities, bidirectional setups can be computationally heavier. Moreover, the overall pipeline remains sensitive to initialization quality and less reliable in settings with noisy or incomplete past observations.}

\paragraph{Normalizing Flows. }
\modified{Finally, \emph{Normalizing Flows} have also gained traction alongside diffusion. Offering invertible and deterministic generative mappings, frameworks like STFlow~\citep{STFlow} and MoFlow~\citep{MoFlow} provide exact log-likelihood estimation to efficiently capture multimodal and geometrically consistent pedestrian motion.}
\modified{Flows offer clear advantages including fast sampling, exact likelihoods, and stable training, but they remain less expressive than diffusion when modeling highly multimodal or long‑tail behaviors. Their invertibility constraints can limit architectural flexibility, and capturing complex social interactions often requires carefully designed coupling layers or graph‑structured transformations. As a result, flows are appealing for real‑time or resource‑constrained settings, but diffusion remains more dominant when modeling rich, high‑entropy futures.}

\modified{
\subsubsection{Large Language Models for HTP.}
A fast‑emerging direction extends generative modeling beyond geometric cues by incorporating \emph{Large Language Models} (LLMs) as semantic priors. Rather than operating directly on trajectory tokens, LLM‑based approaches recast forecasting as a prompt‑driven numerical reasoning task~\citep{LMTraj} or use language and vision-language models (VLMs) to extract high‑level scene semantics and motion prototypes~\citep{SceneAware, LGTraj}. These methods inject commonsense knowledge about intent, affordances, and social conventions into the generative pipeline, helping bridge the gap between low‑level geometric tracking and higher‑level intention understanding. 
While fully multimodal language architectures for HTP have not yet emerged, coupling LLM/VLM 
reasoning with trajectory predictors suggests  a promising direction for integrating richer semantic cues into generative models. At the same time, incorporating LLMs or VLMs into HTP remains challenging: their computational footprint limits real‑time deployment, and their weak geometric grounding makes them complementary to, rather than replacements for, established forecasting backbones.
As such, LLMs function less as sequence models and more as semantic conditioning modules that augment diffusion, CVAE, or autoregressive predictors with structured priors unavailable from trajectory data alone.
} 

\section{Trajectory Prediction and Evaluation}\label{subsection:output}
Beyond the choice of model, how predictions are generated and how performance is measured can significantly influence computational efficiency and accuracy. 

Section~\ref{subsubsec:pred_strat} reviews the four primary prediction strategies used in HTP: \emph{Direct} prediction, \emph{Auto-regressive} prediction, \emph{Goal-conditioned} prediction, and \emph{Prediction Refinement}. Stochasticity modeling is discussed in Section~\ref{subsubsec:stochastic}. The marginal versus joint prediction opposition is addressed in Section~\ref{subsubsec:marginal}. Finally, evaluation metrics are presented in Section~\ref{subsubsec:metrics}.

\subsection{Prediction strategies} \label{subsubsec:pred_strat}

\begin{figure}[t]
    \centering
    \includegraphics[width=0.475\textwidth]{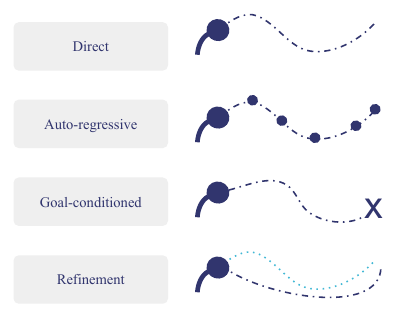}
    \caption{\footnotesize Illustration of the four trajectory-prediction strategies from top to bottom: (1) Direct, which generates the entire future trajectory in a single forward pass; (2) Auto-regressive, which predicts future positions step-by-step by conditioning on past observations and previously generated outputs; (3) Goal-conditioned, where predictions are guided by an explicit or inferred destination; and (4) Refinement, in which an initial coarse trajectory is improved through learned residual corrections.}
    \label{fig:pred}
\end{figure}

We have identified four non-exclusive paradigms used to generate the future pedestrian trajectories, as illustrated in Figure~\ref{fig:pred}.

\textit{Direct prediction} methods predict all future positions in a single forward pass, without any recurrence: given observed states $\mathbf{S}_{1:T_{obs}}$, outputs the full future trajectory $\hat{\mathbf{Y}}_{T_{obs + 1}:T_{pred}} = f(\mathbf{S}_{1:T_{obs}})$ at once.
For example, PTP-STGCN~\citep{PTP-STGCN} relies on 2D convolutional layers for producing bivariate Gaussian distributions from which the predicted trajectories are deduced. Similarly,  IMP~\citep{IMP} uses a MLP to decode features into future trajectories.

\textit{Auto-regressive prediction}, in contrast, generates future positions sequentially, conditioning each predicted step on past observations and previously predicted positions. It can be expressed as $\hat{\mathbf{y}}_t = f(\mathbf{S}_{1:T_{obs}}, \hat{\mathbf{Y}}_{T_{obs + 1}:t-1})$, i.e. the prediction at time $t$ depends on prior outputs.
This paradigm is primarily followed by models that incorporate RNNs, as well as some diffusion models. For example, SBD~\citep{SBD} employs a GRU followed by an MLP to forecast each timestep of the predicted trajectory. 

\textit{Goal-conditioned prediction} extends the prediction process by incorporating explicit or estimated goal information, such as destination points. The predictive model thus becomes conditioned on a latent or observed goal variable $\mathbf{g}$: $\hat{\mathbf{Y}}_{T_{obs + 1}:T_{pred}} = f(\mathbf{S}_{1:T_{obs}}, \mathbf{g})$.
For example, \mbox{TrajPRed~\citep{TrajPRedZhou_2024}} employs a multi-goal estimation module to determine the goal using a CVAE before generating the trajectory itself, \emph{conditioned} on the predicted goal.

\textit{Prediction refinement} methods improve a coarse trajectory by applying learned corrections. Typically, a base model generates a preliminary trajectory $\hat{\mathbf{Y}}$, and a refinement module, $\mathcal{R}$, updates it through a residual:
    $\hat{\mathbf{Y}}_{1:T_{pred}} \leftarrow \hat{\mathbf{Y}}_{1:T_{pred}} + \mathcal{R}\big(\mathbf{S}_{1:T_{obs}}, \hat{\mathbf{Y}}_{1:T_{pred}}, \mathbf{C}\big)$
where $\mathbf{C}$ optionally encodes scene or social context.

For example, DynGroupNet~\citep{DynGroupNet} introduces a prediction refinement module that enhances future trajectory predictions using MLPs. In contrast, SAnchor~\citep{SAnchor} predicts trajectories by separating human motion into broad discrete intents and more specific scene-related adjustments. AC-VRNN~\citep{AC-VRNN} refines its predictions at each time step by utilizing a graph-based attention mechanism that directly modifies the recurrent hidden state to integrate human interactions, and further incorporates refinement through a Kullback-Leibler divergence loss applied to belief maps. Meanwhile, SBD~\citep{SBD} uses an error compensation network at each time step to refine the trajectory. \cite{IAN} develop the Interactive Adjustment Network, an external module that uses individual feedback from past prediction errors to refine current trajectory predictions.


\modified{Taken together, these four prediction strategies illustrate the variety of mechanisms through which HTP models generate future trajectories, each coming with distinct operational trade-offs. Direct prediction methods are computationally efficient as they generate the entire future trajectory in a single forward pass, but they may struggle to dynamically adjust to compounding temporal changes. Auto-regressive prediction mitigates this by generating future positions sequentially and conditioning each step on previously predicted positions, though this sequential nature can lead to slower inference times and the accumulation of errors over longer horizons. Goal-conditioned prediction offers highly structured and interpretable paths by incorporating explicit or estimated goal information, yet its reliability is heavily bottlenecked by the accuracy of the initial goal estimation module itself. Finally, prediction refinement can significantly enhance accuracy by applying learned corrections to a preliminary trajectory, but this inherently introduces additional computational overhead. Ultimately, recognizing this taxonomy helps contextualize the modeling choices adopted in the literature, as researchers must continuously balance temporal structure, behavioral assumptions, and computational efficiency against the specific requirements of real-world applications.}

\subsection{Stochastic trajectory prediction} \label{subsubsec:stochastic}

Most HTP models embrace a stochastic approach rather than a deterministic one because human motion is fundamentally uncertain and shaped by a variety of factors. Deterministic models offer a single predicted trajectory, which often fails to capture the natural variability seen in human behavior and may even fail catastrophically by producing ``averaged'' outputs when faced with inherently multimodal situations. For instance, pedestrians in similar situations may take different routes due to subtle variations in their intents or the context in which they find themselves. Stochastic methods sample multiple potential trajectories from the estimated predictive distributions, better reflecting the human trajectories variability. 
Stochasticity arises not only from agents' motion but also from uncertainties in their goals. 
In \emph{goal-conditioned} model, sampling a goal induces a distribution over future trajectories conditioned on that goal. \emph{Generative models} explicitly handle this uncertainty either through latent variables, as in VAEs and GANs, or by introducing noise within a learned iterative denoising process (e.g., diffusion models), enabling multiple plausible futures from the same observation. 

\paragraph{Goal conditioned.} Goal conditioned predictions estimate a distribution of goals and then sample from this distribution. In a way, this scheme introduces a form of interpretability to the model, by introducing an explicit latent variable, the goal. For instance, the CTSGI model~\citep{CTSGI} handles stochasticity through its Goal Point Estimation module, which leverages a CVAE to predict a distribution of possible goal points for each pedestrian and their neighbors. NSP~\citep{NSP} handles stochasticity by modeling the aleatoric uncertainty as random forces arising from social interactions (based on the social force model of~\cite{Helbing_1995}).
The model predicts distributions for three main factors: goal attraction, collision avoidance, and environment repulsion. The Graph-TERN model~\citep{GraphTern} addresses stochasticity in pedestrian trajectory prediction through control point prediction. It estimates a set of control points that segment the future trajectory, with each control point serving as a potential intermediate destination. To represent the probability distribution of these points, a multivariate Gaussian Mixture Model is employed, capturing the uncertainty in pedestrian movement and enabling the generation of multiple plausible trajectories. The final trajectory endpoint is determined by linking the predicted control points, resulting in a set of possible destinations that reflect the multimodal nature of pedestrian movement. We can also mention~\citep{X4} goal-conditioned model part that operates by retrieving multi-modal goal candidates for test trajectories from an expert repository of training examples then fed into an end-to-end trajectory predictor. KoopCast~\citep{koopcast} follows the same paradigm by first predicting a multimodal goal distribution using a Mixture Density Network (MDN) conditioned on motion history and local map context. Each sampled goal then guides a Koopman operator–based linear dynamical model that refines the trajectory in a lifted polynomial space. This combination provides interpretable, stable, and efficient multi-modal predictions by coupling deep goal inference with linear Koopman dynamics.

\paragraph{Heatmaps.}The use of heatmaps and energy maps is another way to represent stochasticity. For instance, STHGLU~\citep{STHGLU} generates a global scene heatmap based on training data to represent the likelihood of passage in various geographical areas. This heatmap reflects, for example, low 
probabilities of passage in areas with obstacles. A local heatmap is extracted around each pedestrian at each timestep, capturing the evolving environmental information to probabilistically guide the agent's future movements. In SocialCVAE~\citep{SocialCVAE}, a similar approach employs a local interaction energy map to estimate interaction costs between pedestrians and their neighbors. Constructed using an energy-based interaction model, this map predicts the intensity of repulsion between agents and represents future neighborhood occupancy. It serves as a conditioning factor for the CVAE model, enabling it to learn socially plausible movement decisions over time. These methods capture diverse and realistic behaviors by accounting for uncertainties from both social interactions and environmental factors. More recently, SEI~\citep{SEI} introduces an entropy-based stochastic formulation that quantifies uncertainty directly from displacement, velocity, and acceleration distributions. This approach promotes diverse yet socially coherent predictions by aligning stochasticity with interaction dynamics rather than injecting random noise.

\paragraph{Generative models.}  
Generative model-based HTP algorithms naturally account for stochasticity in their outputs. 
In the case of (C)VAE, stochasticity is explicitly introduced in the latent space. During inference, the model samples from the latent variable distribution to generate multiple plausible future trajectories. For instance, the ForceFormer model~\citep{ForceFormer} leverages a CVAE framework to generate multiple possible predictions. GANs also map realizations of random variables from a latent space into trajectories, while the training through a min–max adversarial game is specific. They are used in models such as DTGAN~\citep{DTGAN}, for example. Normalizing flows offer an alternative generative approach based on invertible, deterministic transformations, often implemented through neural ODEs. STFlow~\citep{STFlow} uses such continuous flows with equivariant message passing to keep trajectories geometrically consistent, while MoFlow~\citep{MoFlow} applies a one-step flow-matching method via IMLE for fast and diverse motion predictions. Gaussian Mixture Models (GMMs) express uncertainty in a simple parametric form. PPNet~\citep{PPNet} generates probabilistic goals using a GMM, and~\citep{ganeshaaraj2025enhancing} extends this idea with GMM heads to better capture multi-modal trajectory patterns. An alternative perspective is offered by Resonance~\citep{Resonance}, which decomposes trajectory uncertainty into self- and social-induced randomness, sampling two latent “vibrations’’ whose superposition generates diverse and socially consistent futures. Diffusion models incorporate stochasticity in trajectory prediction by using a Markov chain with a Gaussian distribution (the diffusion process). 
For example, BCDiff~\citep{BCDiff} employs two coupled diffusion models to generate both unobserved historical trajectories and future trajectories, enhancing the accuracy and diversity of predictions, particularly in cases with limited observation time.  We can also mention the model from~\citep{X8}, that combines distributional diffusion with bivariate Gaussian distributions to account for uncertainty in the future pedestrian locations.

\modified{In summary, the various approaches to modeling stochasticity present distinct operational trade-offs that shape their applicability in real-world scenarios. Explicit methods, such as goal-conditioning and heatmaps, offer high interpretability and directly tie uncertainty to spatial semantics or environmental constraints. However, they may struggle to encapsulate complex and non-linear social interactions without requiring heavy, handcrafted heuristics. Conversely, deep generative models excel at representing rich, multimodal distributions and capturing social cues directly from data. Yet, this high expressiveness often comes at the expense of computational efficiency, training stability, and latent space interpretability. Recognizing these trade-offs is essential for selecting the appropriate stochastic framework, as researchers must constantly balance the need for diverse, socially compliant trajectory samples against the specific real-time constraints and safety requirements of the target application.}
   
\subsection{Marginal vs. joint prediction} 
\label{subsubsec:marginal}

A key distinction among HTP models is whether the predictive model focuses on individual agents, yielding a \emph{marginal} distribution over their trajectories, or considers the entire set of agents across consecutive frames, producing a \emph{joint} distribution over their trajectories. 
It should be clear that, for most applications such as monitoring or autonomous driving, reasoning \emph{jointly} over the set of agents in the scene is critical to analyze the whole scene dynamics or to elaborate motion plans, because of the dependencies involved in the agents' motion. Using marginal predictions could lead to producing tuples of individual predictions that contradict the joint predictive distribution.    

Joint prediction is inherently more complex due to the high dimensionality of the problem and the technical challenges it presents, such as handling sequences of observations with varying lengths.  This may explain why most early HTP works~\citep{Trajectron++, Y-Net} have focused on marginal predictions. The main advantage is practical: marginal modeling is simpler and more scalable (lower-dimensional output space), and it aligns well with common benchmark pipelines.
The literature most often reports per-agent metrics such as ADE/FDE and their multimodal variants (e.g.,$\text{min}ADE_K$/$\text{min}FDE_K$). These metrics can also be applied to joint models, but they do not assess scene-level joint consistency. Thus, relying primarily on marginal metrics may hide joint inconsistencies, motivating the use of joint/scene-level measures where appropriate (further explanation associated with metrics (see  Section~\ref{subsubsec:metrics}).

Among the few early works addressing joint predictive distributions, a notable example is SocialGAN~\citep{Gupta_2018}, which shares a single latent variable from a GAN's generator across all agent trajectory decoders in a scene. Since then, more joint approaches have emerged. Most of them employ stochastic strategies that simultaneously infer individual latent variables for all agents~\citep{AgentFormer,GroupNet} and typically incorporate a coupling mechanism, such as graphs or pooling, to condition each latent posterior or prior distribution on the complete set of past and future trajectories. 
Note that even if the predictive model is designed to be able to produce samples from the joint predictive distribution, the loss function used to train the system may not necessarily evaluate the quality of the joint outputs; it is a motivation in recent works such as~\cite{weng2023joint} that applies simple modifications to the loss functions used in HTP to enforce the generation of better \emph{joint samples}.


\modified{The choice between marginal and joint prediction involves a critical trade-off between computational scalability and scene-level consistency. Marginal modeling remains a pragmatic, scalable approach that aligns with standard benchmarks where evaluation is performed on a per-agent basis. However, its inherent inability to guarantee the absence of collisions or contradictory overlapping trajectories limits its reliability for safety-critical applications like autonomous navigation. Conversely, joint prediction directly tackles these inter-agent dependencies, offering a more realistic and physically plausible representation of crowd dynamics. Yet, this realism comes at the cost of high dimensionality, varying sequence lengths, and the need for significantly more complex coupling mechanisms and loss functions. This inevitable shift toward joint modeling renders purely individual evaluation metrics insufficient, underscoring the urgent need for new scene-aware evaluation methods capable of capturing the holistic quality of the predictions, a challenge we detail in the next section.}

\begin{table*}[!th]
    \resizebox{\textwidth}{!}{
    \centering
    \begin{tabular}{p{2cm}p{9cm}cp{3.1cm}}
        \hline
         \textbf{Metric} & \textbf{Definition} & \textbf{Formula}& \textbf{Introduced by} \\ \hline
         ADE & Average $L_2$ distance between predicted and ground-truth positions over all time steps (per agent). &  $\displaystyle 
        \frac{1}{N K \,\Delta T}
        \sum_{k=1}^{K} \sum_{i=1}^{N} \sum_{t=T_{\text{obs}}+1}^{T_{\text{pred}}}
        \left\| \hat{\mathbf{y}}_t^{i,k} - \mathbf{y}_t^i \right\|_2
        $  &  ~\citep{ETH}  \\ \hline
        
         FDE & $L_2$ distance between predicted and ground-truth positions at the final time step (per agent). & $\displaystyle 
        \frac{1}{NK}
        \sum_{k=1}^{K} \sum_{i=1}^{N} 
        \left\| \hat{\mathbf{y}}_{T_{\text{pred}}}^{i,k} - \mathbf{y}_{T_{\text{pred}}}^i \right\|_2
        $  &\citep{Alahi_2016}\\ \hline
        
         $\text{minADE}_K$ & Minimum ADE over $K$ trajectory samples (a.k.a.\ Top-$K$ ADE). & $\displaystyle
        \frac{1}{N \,\Delta T}
        \sum_{i=1}^{N}
        \min_{k \in \llbracket 1, K \rrbracket}
        \sum_{t=T_{\text{obs}}+1}^{T_{\text{pred}}}
        \left\| \hat{\mathbf{y}}_t^{i,k} - \mathbf{y}_t^i \right\|_2
        $& \citep{desire} \\ \hline
        
         $\text{minFDE}_K$ & Minimum FDE over $K$ trajectory samples (a.k.a.\ Top-$K$ FDE). & $\displaystyle
        \frac{1}{N}
        \sum_{i=1}^{N}
        \min_{k \in \llbracket 1, K \rrbracket}
        \left\| \hat{\mathbf{y}}_{T_{\text{pred}}}^{i,k} - \mathbf{y}_{T_{\text{pred}}}^i \right\|_2
        $& \citep{desire} \\ \hline
        
         NLL & Negative log-likelihood of the ground-truth trajectory under the predicted distribution (typically via Gaussian KDE). &
        $\displaystyle
        -\frac{1}{N\,\Delta T}
        \sum_{i=1}^{N}\sum_{t=T_{\text{obs}}+1}^{T_{\text{pred}}}
        \log\!\left(
            \frac{1}{K}\sum_{k=1}^{K}
            \mathcal{N}\!\left(
                \mathbf{y}_t^{i}\,\middle|\,
                \hat{\boldsymbol{\mu}}_{t}^{i,k},\,
                \hat{\boldsymbol{\Sigma}}_{t}^{i,k}
            \right)
        \right)
        $ &  \citep{NLL} \\ \hline
        
         Average Mahalanobis Distance& Average Mahalanobis distance between generated trajectories and ground truth, incorporating prediction variance. & 
         $\displaystyle \frac{1}{NK\, T_{\text{pred}}} 
         \sum_{i=1}^{N}
         \sum_{t=T_{\text{obs}}+1}^{T_{\text{pred}}}
         \sum_{k=1}^{K}
         \sqrt{
        \bigl( \mathbf{y}_t^{i} - \hat{\mathbf{y}}_t^{i,k} \bigr)^{\!\top}
        \boldsymbol{\Sigma}_{t}^{-1}
        \bigl( \mathbf{y}_t^{i} - \hat{\mathbf{y}}_t^{i,k} \bigr)}$
        & \citep{Social-Implicit} \\ \hline
        
         Average Maximum Eigenvalue & Average of the largest eigenvalue of the predicted covariance, measuring distribution spread/confidence. & $\displaystyle
         \frac{1}{KN T_{\text{pred}}}
        \sum_{i=1}^{N}
        \sum_{t=T_{\text{obs}}+1}^{T_{\text{pred}}}
         \sum_{k=1}^{K}
        \lambda_{\max}\!\left( \boldsymbol{\Sigma}_{t}^{i,k} \right)$ 
        & \citep{Social-Implicit}\\ \hline
         
         Collision Rate & Percentage/rate of collisions between an agent and its predicted neighbors (holistic over all $k$ samples or optimistic via the best joint sample). &
         $\displaystyle \frac{1}{N(N-1)\Delta T}
\sum_{t = T_{\text{obs}}+1}^{T_{\text{pred}}} \sum_{(i,j) \in \llbracket 1, N \rrbracket^2} \mathds{1}\left[ \left\| \hat{\mathbf{y}}_t^i - \hat{\mathbf{y}}_t^j \right\|_2 < \epsilon \right].$&  \citep{kothari2021}\\ \hline

         JADE & Joint ADE: average $L_2$ error over all agents and time steps for a joint multi-agent sample. &
        $\displaystyle
        \frac{1}{N \,\Delta T}
        \min_{k \in \llbracket 1, K \rrbracket}
         \sum_{i=1}^{N}
         \sum_{t=T_{\text{obs}}+1}^{T_{\text{pred}}}
        \left\| \hat{\mathbf{y}}_t^{i,k} - \mathbf{y}_t^i \right\|_2
        $ & \citep{weng2023joint} \\ \hline
        
         JFDE & Joint FDE: average $L_2$ error at the final time step over all agents for a joint multi-agent sample. &
        $\displaystyle
        \frac{1}{N}
        \min_{k \in \llbracket 1, K \rrbracket}
        \sum_{i=1}^{N}
        \left\| \hat{\mathbf{y}}_{T_{\text{pred}}}^{i,k} - \mathbf{y}_{T_{\text{pred}}}^i \right\|_2
        $ &  \citep{weng2023joint} \\ \hline
        
         Miss Rate & Percentage of predicted agent coordinates that fall outside a threshold region around ground truth at each frame. &
        $\displaystyle
        \frac{1}{NK\,\Delta T}
        \sum_{i=1}^{N}\sum_{t=T_{\text{obs}}+1}^{T_{\text{pred}}}
        \sum_{k=1}^{K}
        \mathds{1}\!\left[
        \left\|\hat{\mathbf{y}}_{t}^{i,k}-\mathbf{y}_{t}^{i}\right\|_2>\delta
        \right]
        $ & \citep{argoverse}\\ \hline
        
         
         Entropy & Entropy of the distribution of per-agent or collective errors between predicted and ground-truth trajectories. & $\displaystyle
        -\frac{1}{N\,\Delta T}
        \sum_{i=1}^{N}\sum_{t=T_{\text{obs}}+1}^{T_{\text{pred}}}
        \mathbb{E}_{k}\!\big[
        \log p_{t,i}\!\big(\hat{\mathbf{y}}_{t}^{i,k}\big)
        \big]
        $ & \citep{entropy1,entropy2} \\ \hline
        
         AUC & Summarizes the expected min-ADE when randomly sampling $K$ trajectories out of $N$ for all $K\in\{1,\dots,N\}$; evaluates the global quality of the whole prediction set. & 
         $\displaystyle 
         \sum_{i=1}^{N} 
         \sum_{K=1}^{k} \frac{1}{\Delta T} \sum_{j=1}^{k - K + 1}\frac{\binom{k - j}{K - 1}}{\binom{k}{K}} \left\| \hat{\mathbf{Y}}^{i,j} - \mathbf{Y}^{i} \right\|_{2,1}$
         & \citep{auc} \\\hline
         ACE & measures the deviation between predicted and true causal 
effects of each agent, computed from counterfactual removals.  & $\displaystyle 
\frac{1}{\Delta T KN}\sum_{i=1}^{N}\sum_{k=1}^{K} \sum_{t=T_{\text{obs}}+1}^{T_{\text{pred}}}
\bigl|\lVert \hat{\mathbf{y}}_{\varnothing}-\hat{\mathbf{y}}_{t}^{i,k}\rVert_{2}-\mathcal{E}_{i}\bigr|
$ & \citep{ACE} \\ \hline
    \end{tabular}
    }
    \vspace{0.1cm}
    \caption{Overview of pedestrian trajectory metrics. All methods observe $N$ agents during $T_{\text{obs}}$ steps and predict $K$ future trajectory samples $\hat{\mathbf{y}}_{t}^{i,k}$ over $t\in\llbracket T_{\text{obs}}+1, T_{\text{pred}}\rrbracket$, with ground truth $\mathbf{y}_t^i$. We denote $\Delta T = T_{\text{pred}}-T_{\text{obs}}$ and use the Euclidean norm $\|\cdot\|_2$. For probabilistic metrics, $\hat{\boldsymbol{\mu}}_{t}^{i,k}$ and $\hat{\boldsymbol{\Sigma}}_{t}^{i,k}$ are the predicted mean and covariance (and $\lambda_{\max}(\cdot)$ the largest eigenvalue). $\mathds{1}[\cdot]$ is the indicator function; $\epsilon$ is the collision-distance threshold and $\delta$ the miss threshold. For ACE, $\hat{\mathbf{y}}_{\varnothing}$ is the counterfactual prediction with agent $i$ removed, and $\mathcal{E}_i$ is the corresponding ground-truth causal effect for agent $i$. 
 }
    \label{tab:metrics}
\end{table*}

\subsection{Evaluation metrics}
\label{subsubsec:metrics}

Evaluation metrics are paramount to assess the performance of various trajectory generation models through different perspectives. As such, a number of relevant metrics have been introduced by different communities, including computer vision, robotics, and computer graphics, seeking to quantify the accuracy and robustness of the predictions. A structured overview of the most widely used trajectory prediction metrics, their \modified{mathematical expressions} and original references \modified{are} summarized in Table~\ref{tab:metrics}.

The majority of HTP models perform marginal predictions by focusing on predicting trajectories for individual agents and comparing them to the ground truth. In this setting, the most commonly used evaluation metrics for deterministic models are the \emph{average displacement error} (ADE) and the \emph{final displacement error} (FDE)~\citep{ETH}. ADE measures the 
mean L2 distance between the predicted and ground truth positions of an agent's trajectory over a sequence, while FDE measures the L2 distance at the final time step.

In the case of multimodal prediction, the metric should involve the whole predictive distribution and its ground truth counterpart. Unfortunately, the latter is not available, as we just have one sample from the ground truth distribution. Both the ADE and FDE can be extended to multimodal prediction models by computing the $\text{minADE}_{k}$ and $\text{minFDE}_{k}$, i.e., the \emph{minimum ADE} and \emph{minimum FDE}, respectively, over $k$ predictions from the generated distribution. These metrics are also commonly referred to as \emph{Top-k ADE} and \emph{Top-k FDE}, with typical values for $k$ in the literature being $k=3$, $k=5$, $k=10$, and $k=20$. 

However, we note that as $k$ increases, $\text{minADE/minFDE}$ metrics become increasingly biased, since by producing an arbitrarily high number of samples, the generative model can produce deceptively low error values~\citep{kothari2021}. Nevertheless, the ADE/FDE differences in many SOTA HTP models are often only a few centimeters, and even simple linear models can perform surprisingly well~\citep{Social-Implicit,scholler2020constant}. This raises concerns about the reliability of the ranking of such models in the presence of sensor noise and other types of uncertainty. 

Beyond the biases of $\text{minADE}_k$ and $\text{minFDE}_k$, more holistic metrics have been proposed to assess the global quality and consistency of multimodal prediction sets. The \emph{Area Under the minADE Curve} (AUC)~\citep{auc} summarizes the expected minimum ADE obtained when randomly sampling $K$ trajectories among the $N$ generated ones, for all $K \in \{1,\dots,N\}$. Unlike Top-$k$ metrics, which evaluate only the best subset of predictions, AUC provides a global measure of the overall prediction set quality across all sampling budgets.

Another metric for multimodal prediction that focuses on the quality of the predictive distribution is the \emph{negative log-likelihood} (NLL), which evaluates the likelihood of the ground truth trajectories~\citep{IvanovicPavone2019} under an estimate of the predicted distribution, typically obtained by Gaussian kernel density estimation~\citep{parzen1962}. Complementarily, the \emph{Average Causal Effect} (ACE)~\citep{ACE} evaluates whether the predicted trajectories correctly capture the causal influence between interacting agents. It measures the deviation between predicted and true causal effects under counterfactual agent removals, thus going beyond geometric accuracy to assess the causal structure underlying the model’s predictions.
More recently, the \emph{Average Mahalanobis Distance} has been introduced as a robust alternative to $\text{minADE/minFDE}$. This metric accounts not only for the distance between generated trajectories and ground truth, but also for the variance of the predictions~\citep{Social-Implicit}. The \emph{Average Maximum Eigenvalue} of the covariance matrix of the predicted distribution has also been introduced as a complementary metric to evaluate the overall spread of the predictive trajectories, and thus the confidence of the predictions~\citep{Social-Implicit}.

Other metrics for joint prediction have been explored by different communities and can be used for assessing HTP models. Notable among these are the \emph{Percentage/Rate of Collisions} of a given agent with its predicted neighboring agents~\citep{vanToll2021Algorithms}, either by providing a holistic evaluation by considering all $k$ samples~\citep{kothari2021,a2x,weng2023joint} or an optimistic evaluation by considering the sample with the minimum joint ADE~\citep{weng2023joint}. 
In the autonomous vehicle literature, the \emph{Overlap rate}~\citep{waymo} extends the \emph{Percentage/Rate of Collisions} to consider collisions with the static part of the environment, with additional metrics being inspired by object detection, including the 
\emph{Miss Rate}~\citep{argoverse,waymo} that reports the percentage of predicted agents' coordinates at each frame that lie outside of a threshold area defined around the corresponding ground truth coordinates. In computer graphics, the \emph{Entropy} of the distribution of errors between crowd predicted trajectories and ground truth either at a per-agent level~\citep{entropy2} or collectively~\citep{entropy1} has been proposed to assess crowd simulation models. 
The \emph{Entropy} metric accounts for the noise inherently present in real-world datasets and the fact that such datasets are finite and cannot necessarily capture the all ground truth behaviors. However, its applicability has only been tested on physics-based models. 
Similarly,~\cite{charalambous2014data} evaluates physics-based models by casting the problem as an anomaly detection problem, while~\cite{he2020informative} explored average likelihood and KL-divergence evaluation metrics.

\section{Discussion}
\label{sec:discussion}

\citet{rudenko} have raised three central questions:
\begin{itemize}
    \item Are the evaluation techniques to measure prediction performance good enough and follow best practices ?
    \item Have all prediction methods arrived at the same performance level and the choice of the modeling approach does not matter anymore ?
    \item Is motion prediction solved ?
\end{itemize}

Since Rudenko's publication~\citep{Rudenko2020}, these questions remain largely open. Benchmarks have evolved little and substantial room for improvement persists.
This section discusses our observations regarding \emph{evaluation metrics}, \emph{datasets}, \emph{architectural trends}, and \emph{readiness for real-world deployment}. We identify shortcomings in the current HTP practice and propose actionable directions that we believe are necessary for progress beyond incremental leaderboard gains on ETH/UCY-like benchmarks.

\subsection{On metrics: beyond \texorpdfstring{minADE/minFDE}{minADE/minFDE}}
\label{subsec:metric_discussion}

Displacement metrics such as minADE/minFDE remain the dominant evaluation tools in trajectory forecasting, yet they provide only a narrow view of performance. Their best-of-$k$ formulation rewards models that generate large and diverse sets of samples and assumes that a single ground-truth future is the only acceptable outcome. As a result, these scores offer little insight into distributional fidelity, uncertainty quality, or interaction realism; three aspects that are essential for deploying predictive models in real environments.

In practice, relying solely on minADE/minFDE leads to misleading rankings: models can achieve low errors without producing calibrated, socially plausible, or decision-relevant forecasts. These metrics therefore remain useful but insufficient as stand-alone indicators. This aligns with findings that models may yield accurate point estimates yet remain miscalibrated without explicit uncertainty recalibration~\citep{kuleshov}.

\paragraph{Proper scoring and calibration.}

A more meaningful assessment requires evaluating the \emph{entire predictive distribution}. Proper scoring rules such as log-likelihood~\citep{NLL} and entropy~\citep{entropy1} quantify whether uncertainty is correctly concentrated in areas where future motion is likely to occur, such as the energy~\citep{Gneiting} and variogram scores~\citep{Scheuerer} capture spatial spread and temporal consistency that displacement errors systematically overlook. 

Equally important is uncertainty calibration: forecasts should express confidence levels that match empirical accuracy. Coverage–reliability curves, Root Mean Squared Calibration Error (RMSCE), Mean Absolute Calibration Error (MACE), or model-agnostic tools such as Expected Calibration Error (ECE) provide direct diagnostics of over- or under-confidence. Conformal prediction further offers distribution-free coverage guarantees, with post-hoc importance reweighting enabling target coverage without resampling~\citep{canche, Egocentric}. Empirical results consistently show that better-calibrated models support safer and more reliable downstream decision-making.

\paragraph{Interaction-aware evaluation.}


Realistic forecasting requires more than pointwise accuracy: predictions must remain \emph{socially feasible}. \modified{Formally, \textbf{interaction-aware metrics}
evaluate the degree to which a model respects the relational constraints and collective dependencies between agents, ensuring that predicted trajectories preserve the structural coherence of the social system.} Displacement errors alone cannot reveal whether agents respect \modified{task-specific semantic constraints such as} collision avoidance, interpersonal distances, or group cohesion. 
Interaction-aware metrics explicitly test these aspects and often reveal failures that minADE/minFDE entirely miss~\citep{kothari2021}; \modified{moreover, while dyadic metrics focus on pairwise relations, interaction measures derived from higher-order network analysis, such as those utilizing hypergraphs, can reveal group dynamics and collective behaviors that remain invisible in simple graph-based models~\citep{battiston2025higher}.} For instance, predictions that are \modified{socially consistent (e.g., maintaining group structure)} but imprecise from the ground-truth may be preferable to accurate predictions but \modified{violate the underlying relational architecture.}
Evaluating interactions is therefore essential, and metrics that quantify \modified{system-wide coordination} should be treated as a multi-criteria evaluation \modified{where task-dependent indicators (like collision rates) are presented as specific instances of a broader interaction-agnostic framework.}

\paragraph{Top-$K$ as a \emph{curve}, not a point.}

Reporting results at a single $k$ (typically $k{=}20$) masks how effectively a model uses its multimodal capacity. Plotting performance as a function of $k$ exposes whether improvements saturate early, whether models rely on excessive sampling, and how many hypotheses are genuinely useful to downstream modules. Matching $k$ to the computational budget of planners (e.g., $k{=}5$) also yields more meaningful comparisons than reporting performance at arbitrary values. Treating $k$ as a curve rather than a point therefore provides a more transparent and equitable assessment. This is similar to ROC curves, where varying the decision threshold reveals the full performance profile, and to AUC, which summarizes performance over that range.

\paragraph{Joint vs.\ marginal quality.}

Marginal metrics evaluate each agent independently and often hide inconsistencies at the scene level. Selecting the best trajectory per agent can produce combinations that never appear coherently in any sampled future, despite low per-agent error. Joint metrics address this issue by evaluating all agents within the same sample, providing a more faithful assessment of collective behavior and revealing the optimistic bias of marginal scores.

Both perspectives matter: marginal metrics describe individual-agent accuracy, whereas joint metrics test coordinated realism. The appropriate emphasis depends on the application; for instance, robot navigation may prioritize marginal accuracy and safety checks, while crowd-level forecasting requires joint evaluation to avoid contradictory or socially implausible predictions.

\begin{figure}[t]
\centering

\begin{subfigure}{0.45\linewidth}
\centering
\small
\setlength{\tabcolsep}{6pt}
\begin{tabular}{lc}
\hline
Dataset & Density (1/m$^2$) \\
\hline
Zara01 &  0.206 \\
Zara02 & 0.267 \\ 
Univ   & 0.391\\ 
ETH    & 0.148\\ 
\hline
\end{tabular}
\vspace{0.9cm}
\end{subfigure}
\begin{subfigure}{0.48\linewidth}
\centering
\includegraphics[width=\linewidth]{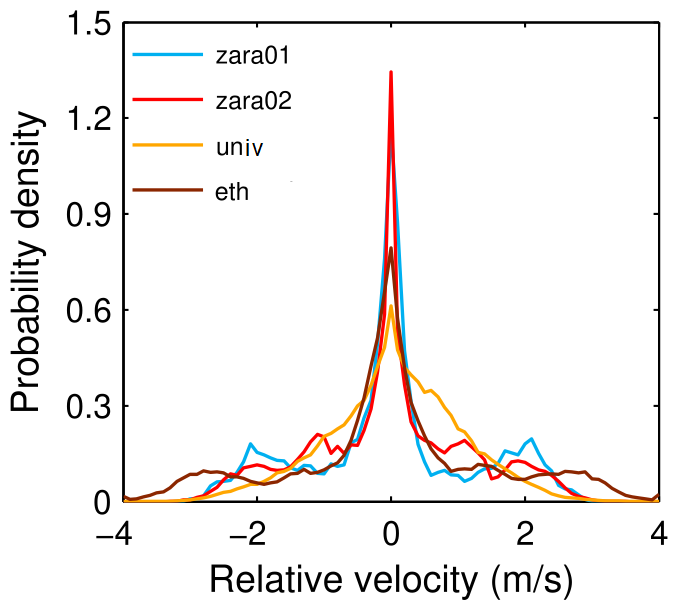}
\end{subfigure}

\caption{\modified{Comparison of ETH/UCY datasets.
(a) Average pedestrian density computed using Edie’s generalized definition~\citep{edie,karamouzas2014universal}.
(b) Distribution of relative pedestrian velocities. Relative velocity measures the rate of change of the separation between two pedestrians: negative values indicate approaching motion, while positive values indicate moving apart.}}
\label{fig:datasets_comp}
\end{figure}

\subsection{\modified{On datasets: diversity, density, and ``openness''}}
\label{subsec:dataset_discussion}

Much of the existing research still depends on a handful of older datasets, notably ETH and UCY. This consistency makes it easier to compare different methods, but it also limits the scope of the conclusions that can be drawn. These datasets are short and only moderately crowded, mostly captured from overhead viewpoints, and offer little variation in 
\modified{interaction conditions (see Figure~\ref{fig:datasets_comp}).} As a result, they fail to reflect more demanding scenarios, such as mixed viewpoints, densely packed groups, or richer social dynamics. Models trained and evaluated under these simplified conditions often struggle when exposed to environments that are more diverse or complex. In addition, datasets that truly represent high-density crowds are still missing, partly because of technical difficulties and ethical concerns. This creates a significant gap between the current benchmarks and the concrete situations where reliable prediction is most needed.


\modified{Because many recent benchmarks are multifaceted, the datasets referenced in the following discussion are illustrative rather than exhaustive, selected because they exemplify particular properties (e.g., high density, multimodal sensing, or privacy‑preserving pipelines). Several datasets span multiple categories, and their inclusion under a specific point reflects the aspect most relevant to that discussion rather than an exclusive characterization. 
}

\paragraph{Increase scenario complexity and density.}

To promote robustness, evaluation datasets should expose models to rigorously defined difficulty levels, in contrast to the common reliance on coarse summary statistics. One approach is to characterize each scenario using explicit “complexity cards” that specify features such as crowd-density patterns~\citep{franchi}, environmental layout (including bottlenecks, intersections, or narrow passages), and interaction types (e.g., overtaking, merging, splitting, crossing paths, and leader–follower dynamics). Table~\ref{tab:dataset} shows that commonly used benchmarks span drastically different crowd regimes (from $\sim$2--3 persons/frame in JRDB/inD to $\sim$80--100 in Grand Central/ATC on average, with peaks up to 1818), so reporting results by density bands is crucial. Evaluating performance across different density and complexity levels can expose failure cases that remain hidden under aggregate metrics.

Findings from social-navigation research already show that model performance deteriorates most noticeably in crowded and spatially constrained settings~\citep{stratton2024}, which emphasizes the importance of benchmarks that capture these situations. A stratified evaluation approach makes it easier to pinpoint when models break down and what conditions cause these errors, especially in scenarios that mirror real-world crowd behaviors.

\paragraph{Perspective and input modalities.}
Broadening the range of viewpoints and sensing modalities is \modified{equally important,} 
when accounting for noise and real-world data. As shown in Table~\ref{tab:dataset}, most existing datasets rely on fixed overhead cameras, yet first-person recordings are often more relevant for robot navigation and human–robot interaction. Newer datasets collected from mobile platforms combine RGB, LiDAR, depth, odometry, and other signals, \modified{capturing motion under realistic occlusions and changing viewpoints, as seen in datasets like JRDB~\citep{jrdb} and THÖR~\citep{thor}, which integrate mobile-platform data with diverse sensor suites}. 

\modified{Furthermore, benchmarks such as THÖR-MAGNI~\citep{magniact} and JRDB-ACT~\citep{jrdbact} leverage motion-capture and rich social annotations (e.g., 3D gaze, group intent) to complement traditional trajectory data.}
Multimodal datasets are essential for training models that can assimilate \modified{heterogeneous signals,} 
integrating noisy and incomplete observations of past positions within conditions that better reflect real sensing situations. Clearly specifying which modalities are included also encourages ablation studies to separate gains due to privileged inputs from those reflecting genuinely robust modeling.

\paragraph{Data protection constraints.}

Ultimately, robust modeling benefits from raw sensor data, but privacy regulations such as the GDPR impose strong limits on releasing large-scale, realistic crowd recordings~\citep{data}. Even so, there are several ways to work within these constraints. Using overhead viewpoints or relying solely on LiDAR can prevent the capture of identifiable facial features while still retaining useful motion data. Sharing only 2D or 3D trajectories, or point-cloud representations instead of full RGB video, further lowers the risk of re-identifying individuals. When RGB footage is necessary, video/image anonymization and pseudonymization procedures can \modified{ensure that identities cannot be tracked across frames~\citep{ano,securing}. This practice is increasingly systematic; for instance, the JRDB and THÖR-MAGNI ecosystems provide raw or processed visual data only after anonymization or pseudonymization.} Synthetic data can also play a valuable role, whether produced via simulations or trajectory augmentation techniques,  like in the Multi-Object Tracking field~\citep{motsynth}, provided its plausibility is validated through tests ensuring it is not easily distinguishable from real data. Even when visual recordings cannot be distributed, providing detailed metadata, documentation of data-collection practices, and evaluation tools remains crucial for reproducibility. With thoughtful safeguards, it is still possible to build datasets that are both more diverse and respectful of privacy requirements.

\subsection{On architectures: interpretability, hybridization}
\label{subsec:architectures_discussion}

\paragraph{Interpretability and diagnosis.}
Many high-performing models remain difficult to interpret, which limits their trust and adoption in safety-critical contexts. In response, the community has developed a range of methods aimed at improving interpretability in spatiotemporal reasoning, from attention visualization and feature attribution~\citep{chefer2021transformer, makansi2021you} to structural probing of latent maps~\citep{hewitt2019structural}. Counterfactual probes such as removing an agent, blocking an exit, or altering map geometry, can expose causal sensitivities. Diagnostic error taxonomies (e.g., missed goal, late yielding, group split) and per-category metrics should become standard reporting practice, as they enable systematic model comparison and data-driven model refinement with human oversight. \cite{makansi2021you} showed that most SOTA architectures ``mostly walk alone'', i.e., they capture temporal continuity better than true multi-agent interaction. Such analyses highlight the importance of interpretability tools to diagnose what aspects of human motion these models actually learn, and to design datasets and metrics that promote causal social understanding.

\paragraph{Hybrid models.}

Hybrid architectures offer a promising path forward, combining the flexibility of learned predictors with structured knowledge such as physics-based priors or game-theoretic reasoning. Purely data-driven models excel at pattern fitting, yet they often fail under distribution shift and provide limited guarantees on feasibility or mutual consistency. Hybrid approaches turn these challenges into opportunities: they embed interpretable quantities, encode invariances and prior constraints, and allow for generalization far beyond the training domain. Ultimately, such models may provide both expressive learning and constraint guarantees.

Recent works illustrate the diversity of such formulations: Bayesian inverse games~\citep{liu2024auto}, mixed-strategy game models using generative diffusion~\citep{sun2025inverse}, and energy-based potential games~\citep{diehl2023energy} all embed strategic interaction into the prediction process. Other frameworks integrate learned priors within dynamic game solvers~\citep{lidard2024blending} or impose structured reasoning about payoffs and intentions for socially compliant motion~\citep{schwarting2019social}. Physics-informed hybrids such as NSP~\citep{NSP} and PPAL~\citep{PPAL} show how explicit physical constraints can complement learning to improve robustness.

Our conviction is that a promising path forward is to determine which components of the prediction stack benefit from structure (e.g., collision avoidance, feasibility, interaction reasoning) and which should remain learned. Strengthening physical and behavioral constraints while keeping models lightweight and explainable appears essential for building prediction systems that generalize reliably beyond their training conditions.

 \subsection{Towards robotics: Closing the sim-to-real loop}
\label{subsec:robotics_discussion}

\paragraph{From open-loop accuracy to decision-aware evaluation.}
In robotics, the limitations of displacement metrics discussed in~\ref{subsec:metric_discussion} become critical: a predictor is useful only insofar as it improves \emph{closed-loop} behavior. Evaluation should therefore complement minADE/minFDE with downstream, decision-aware criteria: navigation success rate, time-to-goal, deadlocks/oscillations, human intervention rate, clearance margins and comfort/proxemics, as well as inference latency on the target platform. This also calls for interaction-aware checks under realistic scenario libraries covering the robot’s operational design domain, where calibration is actionable because it directly conditions risk-sensitive planning~\citep{gao2022evaluation,core,biswas2022socnavbench,nishimura2020risk,dik2024graph,francis2025principles}.

\paragraph{Bridging onboard sensing, tracking noise, and human--robot interaction shift.}
The dataset gaps described in~\ref{subsec:dataset_discussion} are amplified when moving to robots: inputs are not clean trajectories but noisy, partial observations produced by onboard perception. A key sim-to-real barrier is the mismatch between curated tracking data and real robot streams affected by occlusions, viewpoint changes, ego-motion, missed detections and identity switches~\citep{NATRA}. Closing this gap motivates \emph{tracking-to-prediction} benchmarks that evaluate the full chain sensing$\rightarrow$tracking$\rightarrow$prediction, and encourage releasing modality/uncertainty metadata when raw data cannot be shared. Moreover, human motion changes in the presence of a robot, and this shift depends on the robot’s embodiment and behavior; thus, deployable HTP models should be trained or at least validated on robot-in-the-loop data capturing such interaction effects~\citep{zhang2022hri,stratton2024}.

\paragraph{Real-time constraints, interpretability, and hybridization in the loop.}
Finally, the architectural trends discussed in~\ref{subsec:architectures_discussion} must be instantiated under robotics constraints: bounded and predictable runtime, robustness to missing/noisy inputs, and uncertainty that is usable by planners. This favors models that are not only accurate but diagnosable and compatible with safety-oriented decision stacks. Hybrid approaches are particularly relevant here: injecting feasibility and interaction structure can improve robustness under shift while keeping behaviors explainable, \modified{a key ingredient in building trustworthy robotic systems~\citep{leila24,halilovic2025explainable}}. Crucially, pilot deployments help reveal when open-loop gains do not translate to navigation improvements, making iterative closed-loop testing the practical mechanism to align metrics, data, and models and thereby truly close the sim-to-real loop~\citep{poddar2023crowd,aslam2025}.

\section{Conclusion}\label{sec:conclusion}
This survey has reviewed the main components of multi-agent human trajectory prediction, from problem formulation and datasets to modeling strategies and evaluation practices. Our analysis reveals that the field is strongly shaped by academic inertia: widely used datasets, metrics and architectural patterns persist even when they only partially reflect the demands of real-world navigation. This inertia has helped consolidate benchmarks but may now limit progress toward models that handle dense interactions, uncertainty and deployment constraints.
The discussion has highlighted three areas where change is most needed: more informative and diverse datasets, evaluation metrics that reflect decision quality rather than only displacement error, and architectures that balance expressiveness with interpretability and robustness. Challenges at the deployment level, such as closed-loop evaluation or ``tracking-to-prediction'', should be supported.
By outlining these gaps and opportunities, we hope to support methodological choices that strengthen both scientific understanding and practical reliability, and to encourage the development of prediction systems that are increasingly aligned with the demands of real-world multi-agent navigation.

\section*{Acknowledgments}

    This study was supported by the OLICOW project, funded by the French National Research Agency (ANR) under grant ANR-22-CE22-0005. 
    This work was also supported by the National Science Foundation (NSF) under Grant No. IIS-2402338.     

\bibliographystyle{unsrtnat}
\bibliography{references.bib}

@inproceedings{gu2022MID,
      title={Stochastic Trajectory Prediction via Motion Indeterminacy Diffusion},
      author={Gu, Tianpei and Chen, Guangyi and Li, Junlong and Lin, Chunze and Rao, Yongming and Zhou, Jie and Lu, Jiwen},
      booktitle={IEEE/CVF Conference on Computer Vision and Pattern Recognition},
      pages={17113--17122},
      year={2022}
    }

@InProceedings{vandenBerg2011,
  author    = {Jur van den Berg and Stephen J. Guy and Ming Lin and Dinesh Manocha},
  title     = {{Reciprocal n-body Collision Avoidance}},
  booktitle = {Robotics Research: The 14th International Symposium},
  year      = {2011},
  volume    = {70},
  series    = {Springer Tracts in Advanced Robotics},
  pages     = {3--19},
}

@article{Rudenko2020,
author = {Andrey Rudenko and Luigi Palmieri and Michael Herman and Kris M Kitani and Dariu M Gavrila and Kai O Arras},
title ={Human motion trajectory prediction: a survey},
journal = {The International Journal of Robotics Research},
volume = {39},
number = {8},
pages = {895-935},
year = {2020}
}

@INPROCEEDINGS{vandenBerg2008,
  author={van den Berg, Jur and Ming Lin and Manocha, Dinesh},
  booktitle={IEEE International Conference on Robotics and Automation}, 
  title={Reciprocal Velocity Obstacles for real-time multi-agent navigation}, 
  year={2008},
  pages={1928-1935},
}

@inproceedings{Lin2024gigatraj,
title={GigaTraj: Predicting Long-term Trajectories of Hundreds of Pedestrians in Gigapixel Complex Scenes},
author={Lin, Haozhe and Wei, Chunyu and He, Li and Guo, Yuchen and Zhao, Yunqi and Li, Shanglong and Fang, Lu},
booktitle={IEEE/CVF Conference on Computer Vision and Pattern Recognition },
pages={19331-19340},
year={2024},
}

@article{rudenko,
  author       = {Andrey Rudenko and
                  Luigi Palmieri and
                  Michael Herman and
                  Kris M. Kitani and
                  Dariu M. Gavrila and
                  Kai Oliver Arras},
  title        = {Human Motion Trajectory Prediction: {A} Survey},
  journal      = {CoRR},
  volume       = {abs/1905.06113},
  year         = {2019},
  eprinttype    = {arXiv},
  eprint       = {1905.06113}
}

@article{Helbing_1995,
   title={Social force model for pedestrian dynamics},
   volume={51},
   ISSN={1095-3787},
   number={5},
   journal={Physical Review E},
   publisher={American Physical Society (APS)},
   author={Helbing, Dirk and Molnár, Péter},
   year={1995},
   month={05}, pages={4282–4286} }

@article{TrajPRedZhou_2024,
   title={TrajPRed: Trajectory Prediction With Region-Based Relation Learning},
   ISSN={1558-0016},
   journal={IEEE Transactions on Intelligent Transportation Systems},
   publisher={Institute of Electrical and Electronics Engineers (IEEE)},
   author={Zhou, Chen and AlRegib, Ghassan and Parchami, Armin and Singh, Kunjan},
   year={2024},
  volume={25},
  number={8},
  pages={9787-9796}}

@article{Hochreiter_1997,
  title={Long short-term memory},
  author={Hochreiter, Sepp and Schmidhuber, Jürgen},
  journal={Neural computation},
  volume={9},
  number={8},
  pages={1735--1780},
  year={1997},
  publisher={MIT Press}
}

@inproceedings{Goodfellow_2014,
  title={Generative adversarial nets},
  author={Goodfellow, Ian and Pouget-Abadie, Jean and Mirza, Mehdi and Xu, Bing and Warde-Farley, David and Ozair, Sherjil and Courville, Aaron and Bengio, Yoshua},
  booktitle={Advances in Neural Information Processing Systems},
  pages={2672--2680},
  year={2014}
}

@inproceedings{Alahi_2016,
  title={Social LSTM: Human trajectory prediction in crowded spaces},
  author={Alahi, Alexandre and Goel, Kratarth and Ramanathan, Vibhash and Robicquet, Alexandre and Fei-Fei, Li and Savarese, Silvio},
  booktitle={Proceedings of the IEEE conference on computer vision and pattern recognition},
  pages={961--971},
  year={2016}
}

@inproceedings{Gupta_2018,
  title={Social GAN: Socially acceptable trajectories with generative adversarial networks},
  author={Gupta, Agrim and Johnson, Justin and Fei-Fei, Li and Savarese, Silvio and Alahi, Alexandre},
  booktitle={Proceedings of the IEEE Conference on Computer Vision and Pattern Recognition},
  pages={2255--2264},
  year={2018}
}

@inproceedings{NSP,
  title={Human trajectory prediction via neural social physics},
  author={Yue, Jiangbei and Manocha, Dinesh and Wang, He},
  booktitle={European conference on computer vision},
  pages={376--394},
  year={2022},
  organization={Springer}
}

@article{Sharma_2022,
  title={Pedestrian intention prediction for autonomous vehicles: A comprehensive survey},
  author={Sharma, N and Dhiman, C and Indu, S},
  journal={Neurocomputing},
  volume={492},
  pages={9-26},
  year={2022},
  publisher={Elsevier}
}

@article{Zhang_2023,
  title={Pedestrian behavior prediction using deep learning methods for urban scenarios: A review},
  author={Zhang, C and Berger, C},
  journal={IEEE Transactions on Intelligent Transportation Systems},
  year={2023},
  volume={24},
  number={10},
  pages={10279-10301},
  publisher={IEEE}
}

@article{Capy_2023,
  title={Pedestrians and cyclists' intention estimation for the purpose of autonomous driving—A systematic review},
  author={Capy, S and Venture, G and others},
  journal={International Journal of Automotive Engineering},
  volume={14},
  number={1},
  year={2023}
}

@article{Kong_2023,
  title={Mobility trajectory generation: a survey},
  author={Kong, X and Chen, Q and Hou, M and Wang, H and Xia, F},
  journal={Artificial Intelligence Review},
  year={2023},
  publisher={Springer},
  volume={56},
  number={3},
  pages={3057-3098},
}

@article{Singamaneni_2024,
  title={A survey on socially aware robot navigation: Taxonomy and future challenges},
  author={Singamaneni, PT and Bachiller-Burgos, P and others},
  journal={International Journal of Robotics Research},
  year={2024},
volume = {43},
number = {10},
pages = {1533-1572},
  publisher={SAGE}
}

@article{Galvao_2023,
  title={Pedestrian and vehicle behaviour prediction in autonomous vehicle system—A review},
  author={Galvão, LG and Huda, MN},
  journal={Expert Systems with Applications},
  volume={221},
  year={2023},
  publisher={Elsevier}
}

@inproceedings{Almeida_2023,
  title={THOR-Magni: Comparative analysis of deep learning models for role-conditioned human motion prediction},
  author={De Almeida, TR and Rudenko, A and others},
  booktitle={Proceedings of the ICCV 2023 Workshop},
  year={2023},
  pages={2200-2209},
  publisher={IEEE}
}

@article{Fu_2024,
  title={Summary and reflections on pedestrian trajectory prediction in the field of autonomous driving},
  author={Fu, Z and Jiang, K and Xie, C and Xu, Y and Huang, J and Gao, X},
  journal={IEEE Transactions on Intelligent Transportation Systems},
  year={2024},
  publisher={IEEE}
}

@article{Lerner_2007,
  author = {A. Lerner and Y. Chrysanthou and D. Lischinski},
  title = {Crowds by example},
  journal = {Computer Graphics Forum},
  year = {2007},
  volume = {26},
  number = {3},
  pages = {655-664}
}

@article{moher2009preferred,
  title={Preferred reporting items for systematic reviews and meta-analyses: the PRISMA statement},
  author={Moher, David and Liberati, Alessandro and Tetzlaff, Jennifer and Altman, Douglas G and PRISMA Group*, t},
  journal={Annals of internal medicine},
  volume={151},
  number={4},
  pages={264--269},
  year={2009},
  publisher={American College of Physicians}
}

@inproceedings{Y-Net,
  title={From goals, waypoints \& paths to long term human trajectory forecasting},
  author={Mangalam, Karttikeya and An, Yang and Girase, Harshayu and Malik, Jitendra},
  booktitle={IEEE/CVF International Conference on Computer Vision},
  pages={15233--15242},
  year={2021}
}

@inproceedings{M2P3,
  title={M2p3: multimodal multi-pedestrian path prediction by self-driving cars with egocentric vision},
  author={Poibrenski, Atanas and Klusch, Matthias and Vozniak, Igor and M{\"u}ller, Christian},
  booktitle={Proceedings of the 35th Annual ACM Symposium on Applied Computing},
  pages={190--197},
  year={2020}
}

@inproceedings{CARPe,
  title={Carpe posterum: A convolutional approach for real-time pedestrian path prediction},
  author={Mendieta, Mat{\'\i}as and Tabkhi, Hamed},
  booktitle={Proceedings of the AAAI Conference on Artificial Intelligence},
  number={3},
  pages={2346--2354},
  year={2021}
}

@ARTICLE{DTGAN,
  author={Xie, Jiajia and Zhang, Sheng and Xia, Beihao and Xiao, Zhu and Jiang, Hongbo and Zhou, Siwang and Qin, Zheng and Chen, Hongyang},
  journal={IEEE Transactions on Multimedia}, 
  title={Pedestrian Trajectory Prediction Based on Social Interactions Learning With Random Weights}, 
  year={2024},
  volume={26},
  number={},
  pages={7503-7515}
}

@article{DynGroupNet,
title = {Dynamic-group-aware networks for multi-agent trajectory prediction with relational reasoning},
journal = {Neural Networks},
volume = {170},
pages = {564-577},
year = {2024},
issn = {0893-6080},
author = {Chenxin Xu and Yuxi Wei and Bohan Tang and Sheng Yin and Ya Zhang and Siheng Chen and Yanfeng Wang}
}

@INPROCEEDINGS {Eqmotion,
author = {C. Xu and R. T. Tan and Y. Tan and S. Chen and Y. Wang and X. Wang and Y. Wang},
booktitle = {IEEE/CVF Conference on Computer Vision and Pattern Recognition},
title = {EqMotion: Equivariant Multi-Agent Motion Prediction with Invariant Interaction Reasoning},
year = {2023},
volume = {},
issn = {},
pages = {1410-1420},

}

@article{X8,
title = {Uncertainty-aware pedestrian trajectory prediction via distributional diffusion},
journal = {Knowledge-Based Systems},
volume = {296},
pages = {111862},
year = {2024},
issn = {0950-7051},
author = {Yao Liu and Zesheng Ye and Rui Wang and Binghao Li and Quan Z. Sheng and Lina Yao},
}

@inproceedings{AVGCN,
author = {Liu, Congcong and Chen, Yuying and Liu, Ming and Shi, Bertram E.},
title = {AVGCN: Trajectory Prediction using Graph Convolutional Networks Guided by Human Attention},
year = {2021},
booktitle = {IEEE International Conference on Robotics and Automation },
pages = {14234–14240},
}

@article{DERGCN,
title = {DERGCN: Dynamic-Evolving graph convolutional networks for human trajectory prediction},
journal = {Neurocomputing},
volume = {569},
pages = {127117},
year = {2024},
issn = {0925-2312},
author = {Jing Mi and Xuxiu Zhang and Honghai Zeng and Lin Wang}
}

@article{GraphTern, 
title={A Set of Control Points Conditioned Pedestrian Trajectory Prediction}, volume={37}, 
number={5}, 
journal={Proceedings of the AAAI Conference on Artificial Intelligence}, author={Bae, Inhwan and Jeon, Hae-Gon}, 
year={2023}, 
month={06}, 
pages={6155-6165} }

@INPROCEEDINGS {Social-STGCNN,
author = {A. Mohamed and K. Qian and M. Elhoseiny and C. Claudel},
booktitle = {IEEE/CVF Conference on Computer Vision and Pattern Recognition},
title = {Social-STGCNN: A Social Spatio-Temporal Graph Convolutional Neural Network for Human Trajectory Prediction},
year = {2020},
volume = {},
issn = {},
pages = {14412-14420},

}

@INPROCEEDINGS{X2,
  author={Bhujel, Niraj and Yun, Yau Wei and Wang, Han and Dwivedi, Vijay Prakash},
  booktitle={IEEE/RSJ International Conference on Intelligent Robots and Systems}, 
  title={Self-critical Learning of Influencing Factors for Trajectory Prediction using Gated Graph Convolutional Network}, 
  year={2021},
  pages={7904-7910}}

@article{AC-VRNN,
title = {AC-VRNN: Attentive Conditional-VRNN for multi-future trajectory prediction},
journal = {Computer Vision and Image Understanding},
volume = {210},
pages = {103245},
year = {2021},
issn = {1077-3142},
author = {Alessia Bertugli and Simone Calderara and Pasquale Coscia and Lamberto Ballan and Rita Cucchiara}
}

@INPROCEEDINGS {EigenTrajectory,
author = {I. Bae and J. Oh and H. Jeon},
booktitle = {IEEE/CVF International Conference on Computer Vision},
title = {EigenTrajectory: Low-Rank Descriptors for Multi-Modal Trajectory Forecasting},
year = {2023},
volume = {},
issn = {},
pages = {9983-9995},
}

@article{IMP,
author = {Shi, Liushuai and Wang, Le and Long, Chengjiang and Zhou, Sanping and Tang, Wei and Zheng, Nanning and Hua, Gang},
title = {Representing Multimodal Behaviors With Mean Location for Pedestrian Trajectory Prediction},
year = {2023},
issue_date = {Sept. 2023},
publisher = {IEEE Computer Society},
address = {USA},
volume = {45},
number = {9},
issn = {0162-8828},
journal = {IEEE Trans. Pattern Anal. Mach. Intell.},
month = {09},
pages = {11184–11202},
numpages = {19}
}

@inproceedings{SocialVAE ,
   title={{SocialVAE}: Human Trajectory Prediction Using Timewise Latents},
   booktitle={European Conference on Computer Vision},
   publisher={Springer-Verlag},
   author={Xu, Pei and Hayet, Jean-Bernard and Karamouzas, Ioannis},
   year={2022},
   pages={511–528} }

@inproceedings{STAR,
  title={Spatio-temporal graph transformer networks for pedestrian trajectory prediction},
  author={Yu, Cunjun and Ma, Xiao and Ren, Jiawei and Zhao, Haiyu and Yi, Shuai},
  booktitle={European Conference on Computer Vision},
  pages={507--523},
  year={2020},
  organization={Springer}
}

@INPROCEEDINGS {AgentFormer,
author = {Y. Yuan and X. Weng and Y. Ou and K. Kitani},
booktitle = {IEEE/CVF International Conference on Computer Vision},
title = {AgentFormer: Agent-Aware Transformers for Socio-Temporal Multi-Agent Forecasting},
year = {2021},
volume = {},
issn = {},
pages = {9793-9803},
}

@InProceedings{PECNet,
author="Mangalam, Karttikeya
and Girase, Harshayu
and Agarwal, Shreyas
and Lee, Kuan-Hui
and Adeli, Ehsan
and Malik, Jitendra
and Gaidon, Adrien",
editor="Vedaldi, Andrea
and Bischof, Horst
and Brox, Thomas
and Frahm, Jan-Michael",
title="It Is Not the Journey But the Destination: Endpoint Conditioned Trajectory Prediction",
booktitle="European Conference on Computer Vision",
year="2020",
publisher="Springer International Publishing",
pages="759--776",
}

@INPROCEEDINGS {CIDNN,
author = {Y. Xu and Z. Piao and S. Gao},
booktitle = {IEEE/CVF Conference on Computer Vision and Pattern Recognition},
title = {Encoding Crowd Interaction with Deep Neural Network for Pedestrian Trajectory Prediction},
year = {2018},
volume = {},
issn = {},
pages = {5275-5284},

}

@inproceedings{MATRIX,
  title={MATRIX: Multi-Agent Trajectory Generation with Diverse Contexts},
  author={Xu, Zhuo and Zhou, Rui and Yin, Yida and Gao, Huidong and Tomizuka, Masayoshi and Li, Jiachen},
  booktitle={IEEE International Conference on Robotics and Automation},
  pages={12650--12657},
  year={2024},
}

@ARTICLE {SMEMO,
author = {F. Marchetti and F. Becattini and L. Seidenari and A. Bimbo},
journal = {IEEE Transactions on Pattern Analysis \& Machine Intelligence},
title = {SMEMO: Social Memory for Trajectory Forecasting},
year = {2024},
volume = {46},
number = {06},
issn = {1939-3539},
pages = {4410-4425},
}

@InProceedings{Social-SSL,
author="Tsao, Li-Wu
and Wang, Yan-Kai
and Lin, Hao-Siang
and Shuai, Hong-Han
and Wong, Lai-Kuan
and Cheng, Wen-Huang",
editor="Avidan, Shai
and Brostow, Gabriel
and Ciss{\'e}, Moustapha
and Farinella, Giovanni Maria
and Hassner, Tal",
title="Social-SSL: Self-supervised Cross-Sequence Representation Learning Based on Transformers for Multi-agent Trajectory Prediction",
year = {2022},
publisher = {Springer-Verlag},
booktitle = {European Conference on Computer Vision },
pages="234--250",
}

@ARTICLE{LADM,
  author={Lv, Kai and Yuan, Liang and Ni, Xiaoyu},
  journal={IEEE Transactions on Instrumentation and Measurement}, 
  title={Learning Autoencoder Diffusion Models of Pedestrian Group Relationships for Multimodal Trajectory Prediction}, 
  year={2024},
  volume={73},
  number={},
  pages={1-12}}

@inproceedings{GPGraph,
  title={Learning pedestrian group representations for multi-modal trajectory prediction},
  author={Bae, Inhwan and Park, Jin-Hwi and Jeon, Hae-Gon},
  booktitle={European Conference on Computer Vision},
  pages={270--289},
  year={2022},
  organization={Springer}
}

@INPROCEEDINGS {MG-GAN,
author = {P. Dendorfer and S. Elflein and L. Leal-Taixe},
booktitle = {IEEE/CVF International Conference on Computer Vision},
title = {MG-GAN: A Multi-Generator Model Preventing Out-of-Distribution Samples in Pedestrian Trajectory Prediction},
year = {2021},
volume = {},
issn = {},
pages = {13138-13147},
}

@INPROCEEDINGS {ScePT,
author = {Y. Chen and B. Ivanovic and M. Pavone},
booktitle = {Proceedings of the IEEE/CVF Conference on Computer Vision and Pattern Recognition},
title = {ScePT: Scene-consistent, Policy-based Trajectory Predictions for Planning},
year = {2022},
volume = {},
issn = {},
pages = {17082-17091},

}

@inproceedings{TPNMS,
  title={Temporal pyramid network for pedestrian trajectory prediction with multi-supervision},
  author={Liang, Rongqin and Li, Yuanman and Li, Xia and Tang, Yi and Zhou, Jiantao and Zou, Wenbin},
  booktitle={Proceedings of the AAAI Conference on Artificial Intelligence},
  number={3},
  pages={2029--2037},
  year={2021}
}

@INPROCEEDINGS{X6,
  author={Li, Kunming and Shan, Mao and Worrall, Stewart and Nebot, Eduardo},
  booktitle={IEEE International Conference on Intelligent Transportation Systems}, 
  title={Crowd Prediction and Autonomous Navigation with Partial Observations}, 
  year={2022},
  volume={},
  number={},
  pages={1829-1836}}

@inproceedings{SCAN,
  title={SCAN: A spatial context attentive network for joint multi-agent intent prediction},
  author={Sekhon, Jasmine and Fleming, Cody},
  booktitle={Proceedings of the AAAI Conference on Artificial Intelligence},
  number={7},
  pages={6119--6127},
  year={2021}
}

@inproceedings{LB-EBM,
  title={Trajectory Prediction with Latent Belief Energy-Based Model},
  author={Bo Pang and Tianyang Zhao and Xu Xie and Ying Nian Wu},
  booktitle={IEEE/CVF Conference on Computer Vision and Pattern Recognition},
  year={2021},
  pages={11809-11819},
}

@INPROCEEDINGS {SAnchor,
author = {P. Kothari and B. Sifringer and A. Alahi},
booktitle = {IEEE/CVF Conference on Computer Vision and Pattern Recognition},
title = {Interpretable Social Anchors for Human Trajectory Forecasting in Crowds},
year = {2021},
volume = {},
issn = {},
pages = {15551-15561},
}

@article{EPD,
  title={Modeling Pedestrian Intrinsic Uncertainty for Multimodal Stochastic Trajectory Prediction via Energy Plan Denoising},
  author={Liu, Yao and Sheng, Quan Z and Yao, Lina},
  journal={arXiv preprint arXiv:2405.07164},
  year={2024}
}

@article{SGANv2,
  title={Safety-compliant generative adversarial networks for human trajectory forecasting},
  author={Kothari, Parth and Alahi, Alexandre},
  journal={IEEE Transactions on Intelligent Transportation Systems},
  volume={24},
  number={4},
  pages={4251--4261},
  year={2023},
  publisher={IEEE}
}

@article{Group-Obstacle-LSTM,
title = {Embedding group and obstacle information in LSTM networks for human trajectory prediction in crowded scenes},
journal = {Computer Vision and Image Understanding},
volume = {203},
pages = {103126},
year = {2021},
issn = {1077-3142},
author = {Niccoló Bisagno and Cristiano Saltori and Bo Zhang and Francesco G.B. {De Natale} and Nicola Conci}
}

@inproceedings{Social-Implicit,
author = {Mohamed, Abduallah and Zhu, Deyao and Vu, Warren and Elhoseiny, Mohamed and Claudel, Christian},
title = {Social-Implicit: Rethinking Trajectory Prediction Evaluation and The Effectiveness of Implicit Maximum Likelihood Estimation},
year = {2022},
publisher = {Springer-Verlag},
booktitle = {European Conference on Computer Vision},
pages = {463–479},
}

@ARTICLE{STSF-Net,
  author={Wang, Yu and Chen, Shiwei},
  journal={IEEE Transactions on Multimedia}, 
  title={Multi-Agent Trajectory Prediction With Spatio-Temporal Sequence Fusion}, 
  year={2023},
  volume={25},
  number={},
  pages={13-23}}

@INPROCEEDINGS {X4,
author = {H. Zhao and R. P. Wildes},
booktitle = {IEEE/CVF International Conference on Computer Vision},
title = {Where are you heading? Dynamic Trajectory Prediction with Expert Goal Examples},
year = {2021},
volume = {},
issn = {},
pages = {7609-7618},
}

@ARTICLE{X5,
  author={Minoura, Hiroaki and Hirakawa, Tsubasa and Sugano, Yusuke and Yamashita, Takayoshi and Fujiyoshi, Hironobu},
  journal={IEEE Transactions on Intelligent Vehicles}, 
  title={Utilizing Human Social Norms for Multimodal Trajectory Forecasting via Group-Based Forecasting Module}, 
  year={2023},
  volume={8},
  number={1},
  pages={836-850}
}

@ARTICLE {SR-LSTM,
author = {P. Zhang and J. Xue and P. Zhang and N. Zheng and W. Ouyang},
journal = {IEEE Transactions on Pattern Analysis \& Machine Intelligence},
title = {Social-Aware Pedestrian Trajectory Prediction via States Refinement LSTM},
year = {2022},
volume = {44},
number = {05},
issn = {1939-3539},
pages = {2742-2759},
}

@ARTICLE{STGlow,
  author={Liang, Rongqin and Li, Yuanman and Zhou, Jiantao and Li, Xia},
  journal={IEEE Transactions on Neural Networks and Learning Systems}, 
  title={STGlow: A Flow-Based Generative Framework With Dual-Graphormer for Pedestrian Trajectory Prediction}, 
  year={2023},
  volume={35},
  number={11},
  pages={16504-16517}}

@article{FQA,
  title={Multi-agent trajectory prediction with fuzzy query attention},
  author={Kamra, Nitin and Zhu, Hao and Trivedi, Dweep Kumarbhai and Zhang, Ming and Liu, Yan},
  journal={Advances in Neural Information Processing Systems},
  volume={33},
  pages={22530--22541},
  year={2020}
}

@inproceedings{Trajectron++,
  title={Trajectron++: Dynamically-feasible trajectory forecasting with heterogeneous data},
  author={Salzmann, Tim and Ivanovic, Boris and Chakravarty, Punarjay and Pavone, Marco},
  booktitle={European Conference on Computer Vision},
  pages={683--700},
  year={2020},
  organization={Springer}
}

@article{Tri-HGNN,
  title={Tri-HGNN: Learning triple policies fused hierarchical graph neural networks for pedestrian trajectory prediction},
  author={Zhu, Wenjun and Liu, Yanghong and Wang, Peng and Zhang, Mengyi and Wang, Tian and Yi, Yang},
  journal={Pattern Recognition},
  volume={143},
  pages={109772},
  year={2023},
  publisher={Elsevier}
}

@article{GATraj,
title = {GATraj: A graph- and attention-based multi-agent trajectory prediction model},
journal = {ISPRS Journal of Photogrammetry and Remote Sensing},
volume = {205},
pages = {163-175},
year = {2023},
issn = {0924-2716},
author = {Hao Cheng and Mengmeng Liu and Lin Chen and Hellward Broszio and Monika Sester and Michael Ying Yang}
}

@ARTICLE{VIKT,
  author={Zhong, Xian and Yan, Xu and Yang, Zhengwei and Huang, Wenxin and Jiang, Kui and Liu, Ryan Wen and Wang, Zheng},
  journal={IEEE Transactions on Intelligent Transportation Systems}, 
  title={Visual Exposes You: Pedestrian Trajectory Prediction Meets Visual Intention}, 
  year={2023},
  volume={24},
  number={9},
  pages={9390-9400}}

@article{BR-GAN,
  title={BR-GAN: A pedestrian trajectory prediction model combined with behavior recognition},
  author={Pang, Shu Min and Cao, Jin Xin and Jian, Mei Ying and Lai, Jian and Yan, Zhen Ying},
  journal={IEEE Transactions on Intelligent Transportation Systems},
  volume={23},
  number={12},
  pages={24609--24620},
  year={2022},
  publisher={IEEE}
}

@article{CSCNet,
  title={CSCNet: Contextual semantic consistency network for trajectory prediction in crowded spaces},
  author={Xia, Beihao and Wong, Conghao and Peng, Qinmu and Yuan, Wei and You, Xinge},
  journal={Pattern Recognition},
  volume={126},
  pages={108552},
  year={2022},
  publisher={Elsevier}
}

@article{CTSGI,
  title={Causal temporal-spatial pedestrian trajectory prediction with goal point estimation and contextual interaction},
  author={Lian, Jing and Yu, Fengning and Li, Linhui and Zhou, Yafu},
  journal={IEEE Transactions on Intelligent Transportation Systems},
  volume={23},
  number={12},
  pages={24499--24509},
  year={2022},
  publisher={IEEE}
}

@inproceedings{DROGON,
  title={DROGON: A trajectory prediction model based on intention-conditioned behavior reasoning},
  author={Choi, Chiho and Malla, Srikanth and Patil, Abhishek and Choi, Joon Hee},
  booktitle={Conference on Robot Learning},
  pages={49--63},
  year={2021},
  organization={PMLR}
}

@inproceedings{GAIN,
  title={Multi-agent trajectory prediction with graph attention isomorphism neural network},
  author={Liu, Yongkang and Qi, Xuewei and Sisbot, Emrah Akin and Oguchi, Kentaro},
  booktitle={IEEE Intelligent Vehicles Symposium},
  pages={273--279},
  year={2022},
}

@article{HST,
  title={Robots that can see: Leveraging human pose for trajectory prediction},
  author={Salzmann, Tim and Chiang, Hao-Tien Lewis and Ryll, Markus and Sadigh, Dorsa and Parada, Carolina and Bewley, Alex},
  journal={IEEE Robotics and Automation Letters},
  year={2023},
  volume={8},
  number={11},
  pages={7090-7097},
  publisher={IEEE}
}

@inproceedings{Hypertron,
  title={Hypertron: Explicit Social-Temporal Hypergraph Framework for Multi-Agent Forecasting.},
  author={Tian, Yu and Huang, Xingliang and Niu, Ruigang and Yu, Hongfeng and Wang, Peijin and Sun, Xian},
  booktitle={IJCAI},
  pages={1356--1362},
  year={2022}
}

@article{LSSTA,
  title={Long-short term spatio-temporal aggregation for trajectory prediction},
  author={Yang, Cuiliu and Pei, Zhao},
  journal={IEEE Transactions on Intelligent Transportation Systems},
  volume={24},
  number={4},
  pages={4114--4126},
  year={2023},
  publisher={IEEE}
}

@article{Meta-IRLSOT++,
  title={Meta-IRLSOT++: A meta-inverse reinforcement learning method for fast adaptation of trajectory prediction networks},
  author={Yang, Biao and Lu, Yanan and Wan, Rui and Hu, Hongyu and Yang, Changchun and Ni, Rongrong},
  journal={Expert Systems with Applications},
  volume={240},
  pages={122499},
  year={2024},
  publisher={Elsevier}
}

@inproceedings{NMMP,
  title={Collaborative motion prediction via neural motion message passing},
  author={Hu, Yue and Chen, Siheng and Zhang, Ya and Gu, Xiao},
  booktitle={IEEE/CVF Conference on Computer Vision and Pattern Recognition},
  pages={6319--6328},
  year={2020}
}

@inproceedings{SocialCVAE,
  title={SocialCVAE: Predicting Pedestrian Trajectory via Interaction Conditioned Latents},
  author={Xiang, Wei and Haoteng, YIN and Wang, He and Jin, Xiaogang},
  booktitle={Proceedings of the AAAI Conference on Artificial Intelligence},
  number={6},
  pages={6216--6224},
  year={2024}
}

@inproceedings{evolvegraph,
author = {Li, Jiachen and Yang, Fan and Tomizuka, Masayoshi and Choi, Chiho},
title = {EvolveGraph: multi-agent trajectory prediction with dynamic relational reasoning},
year = {2020},
booktitle = {Proceedings of the International Conference on Neural Information Processing Systems},
pages     = {19783--19794},
articleno = {1660},
numpages = {12},
location = {Vancouver, BC, Canada},
series = {NIPS '20}
}

@article{EvolveHypergraph,
  title={Evolvehypergraph: Group-aware dynamic relational reasoning for trajectory prediction},
  author={Li, Jiachen and Hua, Chuanbo and Park, Jinkyoo and Ma, Hengbo and Dax, Victoria and Kochenderfer, Mykel J},
  journal={arXiv preprint arXiv:2208.05470},
  year={2022}
}

@inproceedings{SSA-GAN,
  title={Social-scene-aware generative adversarial networks for pedestrian trajectory prediction},
  author={Huang, Binhao and Ma, Zhenwei and Chen, Lianggangxu and He, Gaoqi},
  booktitle={Computer Graphics International Conference},
  pages={190--201},
  year={2021},
  organization={Springer},
    }

@ARTICLE{GA-STT,
  author={Zhou, Lei and Yang, Dingye and Zhai, Xiaolin and Wu, Shichao and Hu, ZhengXi and Liu, Jingtai},
  journal={IEEE Robotics and Automation Letters}, 
  title={GA-STT: Human Trajectory Prediction With Group Aware Spatial-Temporal Transformer}, 
  year={2022},
  volume={7},
  number={3},
  pages={7660-7667}}

@inproceedings{Transformer2017,
 author = {Vaswani, Ashish and Shazeer, Noam and Parmar, Niki and Uszkoreit, Jakob and Jones, Llion and Gomez, Aidan N and Kaiser, \L ukasz and Polosukhin, Illia},
 booktitle = {Advances in Neural Information Processing Systems},
 editor = {I. Guyon and U. Von Luxburg and S. Bengio and H. Wallach and R. Fergus and S. Vishwanathan and R. Garnett},
 pages = {6000--6010},
 title = {Attention is All you Need},
 year = {2017}
}

@inproceedings{LMTraj,
  title={{Can language beat numerical regression? Language-based multimodal trajectory prediction}},
  author={Bae, Inhwan and Lee, Junoh and Jeon, Hae-Gon},
  booktitle={IEEE/CVF Conference on Computer Vision and Pattern Recognition},
  pages={753--766},
  year={2024}
}

@inproceedings{SGCN,
  title={SGCN: Sparse graph convolution network for pedestrian trajectory prediction},
  author={Shi, Liushuai and Wang, Le and Long, Chengjiang and Zhou, Sanping and Zhou, Mo and Niu, Zhenxing and Hua, Gang},
  booktitle={IEEE/CVF conference on Computer Vision and Pattern Recognition},
  pages={8994--9003},
  year={2021}
}

@article{X11,
  title={Multi-agent dynamic relational reasoning for social robot navigation},
  author={Li, Jiachen and Hua, Chuanbo and Ma, Hengbo and Park, Jinkyoo and Dax, Victoria and Kochenderfer, Mykel J},
  journal={arXiv preprint arXiv:2401.12275},
  year={2024}
}

@article{SSALVM,
  title={Context-aware human trajectories prediction via latent variational model},
  author={Berenguer, Abel D{\'\i}az and Alioscha-Perez, Mitchel and Oveneke, Meshia Cedric and Sahli, Hichem},
  journal={IEEE Transactions on Circuits and Systems for Video Technology},
  volume={31},
  number={5},
  pages={1876--1889},
  year={2020},
  publisher={IEEE}
}

@inproceedings{SSP,
  title={{SSP}: Single shot future trajectory prediction},
  author={Dwivedi, Isht and Malla, Srikanth and Dariush, Behzad and Choi, Chiho},
  booktitle={IEEE/RSJ International Conference on Intelligent Robots and Systems},
  pages={2211--2218},
  year={2020},
}

@inproceedings{STGT,
  title={STGT: Forecasting pedestrian motion using spatio-temporal graph transformer},
  author={Syed, Arsal and Morris, Brendan},
  booktitle={IEEE Intelligent Vehicles Symposium},
  pages={1553--1558},
  year={2021},
}

@article{STHGLU,
  title={An efficient spatial--temporal model based on gated linear units for trajectory prediction},
  author={Liu, Shaohua and Wang, Yisu and Sun, Jingkai and Mao, Tianlu},
  journal={Neurocomputing},
  volume={492},
  pages={593--600},
  year={2022},
  publisher={Elsevier}
}

@inproceedings{UEN,
  title={A Unified Environmental Network for Pedestrian Trajectory Prediction},
  author={Su, Yuchao and Li, Yuanman and Wang, Wei and Zhou, Jiantao and Li, Xia},
  booktitle={Proceedings of the AAAI Conference on Artificial Intelligence},
  number={5},
  pages={4970--4978},
  year={2024}
}

@inproceedings{X13,
  title={Reciprocal learning networks for human trajectory prediction},
  author={Sun, Hao and Zhao, Zhiqun and He, Zhihai},
  booktitle={IEEE/CVF Conference on Computer Vision and Pattern Recognition},
  pages={7416--7425},
  year={2020}
}

@inproceedings{X14,
  title={Conditioned human trajectory prediction using iterative attention blocks},
  author={Postnikov, Aleksey and Gamayunov, Aleksander and Ferrer, Gonzalo},
  booktitle={IEEE International Conference on Robotics and Automation},
  pages={4599--4604},
  year={2022},
}

@article{X16,
  title={Semantic scene upgrades for trajectory prediction},
  author={Syed, Arsal and Morris, Brendan Tran},
  journal={Machine vision and applications},
  volume={34},
  number={2},
  pages={23},
  year={2023},
  publisher={Springer}
}

@article{FLEAM,
  title={Fully convolutional encoder-decoder with an attention mechanism for practical pedestrian trajectory prediction},
  author={Chen, Kai and Song, Xiao and Yuan, Haitao and Ren, Xiaoxiang},
  journal={IEEE Transactions on Intelligent Transportation Systems},
  volume={23},
  number={11},
  pages={20046--20060},
  year={2022},
  publisher={IEEE}
}

@article{SPU-BERT,
  title={SPU-BERT: Faster human multi-trajectory prediction from socio-physical understanding of BERT},
  author={Na, Ki-In and Kim, Ue-Hwan and Kim, Jong-Hwan},
  journal={Knowledge-Based Systems},
  volume={274},
  pages={110637},
  year={2023},
  publisher={Elsevier}
}

@inproceedings{MSRL,
  title={Multi-stream representation learning for pedestrian trajectory prediction},
  author={Wu, Yuxuan and Wang, Le and Zhou, Sanping and Duan, Jinghai and Hua, Gang and Tang, Wei},
  booktitle={Proceedings of the AAAI Conference on Artificial Intelligence},
  number={3},
  pages={2875--2882},
  year={2023}
}

@article{SBD,
  title={A Synchronous Bi-Directional Framework With Temporally Dependent Interaction Modeling for Pedestrian Trajectory Prediction},
  author={Li, Yuanman and Xie, Ce and Liang, Rongqin and Du, Jie and Zhou, Jiantao and Li, Xia},
  journal={IEEE Transactions on Network Science and Engineering},
  year={2023},
  volume={11},
  number={1},
  pages={793-806},
  publisher={IEEE}
}

@inproceedings{BCDiff,
 author = {Li, Rongqing and Li, Changsheng and Ren, Dongchun and Chen, Guangyi and Yuan, Ye and Wang, Guoren},
 booktitle = {Advances in Neural Information Processing Systems},
 editor = {A. Oh and T. Naumann and A. Globerson and K. Saenko and M. Hardt and S. Levine},
 pages = {14400--14413},
 title = {BCDiff: Bidirectional Consistent Diffusion for Instantaneous Trajectory Prediction},
 volume = {36},
 year = {2023}
}

@article{Sparse-GAMPS,
  title={Exploring social posterior collapse in variational autoencoder for interaction modeling},
  author={Tang, Chen and Zhan, Wei and Tomizuka, Masayoshi},
  journal={Advances in Neural Information Processing Systems},
  volume={34},
  pages={8481--8494},
  year={2021}
}

@inproceedings{Leapfrog,
  title={Leapfrog diffusion model for stochastic trajectory prediction},
  author={Mao, Weibo and Xu, Chenxin and Zhu, Qi and Chen, Siheng and Wang, Yanfeng},
  booktitle={IEEE/CVF Conference on Computer Vision and Pattern Recognition},
  pages={5517--5526},
  year={2023}
}

@article{PTP-STGCN,
  title={Ptp-stgcn: pedestrian trajectory prediction based on a spatio-temporal graph convolutional neural network},
  author={Lian, Jing and Ren, Weiwei and Li, Linhui and Zhou, Yafu and Zhou, Bin},
  journal={Applied Intelligence},
  volume={53},
  number={3},
  pages={2862--2878},
  year={2023},
  publisher={Springer}
}

@inproceedings{ForceFormer,
  title={ForceFormer: exploring social force and transformer for pedestrian trajectory prediction},
  author={Zhang, Weicheng and Cheng, Hao and Johora, Fatema T and Sester, Monika},
  booktitle={ IEEE Intelligent Vehicles Symposium },
  pages={1--7},
  year={2023},
}

@inproceedings{TUTR,
  title={Trajectory unified transformer for pedestrian trajectory prediction},
  author={Shi, Liushuai and Wang, Le and Zhou, Sanping and Hua, Gang},
  booktitle={IEEE/CVF International Conference on Computer Vision},
  pages={9675--9684},
  year={2023}
}

@article{X26,
  title={Another vertical view: A hierarchical network for heterogeneous trajectory prediction via spectrums},
  author={Xia, Beihao and Wong, Conghao and Xu, Duanquan and Peng, Qinmu and You, Xinge},
  journal={IEEE Transactions on Pattern Analysis and Machine Intelligence},
  year={2025},
}

@article{PPNet,
  title={Multimodal pedestrian trajectory prediction using probabilistic proposal network},
  author={Chen, Weihuang and Yang, Zhigang and Xue, Lingyang and Duan, Jinghai and Sun, Hongbin and Zheng, Nanning},
  journal={IEEE Transactions on Circuits and Systems for Video Technology},
  volume={33},
  number={6},
  pages={2877--2891},
  year={2022},
  publisher={IEEE}
}

@inproceedings{GTP-Force,
  title={GTP-Force: Game-Theoretic Trajectory Prediction through Distributed Reinforcement Learning},
  author={Emami, Negar and Di Maio, Antonio and Braun, Torsten},
  booktitle={IEEE International Conference on Mobile Ad Hoc and Smart Systems},
  pages={234--242},
  year={2023},
}

@article{SRGAT,
  title={Goal-Guided and Interaction-Aware State Refinement Graph Attention Network for Multi-Agent Trajectory Prediction},
  author={Chen, Xiaobo and Luo, Fengbo and Zhao, Feng and Ye, Qiaolin},
  journal={IEEE Robotics and Automation Letters},
  volume={9},
  number={1},
  pages={57--64},
  year={2023},
  publisher={IEEE}
}

@inproceedings{GroupNet,
  title={Groupnet: Multiscale hypergraph neural networks for trajectory prediction with relational reasoning},
  author={Xu, Chenxin and Li, Maosen and Ni, Zhenyang and Zhang, Ya and Chen, Siheng},
  booktitle={IEEE/CVF Conference on Computer Vision and Pattern Recognition},
  pages={6498--6507},
  year={2022}
}

@article{Review_pedestrian,
  title={Review of pedestrian trajectory prediction methods: Comparing deep learning and knowledge-based approaches},
  author={Korbmacher, Raphael and Tordeux, Antoine},
  journal={IEEE Transactions on Intelligent Transportation Systems},
  volume={23},
  number={12},
  pages={24126--24144},
  year={2022},
 }

@inproceedings{ETH,
  title={You'll never walk alone: Modeling social behavior for multi-target tracking},
  author={Pellegrini, Stefano and Ess, Andreas and Schindler, Konrad and Van Gool, Luc},
  booktitle={IEEE International Conference on Computer Vision},
  pages={261--268},
  year={2009},
}

@inproceedings{MART,
  title={MART: MultiscAle Relational Transformer Networks for Multi-agent Trajectory Prediction},
  author={Lee, Seongju and Lee, Junseok and Yu, Yeonguk and Kim, Taeri and Lee, Kyoobin},
  booktitle={European Conference on Computer Vision},
  pages={89--107},
  year={2025},
  organization={Springer}
}

@inproceedings{NBA,
  title={Multi-modal trajectory prediction of {NBA} players},
  author={Hauri, Sandro and Djuric, Nemanja and Radosavljevic, Vladan and Vucetic, Slobodan},
  booktitle={IEEE/CVF Winter Conference on Applications of Computer Vision},
  pages={1640--1649},
  year={2021}
}

@inproceedings{weng2023joint,
      title={Joint Metrics Matter: A Better Standard for Trajectory Forecasting}, 
      author={Weng, Erica and Hoshino, Hana and Ramanan, Deva and Kitani, Kris},
      booktitle={IEEE/CVF International Conference on Computer Vision },
    pages={20315--20326},
      year={2023}
}

@article{monosurvey,
  title={Pedestrian trajectory prediction in pedestrian-vehicle mixed environments: A systematic review},
  author={Golchoubian, Mahsa and Ghafurian, Moojan and Dautenhahn, Kerstin and Azad, Nasser Lashgarian},
  journal={IEEE Transactions on Intelligent Transportation Systems},
  year={2023},
  volume={24},
  number={11},
  pages={11544-11567},
  publisher={IEEE}
}

@inproceedings{waymo,
  title={Scalability in perception for autonomous driving: Waymo open dataset},
  author={Sun, Pei and Kretzschmar, Henrik and Dotiwalla, Xerxes and Chouard, Aurelien and Patnaik, Vijaysai and Tsui, Paul and Guo, James and Zhou, Yin and Chai, Yuning and Caine, Benjamin and others},
  booktitle={IEEE/CVF Conference on Computer Vision and Pattern Recognition},
  pages={2446--2454},
  year={2020}
}

@article{Human--robot,
  title={Human--robot interaction: a survey},
  author={Goodrich, Michael A and Schultz, Alan C and others},
  journal={Foundations and Trends{\textregistered} in Human--Computer Interaction},
  volume={1},
  number={3},
  pages={203--275},
  year={2008},
  publisher={Now Publishers, Inc.}
}

@inproceedings{CARLA,
  title={OpenScenario: a flexible integrated environment to develop educational activities based on pedagogical scenarios},
  author={Jullien, Jean-Michel and Martel, Christian and Vignollet, Laurence and Wentland, Maia},
  booktitle={2009 Ninth IEEE International Conference on Advanced Learning Technologies},
  pages={509--513},
  year={2009},
  organization={IEEE}
}

@article{GRU,
  title={Empirical evaluation of gated recurrent neural networks on sequence modeling},
  author={Chung, Junyoung and Gulcehre, Caglar and Cho, KyungHyun and Bengio, Yoshua},
  journal={arXiv preprint arXiv:1412.3555},
  year={2014}
}

@article{rnn,
  title={Learning representations by back-propagating errors},
  author={Rumelhart, David E and Hinton, Geoffrey E and Williams, Ronald J},
  journal={Nature},
  volume={323},
  number={6088},
  pages={533--536},
  year={1986},
  publisher={Nature Publishing Group UK London}
}

@article{vae,
  title={Auto-encoding variational bayes},
  author={Kingma, Diederik P and Welling, Max},
  journal={arXiv preprint arXiv:1312.6114},
  year={2013}
}

@article{Kun2024,
title = "Evacuation trajectory prediction of passengers in transport aircraft based on social-implicit model",
journal = "Journal of Traffic and Transportation Engineering",
volume = {24},
number = {1671-1637},
pages = {270},
year = {2024},
issn = {1671-1637},
author = {CHEN Kun and LI Fang and FENG Zhen-yu and CHEN Xiang-ming and DUAN Long-kun},
}

@article{Survey_crowd_modelling,
  title={Data-driven crowd modeling techniques: A survey},
  author={Zhong, Jinghui and Li, Dongrui and Huang, Zhixing and Lu, Chengyu and Cai, Wentong},
  journal={ACM Transactions on Modeling and Computer Simulation },
  volume={32},
  number={1},
  pages={1--33},
  year={2022},
  publisher={ACM New York, NY}
}

@ARTICLE{retail,

  author={Galdelli, Alessandro and Pietrini, Rocco and Mancini, Adriano and Zingaretti, Primo},

  journal={IEEE Access}, 

  title={Retail Robot Navigation: A Shopper Behavior-Centric Approach to Path Planning}, 

  year={2024},
  

  volume={12},

  number={},

  pages={50154-50164} }

@Inproceedings{IvanovicPavone2019,
  author       = {Ivanovic, B. and Pavone, M.},
  title        = {The {Trajectron}: Probabilistic Multi-Agent Trajectory Modeling with Dynamic Spatiotemporal Graphs},
  booktitle    = {{IEEE/CVF International Conference on Computer Vision}},
   pages={2375-2381},
  year         = {2019},
}

@article{parzen1962,
    title = "On Estimation of a Probability Density Function and Mode",
    author = "Emanuel Parzen",
    journal = "The Annals of Mathematical Statistics",
    volume = "33",
    pages = "1065-1076",
    year = "1962"
}

@article{vanToll2021Algorithms,
    author = {van Toll, W. and Pettr\'{e}, J.},
    title = {{Algorithms for Microscopic Crowd Simulation: Advancements in the 2010s}},
    journal = {Computer Graphics Forum},
    volume = {40},
    number = {2},
    pages = {731--754},
    year = {2021}
}

@article{entropy1,
  title={A statistical similarity measure for aggregate crowd dynamics},
  author={Guy, Stephen J and Van Den Berg, Jur and Liu, Wenxi and Lau, Rynson and Lin, Ming C and Manocha, Dinesh},
  journal={ACM Transactions on Graphics },
  volume={31},
  number={6},
  pages={190},
  year={2012},
}

@article{entropy2,
  title={Crowd Space: A Predictive Crowd Analysis Technique},
  author={Ioannis Karamouzas and Nick Sohre and Ran Hu and Stephen J. Guy},
  journal = {ACM Transactions on Graphics},
 volume = {37},
 number = {6},
 year = {2018},
 pages = {1--14}}

@article{scholler2020constant,
  title={What the constant velocity model can teach us about pedestrian motion prediction},
  author={Sch{\"o}ller, Christoph and Aravantinos, Vincent and Lay, Florian and Knoll, Alois},
  journal={IEEE Robotics and Automation Letters},
  volume={5},
  number={2},
  pages={1696--1703},
  year={2020},
}

@INPROCEEDINGS{argoverse,
  author={Chang, Ming-Fang and Lambert, John and Sangkloy, Patsorn and Singh, Jagjeet and Bak, Slawomir and Hartnett, Andrew and Wang, De and Carr, Peter and Lucey, Simon and Ramanan, Deva and Hays, James},
  booktitle={IEEE/CVF Conference on Computer Vision and Pattern Recognition}, 
  title={Argoverse: 3D Tracking and Forecasting With Rich Maps}, 
  year={2019},
  volume={},
  number={},
  pages={8740-8749}
}

@inproceedings{a2x,
author = {Sohn, Samuel S. and Lee, Mihee and Moon, Seonghyeon and Qiao, Gang and Usman, Muhammad and Yoon, Sejong and Pavlovic, Vladimir and Kapadia, Mubbasir},
title = {A2X: An Agent and Environment Interaction Benchmark for Multimodal Human Trajectory Prediction},
year = {2021},
booktitle = {Proceedings of the 14th ACM SIGGRAPH Conference on Motion, Interaction and Games},
articleno = {19},
numpages = {9},
pages={1--9}
}

@inproceedings{charalambous2014data,
  title={A data-driven framework for visual crowd analysis},
  author={Charalambous, Panayiotis and Karamouzas, Ioannis and Guy, Stephen J and Chrysanthou, Yiorgos},
  booktitle={Computer Graphics Forum},
  pages={41--50},
  year={2014},
}

@article{he2020informative,
  title={Informative scene decomposition for crowd analysis, comparison and simulation guidance},
  author={He, Feixiang and Xiang, Yuanhang and Zhao, Xi and Wang, He},
  journal={ACM Transactions on Graphics },
  volume={39},
  number={4},
  pages={50--1},
  year={2020},
}

@article{huang2022survey,
  title={A survey on trajectory-prediction methods for autonomous driving},
  author={Huang, Yanjun and Du, Jiatong and Yang, Ziru and Zhou, Zewei and Zhang, Lin and Chen, Hong},
  journal={IEEE Transactions on Intelligent Vehicles},
  volume={7},
  number={3},
  pages={652--674},
  year={2022},
  publisher={IEEE}
}

@article{core,
  title={Core challenges of social robot navigation: A survey},
  author={Mavrogiannis, Christoforos and Baldini, Francesca and Wang, Allan and Zhao, Dapeng and Trautman, Pete and Steinfeld, Aaron and Oh, Jean},
  journal={ACM Transactions on Human-Robot Interaction},
  volume={12},
  number={3},
  pages={1--39},
  year={2023},
  publisher={ACM New York, NY}
}

@article{magni,
  title={TH{\"O}R-MAGNI: a large-scale indoor motion capture recording of human movement and robot interaction},
  author={Schreiter, Tim and Rodrigues de Almeida, Tiago and Zhu, Yufei and Gutierrez Maestro, Eduardo and Morillo-Mendez, Lucas and Rudenko, Andrey and Palmieri, Luigi and Kucner, Tomasz P and Magnusson, Martin and Lilienthal, Achim J},
  journal={The International Journal of Robotics Research},
  volume={44},
  number={4},
  pages={568--591},
  year={2025},
  publisher={Sage Publications Sage UK: London, England}
}

@article{muchen2024mixed,
  title={Mixed strategy Nash equilibrium for crowd navigation},
  author={Muchen Sun, Max and Baldini, Francesca and Hughes, Katie and Trautman, Peter and Murphey, Todd},
  journal={The International Journal of Robotics Research},
  pages={02783649241302342},
  year={2024},
  publisher={SAGE Publications Sage UK: London, England}
}

@article{trautman2015robot,
  title={Robot navigation in dense human crowds: Statistical models and experimental studies of human--robot cooperation},
  author={Trautman, Pete and Ma, Jeremy and Murray, Richard M and Krause, Andreas},
  journal={The International Journal of Robotics Research},
  volume={34},
  number={3},
  pages={335--356},
  year={2015},
  publisher={SAGE Publications Sage UK: London, England}
}

@article{zhao2020noticing,
  title={Noticing motion patterns: A temporal cnn with a novel convolution operator for human trajectory prediction},
  author={Zhao, Dapeng and Oh, Jean},
  journal={IEEE Robotics and Automation Letters},
  volume={6},
  number={2},
  pages={628--634},
  year={2020},
  publisher={IEEE}
}

@inproceedings{ind,
  title={The ind dataset: A drone dataset of naturalistic road user trajectories at german intersections},
  author={Bock, Julian and Krajewski, Robert and Moers, Tobias and Runde, Steffen and Vater, Lennart and Eckstein, Lutz},
  booktitle={IEEE Intelligent Vehicles Symposium (IV)},
  pages={1929--1934},
  year={2020},
}

@inproceedings{round,
  title={The round dataset: A drone dataset of road user trajectories at roundabouts in germany},
  author={Krajewski, Robert and Moers, Tobias and Bock, Julian and Vater, Lennart and Eckstein, Lutz},
  booktitle={IEEE International Conference on Intelligent Transportation Systems},
  pages={1--6},
  year={2020},
}

@inproceedings{sdd,
  title={Learning social etiquette: Human trajectory understanding in crowded scenes},
  author={Robicquet, Alexandre and Sadeghian, Amir and Alahi, Alexandre and Savarese, Silvio},
  booktitle={European conference on computer vision},
  pages={549--565},
  year={2016},
  organization={Springer}
}

@inproceedings{gcs,
  title={Understanding collective crowd behaviors: Learning a mixture model of dynamic pedestrian-agents},
  author={Zhou, Bolei and Wang, Xiaogang and Tang, Xiaoou},
  booktitle={IEEE/CVF conference on Computer Vision and Pattern Recognition},
  pages={2871--2878},
  year={2012},
  organization={IEEE}
}

@inproceedings{town,
  title={Stable multi-target tracking in real-time surveillance video},
  author={Benfold, Ben and Reid, Ian},
  booktitle={IEEE/CVF Conference on Computer Vision and Pattern Recognition},
  pages={3457--3464},
  year={2011},
}

@article{atc,
  title={Person tracking in large public spaces using 3-D range sensors},
  author={Br{\v{s}}{\v{c}}i{\'c}, Dra{\v{z}}en and Kanda, Takayuki and Ikeda, Tetsushi and Miyashita, Takahiro},
  journal={IEEE Transactions on Human-Machine Systems},
  volume={43},
  number={6},
  pages={522--534},
  year={2013},
  publisher={IEEE}
}

@inproceedings{daimier,
  title={Pedestrian path prediction with recursive bayesian filters: A comparative study},
  author={Schneider, Nicolas and Gavrila, Dariu M},
  booktitle={German conference on pattern recognition},
  pages={174--183},
  year={2013},
  organization={Springer}
}

@inproceedings{lcas,
  title={Online learning for human classification in 3D LiDAR-based tracking},
  author={Yan, Zhi and Duckett, Tom and Bellotto, Nicola},
  booktitle={IEEE/RSJ International Conference on Intelligent Robots and Systems},
  pages={864--871},
  year={2017},
}

@article{trajimpute,
  title={Pedestrian trajectory prediction with missing data: Datasets, imputation, and benchmarking},
  author={Chib, Pranav Singh and Singh, Pravendra},
  journal={Advances in Neural Information Processing Systems},
  volume={37},
  pages={124530--124546},
  year={2024}
}

@article{lights,
  title={Dense Crowd Dynamics and Pedestrian Trajectories: A Multiscale Field Dataset from the Festival of Lights in Lyon},
  author={Dufour, Oscar and Dang, Huu-Tu and Cordes, Jakob and Korbmacher, Raphael and Chraibi, Mohcine and Gaudou, Benoit and Nicolas, Alexandre and Tordeux, Antoine},
  journal={Scientific Data},
  volume={12},
  number={1},
  pages={718},
  year={2025},
  publisher={Nature Publishing Group UK London}
}

@article{sit,
  title={Sit dataset: socially interactive pedestrian trajectory dataset for social navigation robots},
  author={Bae, Jong Wook and Kim, Jungho and Yun, Junyong and Kang, Changwon and Choi, Jeongseon and Kim, Chanhyeok and Lee, Junho and Choi, Jungwook and Choi, Jun Won},
  journal={Advances in neural information processing systems},
  volume={36},
  pages={24552--24563},
  year={2023}
}

@article{jrdb,
  title={Jrdb: A dataset and benchmark of egocentric robot visual perception of humans in built environments},
  author={Martin-Martin, Roberto and Patel, Mihir and Rezatofighi, Hamid and Shenoi, Abhijeet and Gwak, JunYoung and Frankel, Eric and Sadeghian, Amir and Savarese, Silvio},
  journal={IEEE transactions on pattern analysis and machine intelligence},
  volume={45},
  number={6},
  pages={6748--6765},
  year={2021},
  publisher={IEEE}
}

@inproceedings{jrdbact,
  title={Jrdb-act: A large-scale dataset for spatio-temporal action, social group and activity detection},
  author={Ehsanpour, Mahsa and Saleh, Fatemeh and Savarese, Silvio and Reid, Ian and Rezatofighi, Hamid},
  booktitle={IEEE/CVF Conference on Computer Vision and Pattern Recognition},
  pages={20983--20992},
  year={2022}
}

@article{oxfordihm,
  title={Motion planning in dynamic environments using context-aware human trajectory prediction},
  author={Finean, Mark Nicholas and Petrovi{\'c}, Luka and Merkt, Wolfgang and Markovi{\'c}, Ivan and Havoutis, Ioannis},
  journal={Robotics and autonomous systems},
  volume={166},
  pages={104450},
  year={2023},
  publisher={Elsevier}
}

@inproceedings{magniact,
  title={TH{\"O}R-MAGNI Act: Actions for Human Motion Modeling in Robot-Shared Industrial Spaces},
  author={De Almeida, Tiago Rodrigues and Schreiter, Tim and Rudenko, Andrey and Palmieri, Luigi and Stork, Johannes A and Lilienthal, Achim J},
  booktitle={ACM/IEEE International Conference on Human-Robot Interaction},
  pages={1083--1087},
  year={2025},
}

@article{thor,
  title={Th{\"o}r: Human-robot navigation data collection and accurate motion trajectories dataset},
  author={Rudenko, Andrey and Kucner, Tomasz P and Swaminathan, Chittaranjan S and Chadalavada, Ravi T and Arras, Kai O and Lilienthal, Achim J},
  journal={IEEE Robotics and Automation Letters},
  volume={5},
  number={2},
  pages={676--682},
  year={2020},
  publisher={IEEE}
}

@inproceedings{worldtrace,
  title={en},
  author={Zhu, Yuanshao and Yu, James Jianqiao and Zhao, Xiangyu and Zhou, Xun and Han, Liang and Wei, Xuetao and Liang, Yuxuan},
  booktitle={Proceedings of Advances in Neural Information Processing Systems},
  volume={38},
  year={2025}
}

@inproceedings{desire,
  title={Desire: Distant future prediction in dynamic scenes with interacting agents},
  author={Lee, Namhoon and Choi, Wongun and Vernaza, Paul and Choy, Christopher B and Torr, Philip HS and Chandraker, Manmohan},
  booktitle={IEEE/CVF Conference on Computer Vision and Pattern Recognition},
  pages={336--345},
  year={2017}
}

@inproceedings{NLL,
  title={Long-term on-board prediction of people in traffic scenes under uncertainty},
  author={Bhattacharyya, Apratim and Fritz, Mario and Schiele, Bernt},
  booktitle={IEEE/CVF Conference on Computer Vision and Pattern Recognition},
  pages={4194--4202},
  year={2018}
}

@article{auc,
  title={Beyond minimum-of-N: Rethinking the evaluation and methods of pedestrian trajectory prediction},
  author={Li, Linhui and Lin, Xiaotong and Huang, Yejia and Zhang, Zizhen and Hu, Jian-Fang},
  journal={IEEE Transactions on Circuits and Systems for Video Technology},
  year={2024},
  volume={34},
  number={12},
  pages={12880-12893},
  publisher={IEEE}
}

@article{DSTIGCN,
  title={DSTIGCN: Deformable Spatial-Temporal Interaction Graph Convolution Network for Pedestrian Trajectory Prediction},
  author={Chen, Wangxing and Sang, Haifeng and Wang, Jinyu and Zhao, Zishan},
  journal={IEEE Transactions on Intelligent Transportation Systems},
  year={2025},
  publisher={IEEE}
}

@article{MHTraj:,
  title={MHTraj: A Multi-Domain Hybrid Graph Neural Network With Causal-Spatial Modeling for Multi-Agent Trajectory Prediction},
  author={Chen, Jiuyu and Jia, Chunxiao and Xie, Wei and Zhu, Donglin and Shao, Xiaotao and Wang, Zhongli},
  journal={IEEE Transactions on Network Science and Engineering},
  year={2025},
  publisher={IEEE}
}

@article{MSWTE-GNN,
  title={Multi-scale wavelet transform enhanced graph neural network for pedestrian trajectory prediction},
  author={Lin, Xuanqi and Zhang, Yong and Wang, Shun and Hu, Yongli and Yin, Baocai},
  journal={Physica A: Statistical Mechanics and its Applications},
  volume={659},
  pages={130319},
  year={2025},
  publisher={Elsevier}
}

@article{OST-HGCN,
  title={OST-HGCN: Optimized Spatial--Temporal Hypergraph Convolution Network for Trajectory Prediction},
  author={Lin, Xuanqi and Zhang, Yong and Wang, Shun and Hu, Yongli and Yin, Baocai},
  journal={IEEE Transactions on Intelligent Transportation Systems},
  year={2025},
  volume={26},
  number={3},
  pages={3056-3070},
  publisher={IEEE}
}

@article{PCHGCN,
  title={PCHGCN: Physically Constrained Higher-order Graph Convolutional Network for Pedestrian Trajectory Prediction},
  author={Chen, Wangxing and Sang, Haifeng and Zhao, Zishan},
  journal={IEEE Internet of Things Journal},
  year={2025},
  publisher={IEEE}
}

@article{CMPT,
  title={A Completing Missing Pedestrian Trajectories Method Driven by Prior-Posterior Knowledge and Interactive Information},
  author={Duan, Mingxing and Zheng, Xinyue and Pi, Huilong and Ding, Yan and Tang, Zhuo},
  journal={IEEE Transactions on Intelligent Transportation Systems},
  year={2025},
  publisher={IEEE}
}

@article{hu2025cvae,
  title={A CVAE Combined With Diffusion Mechanism to Pedestrian Trajectory Prediction},
  author={Hu, Chuan and Lang, Peichuan and Chen, Hao and Yang, Biao and Zhu, Dianchen},
  journal={IEEE Robotics and Automation Letters},
  year={2025},
  volume={10},
  number={7},
  pages={7358-7364},
  publisher={IEEE}
}

@article{AFC-RNN,
  title={AFC-RNN: Adaptive Forgetting-Controlled Recurrent Neural Network for Pedestrian Trajectory Prediction},
  author={Dong, Yonghao and Wang, Le and Zhou, Sanping and Tang, Wei and Hua, Gang and Sun, Changyin},
  journal={IEEE Transactions on Pattern Analysis and Machine Intelligence},
  year={2025},
  publisher={IEEE}
}

@inproceedings{AMD,
  title={AMD: Adaptive Momentum and Decoupled Contrastive Learning Framework for Robust Long-Tail Trajectory Prediction},
  author={Rao, Bin and Liao, Haicheng and Guan, Yanchen and Wang, Chengyue and Wang, Bonan and Zhang, Jiaxun and Li, Zhenning},
  booktitle = {Proceedings of IEEE/CVF International Conference on Computer Vision},
  year={2025}
}

@article{ASTRA,
title={{ASTRA}: A Scene-aware Transformer-based Model for Trajectory Prediction},
author={Izzeddin Teeti and Aniket Thomas and Munish Monga and Sachin Kumar Giroh and Uddeshya Singh and Andrew Bradley and Biplab Banerjee and Fabio Cuzzolin},
journal={Transactions on Machine Learning Research},
issn={2835-8856},
year={2025},
note={}
}

@article{CINet,
  title={Completed Interaction Networks for Pedestrian Trajectory Prediction},
  author={Zhang, Zhong and Zhou, Jianglin and Liu, Shuang and Xiao, Baihua},
  journal={IEEE Transactions on Multimedia},
  year={2025},
  volume={27},
  number={},
  pages={5119-5129},
  publisher={IEEE}
}

@article{ganeshaaraj2025enhancing,
  title={Enhancing predictive performance on long-tail trajectories via clustering and specialized decoders},
  author={Ganeshaaraj, G and Fernando, Tharindu and Sridharan, Sridha and Fookes, Clinton},
  journal={Pattern Recognition},
  pages={112315},
  year={2025},
  publisher={Elsevier}
}

@article{STFlow,
  title={Flow Matching for Geometric Trajectory Simulation},
  author={Brinke, Kiet Bennema ten and Minartz, Koen and Menkovski, Vlado},
  journal={arXiv preprint arXiv:2505.18647},
  year={2025}
}

@inproceedings{mamba,
  title={Mamba: Linear-time sequence modeling with selective state spaces},
  author={Gu, Albert and Dao, Tri},
  booktitle={First conference on language modeling},
  year={2024}
}

@article{su2025improving,
  title={Improving generative trajectory prediction via collision-free modeling and goal scene reconstruction},
  author={Su, Zhaoxin and Huang, Gang and Zhou, Zhou and Li, Yongfu and Zhang, Sanyuan and Hua, Wei},
  journal={Pattern Recognition Letters},
  volume={188},
  pages={117--124},
  year={2025},
  publisher={Elsevier}
}

@inproceedings{LCD,
  title={Instantaneous Trajectory Prediction via Latent Bidirectional Cooperative Diffusion},
  author={Ma, Kun and Han, Qilong and Yao, Jingzheng and Wu, Changmao and Na, Chunrui},
  booktitle={IEEE International Conference on Acoustics, Speech and Signal Processing},
  pages={1--5},
  year={2025},
}

@article{IDM,
  title={Intention-aware denoising diffusion model for trajectory prediction},
  author={Liu, Chen and He, Shibo and Liu, Haoyu and Chen, Jiming},
  journal={IEEE Transactions on Intelligent Transportation Systems},
  year={2025},
  volume={26},
  number={5},
  pages={5915-5930},
}

@article{koopcast,
  title={KoopCast: Trajectory Forecasting via Koopman Operators},
  author={Lee, Jungjin and Shin, Jaeuk and Kim, Gihwan and Han, Joonho and Yang, Insoon},
  journal={arXiv preprint arXiv:2509.15513},
  year={2025 }
}

@article{huang2025learning,
  title={Learning Velocity and Acceleration: Self-Supervised Motion Consistency for Pedestrian Trajectory Prediction},
  author={Huang, Yizhou and Cheng, Yihua and Wang, Kezhi},
  journal={arXiv preprint arXiv:2503.24272},
  year={2025 }
}

@article{hop,
  title={Pedestrian trajectory prediction model based on self-supervised spatiotemporal graph network},
  author={Yang, Shiji and Xiao, Xuezhong},
  journal={Intelligent Systems with Applications},
  pages={200533},
  year={2025},
  publisher={Elsevier}
}

@inproceedings{shi2025leveraging,
  title={Leveraging Maps of Spatial Motion Patterns to Enhance Long-Term Adaptive Trajectory Prediction with Diffusion Models},
  author={Shi, Junyi and Kucner, Tomasz Piotr},
  booktitle={European Conference on Mobile Robots},
  pages={1--8},
  year={2025},
  organization={IEEE}
}

@article{STG-KNet,
  title={STG-KNet: A Kernel-mapping-based spatial-temporal graph convolution network for pedestrian trajectory prediction},
  author={Xu, Yuanzi and Yang, Jiafu and Cheng, Rongjun},
  journal={Physica A: Statistical Mechanics and its Applications},
  pages={130985},
  year={2025},
  publisher={Elsevier}
}

@article{UniEdge,
  title={Unified Spatial-Temporal Edge-Enhanced Graph Networks for Pedestrian Trajectory Prediction},
  author={Li, Ruochen and Qiao, Tanqiu and Katsigiannis, Stamos and Zhu, Zhanxing and Shum, Hubert PH},
  journal={IEEE Transactions on Circuits and Systems for Video Technology},
  year={2025},
  volume={35},
  number={7},
  pages={7047-7060},
  publisher={IEEE}
}

@article{MINet,
  title={MINet: A Pedestrian Trajectory Forecasting Method with Multi-Information Feature Fusion},
  author={Yuan, Tao and Han, Xiaohong},
  journal={Computing and Informatics},
  volume={44},
  number={1},
  pages={202--222},
  year={2025}
}

@inproceedings{MoFlow,
  title={Moflow: One-step flow matching for human trajectory forecasting via implicit maximum likelihood estimation based distillation},
  author={Fu, Yuxiang and Yan, Qi and Wang, Lele and Li, Ke and Liao, Renjie},
  booktitle={Proceedings of the Computer Vision and Pattern Recognition Conference},
  pages={17282--17293},
  year={2025}
}

@article{BRAINet,
  title={Traffic agents trajectory prediction based on enhanced bidirectional recurrent network and adaptive social interaction model},
  author={Chen, Xiaobo and Liang, Yuwen and Hu, Chuan and Wang, Hai and Ye, Qiaolin},
  journal={IEEE Transactions on Automation Science and Engineering},
  year={2025},
  publisher={IEEE}
}

@article{FTPN,
  title={Predict Multiple Steps At Once: A Fragment Trajectory Prediction Network},
  author={Qiu, Ruiqi and Gong, Jun and Cen, Yi and Jin, Jingyuan and Hou, Jinbao and Luo, Siqi},
  journal={IEEE Robotics and Automation Letters},
  year={2025},
  volume={10},
  number={7},
  pages={6720-6727},
  publisher={IEEE}
}

@article{FMTP,
  title={Remember and Recall: Associative-Memory-Based Trajectory Prediction},
  author={Guo, Hang and Zhang, Yuzhen and Gao, Tianci and Su, Junning and Lv, Pei and Xu, Mingliang},
  journal={IEEE Transactions on Intelligent Transportation Systems},
  year={2025},
  volume={26},
  number={9},
  pages={13898-13908},
  publisher={IEEE}
}

@article{SEI,
  title={Social Entropy Informer: A Multi-Scale Model-Data Dual-Driven Approach for Pedestrian Trajectory Prediction},
  author={Jiang, Zihan and Qin, Chengxuan and Yang, Rui and Shi, Bingyu and Alsaadi, Fuad E and Wang, Zidong},
  journal={IEEE Transactions on Intelligent Transportation Systems},
  year={2025},
  volume={26},
  number={10},
  pages={16438-16453},
  publisher={IEEE}
}

@inproceedings{damirchi2025socially,
  title={Socially-informed reconstruction for pedestrian trajectory forecasting},
  author={Damirchi, Haleh and Etemad, Ali and Greenspan, Michael},
  booktitle={IEEE/CVF Winter Conference on Applications of Computer Vision},
  pages={7460--7469},
  year={2025},
}

@inproceedings{SocialMOIF,
  title={SocialMOIF: Multi-Order Intention Fusion for Pedestrian Trajectory Prediction},
  author={Chen, Kai and Zhao, Xiaodong and Huang, Yujie and Fang, Guoyu and Song, Xiao and Wang, Ruiping and Wang, Ziyuan},
  booktitle={Proceedings of the Computer Vision and Pattern Recognition Conference},
  pages={22465--22475},
  year={2025}
}

@article{SocialTrans,
  title={SocialTrans: Transformer based social intentions interaction for pedestrian trajectory prediction},
  author={Chen, Kai and Zhao, Xiaodong and Huang, Yujie and Fang, Guoyu},
  journal={Physica A: Statistical Mechanics and its Applications},
  volume={663},
  pages={130435},
  year={2025},
  publisher={Elsevier}
}

@article{SceneAware,
  title={SceneAware: Scene-Constrained Pedestrian Trajectory Prediction with LLM-Guided Walkability},
  author={Bai, Juho and Shim, Inwook},
  journal={arXiv preprint arXiv:2506.14144},
  year={2025}
}

@article{kothari2021,
  title={Human trajectory forecasting in crowds: A deep learning perspective},
  author={Kothari, Parth and Kreiss, Sven and Alahi, Alexandre},
  journal={IEEE Transactions on Intelligent Transportation Systems},
  volume={23},
  number={7},
  pages={7386--7400},
  year={2021},
}

@inproceedings{liu2024auto,
  title={Auto-encoding bayesian inverse games},
  author={Liu, Xinjie and Peters, Lasse and Alonso-Mora, Javier and Topcu, Ufuk and Fridovich-Keil, David},
  booktitle={Proceedings of the Workshop on Algorithmic Foundations of Robotics},
  year={2024}
}

@article{sun2025inverse,
  title={Inverse Mixed Strategy Games with Generative Trajectory Models},
  author={Sun, Max Muchen and Trautman, Pete and Murphey, Todd},
  journal={arXiv preprint arXiv:2502.03356},
  year={2025 }
}

@article{diehl2023energy,
  title={Energy-based potential games for joint motion forecasting and control},
  author={Diehl, Christopher and Klosek, Tobias and Krueger, Martin and Murzyn, Nils and Osterburg, Timo and Bertram, Torsten},
  journal={arXiv preprint arXiv:2312.01811},
  year={2023}
}

@inproceedings{lidard2024blending,
  title={Blending data-driven priors in dynamic games},
  author={Lidard, Justin and Hu, Haimin and Hancock, Asher and Zhang, Zixu and Contreras, Albert Gim{\'o} and Modi, Vikash and DeCastro, Jonathan and Gopinath, Deepak and Rosman, Guy and Leonard, Naomi Ehrich and others},
  pages={20-39},
  booktitle={Proceedings of Robotics: Science and Systems},
  year={2024}
}

@article{schwarting2019social,
  title={Social behavior for autonomous vehicles},
  author={Schwarting, Wilko and Pierson, Alyssa and Alonso-Mora, Javier and Karaman, Sertac and Rus, Daniela},
  journal={Proceedings of the National Academy of Sciences},
  volume={116},
  number={50},
  pages={24972--24978},
  year={2019},
  publisher={National Academy of Sciences}
}

@inproceedings{MTP,
  title={Multimodal trajectory prediction via topological invariance for navigation at uncontrolled intersections},
  author={Roh, Junha and Mavrogiannis, Christoforos and Madan, Rishabh and Fox, Dieter and Srinivasa, Siddhartha},
  booktitle={Conference on Robot Learning},
  pages={2216--2227},
  year={2021},
  organization={PMLR}
}

@article{SHINE,
  title={SHINE: Social homology identification for navigation in crowded environments},
  author={Martinez-Baselga, Diego and de Groot, Oscar and Knoedler, Luzia and Riazuelo, Luis and Alonso-Mora, Javier and Montano, Luis},
  journal={The International Journal of Robotics Research},
  pages={02783649251344639},
  year={2024},
  publisher={SAGE Publications Sage UK: London, England}
}

@inproceedings{SRefiner,
  title={SRefiner: Soft-Braid Attention for Multi-Agent Trajectory Refinement},
  author={Xiao, Liwen and Pan, Zhiyu and Wang, Zhicheng and Cao, Zhiguo and Li, Wei},
  booktitle={IEEE/CVF International Conference on Computer Vision},
  pages={960--969},
  year={2025}
}

@article{stratton2024,
  title={Characterizing the Complexity of Social Robot Navigation Scenarios},
  author={Stratton, Andrew and Hauser, Kris and Mavrogiannis, Christoforos},
  journal={IEEE Robotics and Automation Letters},
  year={2024},
  volume={10},
  number={1},
  pages={184-191},
  publisher={IEEE}
}

@misc{NBAdata,   
    title = {SportVU Data},
    year = {2016},
    howpublished={Available at \url{https://github.com/rajshah4/BasketballData/tree/master/2016.NBA.Raw.SportVU.Game.Logs}},
}

@article{francis2025principles,
  title={Principles and guidelines for evaluating social robot navigation algorithms},
  author={Francis, Anthony and P{\'e}rez-d’Arpino, Claudia and Li, Chengshu and Xia, Fei and Alahi, Alexandre and Alami, Rachid and Bera, Aniket and Biswas, Abhijat and Biswas, Joydeep and Chandra, Rohan and others},
  journal={ACM Transactions on Human-Robot Interaction},
  volume={14},
  number={2},
  pages={1--65},
  year={2025},
  publisher={ACM New York, NY}
}

@article{zhang2022hri,
  title={From {H}{R}{I} to {C}{R}{I}: crowd Robot interaction—understanding the effect of robots on crowd motion: empirical study of pedestrian dynamics with a wheelchair and a {P}epper robot},
  author={Zhang, Bingqing and Amirian, Javad and Eberle, Harry and Pettr{\'e}, Julien and Holloway, Catherine and Carlson, Tom},
  journal={International Journal of Social Robotics},
  volume={14},
  number={3},
  pages={631--643},
  year={2022},
  publisher={Springer}
}

@article{Gneiting,
  title   = {Strictly Proper Scoring Rules, Prediction, and Estimation},
  author  = {Gneiting, Tilmann and Raftery, Adrian E.},
  journal = {Journal of the American Statistical Association},
  volume  = {102},
  number  = {477},
  pages   = {359--378},
  year    = {2007}
}

@article{Scheuerer,
  title   = {Variogram-Based Proper Scoring Rules for Probabilistic Forecasts of Multivariate Quantities},
  author  = {Scheuerer, Michael and Hamill, Thomas M.},
  journal = {Monthly Weather Review},
  volume  = {143},
  number  = {4},
  pages   = {1321--1334},
  year    = {2015},
}

@article{canche,
  title={Calibrating uncertainties in human trajectory forecasting},
  author={Canche, Mario and Morales Quispe, Marcela and Hayet, Jean-Bernard},
  journal={Machine Vision and Applications},
  volume={36},
  number={3},
  pages={1--13},
  year={2025},
  publisher={Springer}
}

@inproceedings{poddar2023crowd,
  title={From crowd motion prediction to robot navigation in crowds},
  author={Poddar, Sriyash and Mavrogiannis, Christoforos and Srinivasa, Siddhartha S},
  booktitle={IEEE/RSJ International Conference on Intelligent Robots and Systems},
  pages={6765--6772},
  year={2023},
}

@inproceedings{nishimura2020risk,
  title={Risk-sensitive sequential action control with multi-modal human trajectory forecasting for safe crowd-robot interaction},
  author={Nishimura, Haruki and Ivanovic, Boris and Gaidon, Adrien and Pavone, Marco and Schwager, Mac},
  booktitle={IEEE/RSJ International Conference on Intelligent Robots and Systems},
  pages={11205--11212},
  year={2020},
}

@inproceedings{securing,
  title={Securing multimedia-based personal data: towards a methodology for automated anonymization risk assessment seeking GDPR compliance},
  author={Aramburu, Mikel and Red{\'o}, David and Garcia-Casta{\~n}o, Jorge},
  booktitle={Artificial Intelligence for Security and Defence Applications II},
  volume={13206},
  pages={106--120},
  year={2024},
  organization={SPIE}
}

@article{ano,
    author = {Weitzenboeck, Emily M and Lison, Pierre and Cyndecka, Malgorzata and Langford, Malcolm},
    title = {The GDPR and unstructured data: is anonymization possible?},
    journal = {International Data Privacy Law},
    volume = {12},
    number = {3},
    pages = {184-206},
    year = {2022},
    month = {03},
    issn = {2044-3994},
    eprint = {https://academic.oup.com/idpl/article-pdf/12/3/184/45690911/ipac008.pdf},
}

@inproceedings{motsynth,
  title={Motsynth: How can synthetic data help pedestrian detection and tracking?},
  author={Fabbri, Matteo and Bras{\'o}, Guillem and Maugeri, Gianluca and Cetintas, Orcun and Gasparini, Riccardo and O{\v{s}}ep, Aljo{\v{s}}a and Calderara, Simone and Leal-Taix{\'e}, Laura and Cucchiara, Rita},
  booktitle={Proceedings of the IEEE/CVF International Conference on Computer Vision},
  pages={10849--10859},
  year={2021}
}

@inproceedings{chefer2021transformer,
  title={Transformer interpretability beyond attention visualization},
  author={Chefer, Hila and Gur, Shir and Wolf, Lior},
  booktitle={IEEE/CVF Conference on Computer Vision and Pattern Recognition},
  pages={782--791},
  year={2021}
}

@article{makansi2021you,
  title={You mostly walk alone: Analyzing feature attribution in trajectory prediction},
  author={Makansi, Osama and Von K{\"u}gelgen, Julius and Locatello, Francesco and Gehler, Peter and Janzing, Dominik and Brox, Thomas and Sch{\"o}lkopf, Bernhard},
  journal={arXiv preprint arXiv:2110.05304},
  year={2021}
}

@inproceedings{hewitt2019structural,
  title={A structural probe for finding syntax in word representations},
  author={Hewitt, John and Manning, Christopher D},
  booktitle={Proceedings of the 2019 Conference of the North American Chapter of the Association for Computational Linguistics: Human Language Technologies, Volume 1 (Long and Short Papers)},
  pages={4129--4138},
  year={2019}
}

@article{zucker,
title={A Study in Zucker: Insights on Interactions Between Humans and Small Service Robots}, 
author={Day, Alex and Karamouzas, Ioannis},
journal = {IEEE Robotics and Automation Letters},
volume  = {9},
number={3},
year={2024}
}

@misc{gpt,
  title={Improving language understanding by generative pre-training},
  author={Radford, Alec and Narasimhan, Karthik and Salimans, Tim and Sutskever, Ilya and others},
  year={2018},
  publisher={San Francisco, CA, USA}
}

@article{ARP-STGCN,
  title={ARP-STGCN: a fast attraction--repulsion-potential based spatio-temporal graph convolutional network with imputation for pedestrian trajectory prediction},
  author={Fang, Bin and Qin, Fangtao and Wang, Yi},
  journal={Complex \& Intelligent Systems},
  volume={11},
  number={11},
  pages={1--20},
  year={2025},
  publisher={Springer}
}

@article{CICR,
  title={Causal Intervention and Counterfactual Reasoning for Multimodal Pedestrian Trajectory Prediction},
  author={Han, Xinyu and Xu, Huosheng},
  journal={Journal of Imaging},
  volume={11},
  number={11},
  pages={379},
  year={2025},
  publisher={MDPI}
}

@inproceedings{zou2025walks,
  title={Who walks with you matters: Perceiving social interactions with groups for pedestrian trajectory prediction},
  author={Zou, Ziqian and Wong, Conghao and Xia, Beihao and You, Xinge},
  booktitle={IEEE/CVF International Conference on Computer Vision},
  pages={4844--4853},
  year={2025}
}

@article{CDDM,
  title={Collaborative-Distilled Diffusion Models (CDDM) for Accelerated and Lightweight Trajectory Prediction},
  author={Wang, Bingzhang and Chen, Kehua and Wang, Yinhai},
  journal={arXiv preprint arXiv:2510.00627},
  year={2025}
}

@inproceedings{DiffRefiner,
  title={Diff-Refiner: Enhancing Multi-Agent Trajectory Prediction with a Plug-and-Play Diffusion Refiner},
  author={Zhou, Xiangzheng and Chen, Xiaobo and Yang, Jian},
  booktitle={IEEE International Conference on Robotics and Automation},
  pages={10779--10785},
  year={2025},
}

@article{Diffusion2,
  title={Diffusion\^{} 2: Dual Diffusion Model with Uncertainty-Aware Adaptive Noise for Momentary Trajectory Prediction},
  author={Luo, Yuhao and Zhang, Yuang and Chen, Kehua and Zheng, Xinyu and Zhang, Shucheng and Chen, Sikai and Wang, Yinhai},
  journal={arXiv preprint arXiv:2510.04365},
  year={2025 }
}

@inproceedings{Egocentric,
  title={Egocentric Conformal Prediction for Safe and Efficient Navigation in Dynamic Cluttered Environments},
booktitle={Proceedings of the  IEEE Conference on Decision and Control (CDC)},
  author={Shin, Jaeuk and Lee, Jungjin and Yang, Insoon},
  year={2025}
  }

@article{egotraj,
  title={EgoTraj-Bench: Towards Robust Trajectory Prediction Under Ego-view Noisy Observations},
  author={Liu, Jiayi and Zhou, Jiaming and Ye, Ke and Lin, Kun-Yu and Wang, Allan and Liang, Junwei},
  journal={arXiv preprint arXiv:2510.00405},
  year={2025}
}

@inproceedings{ren2025totp,
  title={TOTP: Transferable Online Pedestrian Trajectory Prediction with Temporal-Adaptive Mamba Latent Diffusion},
  author={Ren, Ziyang and Wei, Ping and Deng, Shangqi and Tang, Haowen and Li, Jiapeng and Li, Huan},
  booktitle={Proceedings of the IEEE/CVF International Conference on Computer Vision},
  pages={26263--26272},
  year={2025}
}

@inproceedings{ren2025stochastic,
  title={Stochastic-Aware Mamba Diffusion for Pedestrian Trajectory Prediction},
  author={Ren, Ziyang and Wei, Ping and Tang, Haowen and Li, Huan and Yang, Jin and Qin, Jialu},
  booktitle={IEEE International Conference on Acoustics, Speech and Signal Processing},
  pages={1--5},
  year={2025},
}

@inproceedings{TBD,
  title={TBD pedestrian data collection: Towards rich, portable, and large-scale natural pedestrian data},
  author={Wang, Allan and Sato, Daisuke and Corzo, Yasser and Simkin, Sonya and Biswas, Abhijat and Steinfeld, Aaron},
  booktitle={IEEE International Conference on Robotics and Automation},
  pages={637--644},
  year={2024},
  organization={IEEE}
}

@article{OVSKTGCNN,
  title={Enhanced pedestrian trajectory prediction via overlapping field-of-view domains and integrated Kolmogorov-Arnold networks},
  author={Wang, Hongxia and Liu, Yang and Nie, Zhenkai},
  journal={PLoS One},
  volume={20},
  number={6},
  pages={e0322722},
  year={2025},
  publisher={Public Library of Science San Francisco, CA USA}
}

@inproceedings{IAN,
  title={Interactive Adjustment for Human Trajectory Prediction with Individual Feedback},
  author={Sun, Jianhua and Li, Yuxuan and Chai, Liang and Lu, Cewu},
    year = 2025,
  pages = {13164--13183},
  booktitle={International Conference on Learning Representations}
}

@inproceedings{LGTraj,
  title={Lg-traj: Llm guided pedestrian trajectory prediction},
  author={Chib, Pranav Singh and Singh, Pravendra},
  booktitle={Proceedings of the IEEE/CVF International Conference on Computer Vision},
  pages={6802--6812},
  year={2025}
}

@article{MambaPTP,
  title={MambaPTP: Exploring the Potential of Mamba for Pedestrian Trajectory Prediction},
  author={Zhang, Shuangqing and Zhao, Gangming and Lyu, Fan and Wang, Songping and Zhang, Zhang and Zhao, Fang and Li, Jinpeng and Shan, Caifeng and Wang, Liang},
  journal={IEEE Transactions on Circuits and Systems for Video Technology},
  year={2025},
  pages={1-1},
  publisher={IEEE}
}

@article{aslam2025,
  title={Model Predictive Control for Crowd Navigation via Learning-Based Trajectory Prediction},
  author={Aslam, Mohamed Parvez and Derajic, Bojan and Bouzidi, Mohamed-Khalil and Bernhard, Sebastian and Ringert, Jan Oliver},
  journal={arXiv preprint arXiv:2508.07079},
  year={2025}
}

@inproceedings{NATRA,
  title={NATRA: Noise-Agnostic Framework for Trajectory Prediction with Noisy Observations},
  author={Li, Rongqing and Li, Changsheng and Lv, Ruilin and Li, Yuhang and Gao, Yang and Zhang, Xiaolu and Zhou, Jun},
  booktitle={Proceedings of the IEEE/CVF International Conference on Computer Vision},
  pages={27872--27884},
  year={2025}
}

@inproceedings{NMRF,
  title={Neuralized Markov Random Field for Interaction-Aware Stochastic Human Trajectory Prediction},
  author={Fang, Zilin and Hsu, David and Lee, Gim Hee},
  booktitle={The Thirteenth International Conference on Learning Representations},
 pages = {73878--73893},
  year={2025}
}

@article{GDDDL,
  title={Pedestrian trajectory prediction using goal-driven and dynamics-based deep learning framework},
  author={Wang, Honghui and Zhi, Weiming and Batista, Gustavo and Chandra, Rohitash},
  journal={Expert Systems with Applications},
  volume={271},
  pages={126557},
  year={2025},
  publisher={Elsevier}
}

@article{PPAL,
  title={Pedestrian trajectory prediction via physical-guided position association learning},
  author={Xu, Yueyun and Qin, Hongmao and Bian, Yougang and Ding, Rongjun},
  journal={Engineering Science and Technology, an International Journal},
  volume={64},
  pages={102008},
  year={2025},
  publisher={Elsevier}
}

@article{SASGM,
  title={Probabilistic and Interaction-Aware Trajectory Prediction Using Score-Based Diffusion Models},
  author={Han, Peihua and Zhu, Mingda and Tian, Weiwei and Zhang, Houxiang},
  journal={IEEE Transactions on Industrial Informatics},
  year={2025},
  pages={1-10},
  publisher={IEEE}
}

@inproceedings{uhlemann2025snapshot,
  title={Snapshot: Towards Application-centered Models for Pedestrian Trajectory Prediction in Urban Traffic Environments},
  author={Uhlemann, Nico and Zhou, Yipeng and Mohr, Tobias Simeon and Lienkamp, Markus},
  booktitle={Proceedings of the Winter Conference on Applications of Computer Vision},
  pages={1152--1162},
  year={2025}
}

@inproceedings{resonance,
  title={Resonance: Learning to Predict Social-Aware Pedestrian Trajectories as Co-Vibrations},
  author={Wong, Conghao and Zou, Ziqian and Xia, Beihao},
  booktitle={Proceedings of the IEEE/CVF International Conference on Computer Vision},
  pages={25788--25799},
  year={2025}
}

@inproceedings{ACE,
  title={Sim-to-real causal transfer: A metric learning approach to causally-aware interaction representations},
  author={Rahimi, Ahmad and Luan, Po-Chien and Liu, Yuejiang and Raji{\v{c}}, Frano and Alahi, Alexandre},
  booktitle={Proceedings of the Computer Vision and Pattern Recognition Conference},
  pages={17271--17281},
  year={2025}
}

@techreport{data,
  title={Data, privacy laws and firm production: Evidence from the GDPR},
  author={Demirer, Mert and Hern{\'a}ndez, Diego J Jim{\'e}nez and Li, Dean and Peng, Sida},
  year={2024},
  institution={National Bureau of Economic Research}
}

@article{biswas2022socnavbench,
  title={Socnavbench: A grounded simulation testing framework for evaluating social navigation},
  author={Biswas, Abhijat and Wang, Allan and Silvera, Gustavo and Steinfeld, Aaron and Admoni, Henny},
  journal={ACM Transactions on Human-Robot Interaction (THRI)},
  volume={11},
  number={3},
  pages={1--24},
  year={2022},
  publisher={ACM New York, NY}
}

@article{gao2022evaluation,
  title={Evaluation of socially-aware robot navigation},
  author={Gao, Yuxiang and Huang, Chien-Ming},
  journal={Frontiers in Robotics and AI},
  volume={8},
  pages={721317},
  year={2022},
  publisher={Frontiers Media SA}
}

@article{dik2024graph,
  title={Graph network-based human movement prediction for socially-aware robot navigation in shared workspaces},
  author={Dik, Casper and Emmanouilidis, Christos and Duqueroie, Bertrand},
  journal={Neural Computing and Applications},
  volume={36},
  number={34},
  pages={21743--21759},
  year={2024},
  publisher={Springer}
}

@inproceedings{franchi,
  title={Crowd behavior characterization for scene tracking},
  author={Franchi, Gianni and Aldea, Emanuel and Dubuisson, S{\'e}verine and Bloch, Isabelle},
  booktitle={IEEE International Conference on Advanced Video and Signal Based Surveillance},
  pages={1--8},
  year={2019},
}

@inproceedings{kuleshov,
  title={Accurate uncertainties for deep learning using calibrated regression},
  author={Kuleshov, Volodymyr and Fenner, Nathan and Ermon, Stefano},
  booktitle={International Conference on Machine Learning},
  pages={2796--2804},
  year={2018},
  organization={PMLR}
}

@article{battiston2025higher,
  title={Higher-order interactions shape collective human behaviour},
  author={Battiston, Federico and Capraro, Valerio and Karimi, Fariba and Lehmann, Sune and Migliano, Andrea Bamberg and Sadekar, Onkar and S{\'a}nchez, Angel and Perc, Matja{\v{z}}},
  journal={Nature Human Behaviour},
  pages={1--17},
  year={2025},
  publisher={Nature Publishing Group UK London}
}

@article{xu2024sports,
  title={Sports-traj: A unified trajectory generation model for multi-agent movement in sports},
  author={Xu, Yi and Fu, Yun},
  journal={arXiv preprint arXiv:2405.17680},
  year={2024}
}

@book{pelechano2008virtual,
  title={Virtual Crowds: Methods, Simulation, and Control},
  author={Pelechano, Nuria and Allbeck, Jan M. and Badler, Norman I.},
  year={2008},
  publisher={Morgan \& Claypool Publishers}
}

@book{pelechano2016simulating,
  title={Simulating Heterogeneous Crowds with Interactive Behaviors},
  author={Pelechano, Nuria and Allbeck, Jan and Kapadia, Mubbasir and Badler, Norman I.},
  year={2016},
  publisher={CRC Press}
}

@article{lemonari2022authoring,
  title={Authoring Virtual Crowds: A Survey},
  author={Lemonari, Marilena and Blanco, Rafael and Charalambous, Panayiotis and Pelechano, Nuria and Avraamides, Marios N. and Pettr{\'e}, Julien and Chrysanthou, Yiorgos},
  journal={Computer Graphics Forum},
  volume={41},
  number={2},
  pages={677--701},
  year={2022},
  publisher={Wiley Online Library}
}

@article{li2019mobile,
  title={Mobile targeting using customer trajectory patterns},
  author={Li, Beibei and Luo, Xueming and Zhang, Xiaoyi and Wang, Fang},
  journal={Management Science},
  volume={65},
  number={11},
  pages={5027--5049},
  year={2019},
  publisher={INFORMS}
}

@inproceedings{halilovic2025explainable,
  title={Explainable robot navigation},
  author={Halilovic, Amar},
  booktitle={Proceedings of the AAAI Conference on Artificial Intelligence},
  number={28},
  pages={29261--29262},
  year={2025}
}

@article{leila24,
author = {Methnani, Leila and Chiou, Manolis and Dignum, Virginia and Theodorou, Andreas},
title = {Who’s in Charge Here? A Survey on Trustworthy AI in Variable Autonomy Robotic Systems},
year = {2024},
issue_date = {July 2024},
volume = {56},
number = {7},
doi = {10.1145/3645090},
journal = {ACM Comput. Surv.},
}

@article{wang2025trends,
  title={Trends in motion prediction toward deployable and generalizable autonomy: A revisit and perspectives},
  author={Wang, Letian and Lavoie, Marc-Antoine and Papais, Sandro and Nisar, Barza and Chen, Yuxiao and Ding, Wenhao and Ivanovic, Boris and Shao, Hao and Abuduweili, Abulikemu and Cook, Evan and others},
  journal={arXiv preprint arXiv:2505.09074},
  year={2025}
}

@inproceedings{brown2020,
author = {Brown, Tom B. and Mann, Benjamin and Ryder, Nick and Subbiah, Melanie and Kaplan, Jared and Dhariwal, Prafulla and Neelakantan, Arvind and Shyam, Pranav and Sastry, Girish and Askell, Amanda and Agarwal, Sandhini and Herbert-Voss, Ariel and Krueger, Gretchen and Henighan, Tom and Child, Rewon and Ramesh, Aditya and Ziegler, Daniel M. and Wu, Jeffrey and Winter, Clemens and Hesse, Christopher and Chen, Mark and Sigler, Eric and Litwin, Mateusz and Gray, Scott and Chess, Benjamin and Clark, Jack and Berner, Christopher and McCandlish, Sam and Radford, Alec and Sutskever, Ilya and Amodei, Dario},
title = {Language models are few-shot learners},
year = {2020},
ooktitle = {International Conference on Neural Information Processing Systems},
}

@article{hu2020predicting,
  title={Predicting crowd egress and environment relationships to support building design optimization},
  author={Hu, Kaidong and Yoon, Sejong and Pavlovic, Vladimir and Faloutsos, Petros and Kapadia, Mubbasir},
  journal={Computers \& Graphics},
  volume={88},
  pages={83--96},
  year={2020},
  publisher={Elsevier}
}

@article{egress2017,
author = {Cassol, Vincius J. and Smania Testa, Est\^{e}v\~{a}o and Rosito Jung, Cl\'{a}udio and Usman, Muhammad and Faloutsos, Petros and Berseth, Glen and Kapadia, Mubbasir and Badler, Norman I. and Raupp Musse, Soraia},
title = {Evaluating and Optimizing Evacuation Plans for Crowd Egress},
year = {2017},
issue_date = {2017},
volume = {37},
number = {4},
doi = {10.1109/MCG.2017.3271454},
journal = {IEEE Computer Graphics and Applications},
pages = {60–71},
numpages = {12}
}

@inproceedings{houston2021one,
  title={One thousand and one hours: Self-driving motion prediction dataset},
  author={Houston, John and Zuidhof, Guido and Bergamini, Luca and Ye, Yawei and Chen, Long and Jain, Ashesh and Omari, Sammy and Iglovikov, Vladimir and Ondruska, Peter},
  booktitle={Conference on Robot Learning},
  pages={409--418},
  year={2021},
  organization={PMLR}
}

@inproceedings{edie,
    author = {Edie, Leslie C},
    title = {Discussion of traffic stream measurements and definitions} ,
    booktitle = {Proceedings of the 2nd International Symposium on the Theory
of Traffic Flow},
    year = {1963}, 
    pages = {139-154}
}

@article{karamouzas2014universal,
  title={Universal power law governing pedestrian interactions},
  author={Karamouzas, Ioannis and Skinner, Brian and Guy, Stephen J},
  journal={Physical Review Letters},
  volume={113},
  number={23},
  pages={238701},
  year={2014},
}
\end{document}